\documentclass{article}

\PassOptionsToPackage{numbers}{natbib}
    \usepackage[final]{neurips_2025}

\usepackage[utf8]{inputenc} % allow utf-8 input
\usepackage[T1]{fontenc}    % use 8-bit T1 fonts
\usepackage{hyperref}       % hyperlinks
\usepackage{url}            % simple URL typesetting
\usepackage{booktabs}       % professional-quality tables
\usepackage{amsfonts}       % blackboard math symbols
\usepackage{nicefrac}       % compact symbols for 1/2, etc.
\usepackage{microtype}      % microtypography
\usepackage{xcolor}         % colors
\definecolor{grey}{rgb}{0.8,0.8,0.8}
\definecolor{aqua}{rgb}{0, 1, 1}
\definecolor{steel}{rgb}{0.2734, 0.5078, 0.7031}
\definecolor{slate}{rgb}{0.1836, 0.3086, 0.3086}
\newcommand{\hlg}[2]{\setlength{\fboxsep}{0.3pt}\colorbox{green!#2}{\rule[-.05\baselineskip]{0pt}{.7\baselineskip}{#1}}}
\newcommand{\hlr}[2]{\setlength{\fboxsep}{0.3pt}\colorbox{red!#2}{\rule[-.05\baselineskip]{0pt}{.7\baselineskip}{#1}}}
\newcommand{\hlb}[2]{\setlength{\fboxsep}{0.3pt}\colorbox{aqua!#2}{\rule[-.05\baselineskip]{0pt}{.7\baselineskip}{#1}}}

\usepackage{amsmath,amsfonts,bm}

\def\eqref#1{equation~\ref{#1}}
\def\1{\bm{1}}

\def\mF{{\bm{F}}}

\def\mH{{\bm{H}}}
\def\mI{{\bm{I}}}

\def\mQ{{\bm{Q}}}

\def\mLambda{{\bm{\Lambda}}}

\DeclareMathAlphabet{\mathsfit}{\encodingdefault}{\sfdefault}{m}{sl}
\SetMathAlphabet{\mathsfit}{bold}{\encodingdefault}{\sfdefault}{bx}{n}

\newcommand{\mtheta}{\boldsymbol{\theta}}
\newcommand{\mphi}{\boldsymbol{\phi}}
\newcommand{\malpha}{\boldsymbol{\alpha}}

\newcommand{\dG}{\mathcal{G}}
\newcommand{\dL}{\mathcal{L}}

\usepackage{graphicx}
\usepackage{subfig}
\usepackage{enumitem}
\usepackage{xspace}
\usepackage{amsthm}
\usepackage{booktabs}
\usepackage{multirow}
\usepackage{subfig}

\usepackage{amsfonts}
\newtheorem{proposition}{Proposition}

\usepackage{url}
\usepackage{adjustbox}
\usepackage{multirow}
\usepackage{float}

\newcommand{\proj}{DWiSE-FT\xspace}
\newcommand{\projj}{\textsc{RoFt-Mol}\xspace}

\title{RoFt-Mol: Benchmarking \underline{Ro}bust \underline{F}ine-\underline{T}uning with \underline{Mol}ecular Graph Foundation Models}

\author{%
  Shikun Liu\thanks{Equal contribution.} $\ ^1$
  \quad Deyu Zou\footnotemark[1]  $\ ^2$
  \quad Nima Shoghi  $^1$
  \quad Victor Fung  $^1$
  \quad Kai Liu  $^3$
  \quad Pan Li  $^{1}$\\
    $^1\ $Georgia Institute of Technology
    $^2\ $The Chinese University of Hong Kong\\
    $^3\ $SES AI\\
}

\begin{document}

\maketitle

\vspace{-5mm}
\begin{abstract}
In the era of foundation models, fine-tuning pre-trained models for specific downstream tasks has become crucial. This drives the need for robust fine-tuning methods to address challenges such as model overfitting and sparse labeling. Molecular graph foundation models (MGFMs) face unique difficulties that complicate fine-tuning. These models are limited by smaller pre-training datasets and more severe data scarcity for downstream tasks, both of which require enhanced model generalization. Moreover, MGFMs must accommodate diverse objectives, including both regression and classification tasks. To better understand and improve fine-tuning techniques under these conditions, we classify eight fine-tuning methods into three mechanisms: weight-based, representation-based, and partial fine-tuning. We benchmark these methods on downstream regression and classification tasks across supervised and self-supervised pre-trained models in diverse labeling settings. This extensive evaluation provides valuable insights and informs the design of a refined robust fine-tuning method, \projj. This approach combines the strengths of simple post-hoc weight interpolation with more complex weight ensemble fine-tuning methods, delivering improved performance across both task types while maintaining the ease of use inherent in post-hoc weight interpolation. \footnote{
Our code and data are available at: \url{https://github.com/Graph-COM/RoFt-Mol}}
\end{abstract}

\vspace{-5mm}
\section{Introduction}
\label{sec:intro}

In recent years, foundation models~\citep{bommasani2021opportunities, zhou2023comprehensive} have achieved success in learning high-quality, general-purpose representations of images and text through pre-training on diverse datasets~\citep{radford2021learning, kirillov2023segment, ramesh2022hierarchical, touvron2023llama, bubeck2023sparks, zhao2023survey}. To adapt these pre-trained models for downstream applications, additional training on task-specific data, known as fine-tuning, is often required. However, vanilla fine-tuning frequently encounters challenges, including model overfitting~\citep{howard2018universal, lirethinking, kornblith2019better}, catastrophic forgetting of pre-trained knowledge~\citep{leesurgical, lidelta, xuhong2018explicit, lubana2022quadratic}, and distribution shifts between fine-tuned and test samples, which can lead to negative transfer~\citep{wang2019characterizing, chen2019catastrophic}. These challenges highlight the need for robust fine-tuning strategies~\citep{shen2021partial, wortsman2022robust, kumar2022fine, shu2023clipood, andreassenevolution, kirichenkolast}.

Recently, the advantages of foundation models have been extended to various scientific applications~\citep{golling2024masked, leung2024towards, nguyen2023climax}. Among these, molecular graph foundation models (MGFMs) have gained significant attention for their promising potential in biochemistry~\citep{hu2020strategies, hou2022graphmae, xiamole, suresh2021adversarial, shoghimolecules, beainitowards, zheng2023towards, ross2022large, rong2020self, maoposition}. While MGFMs exhibit scaling behaviors similar to foundation models in other domains~\citep{chen2024uncovering}, they face unique challenges related to data and tasks. %, which necessitate a more careful design of their fine-tuning strategies.

A primary challenge stems from the significantly smaller pre-training datasets in this domain, typically consisting of at most $O(100M)$ molecular samples, compared to the billions of samples used in other domains~\citep{sun2022does}. This limitation restricts the parameter scale of MGFMs ($O(100M)$ parameters) and their generalization capacity~\citep{wang2024evaluating, akhondzadeh2023probing}. Furthermore, downstream tasks in this domain often involve limited data for fine-tuning, with datasets containing only tens or a few hundred labeled samples~\citep{wijaya2024two}, exacerbating the difficulty of achieving robust model generalization. In addition to data constraints, many downstream tasks, such as molecular property prediction, are regression-based~\citep{wu2018moleculenet, hou2022accurate}. These tasks require models to capture fine-grained numerical patterns, which presents a distinct requirement compared to the coarse-grained feature reliance typical in classification tasks in CV and NLP. These factors collectively highlight the need for a careful examination of fine-tuning strategies for MGFMs and their appropriate improvement.
% Consequently, understanding fine-tuning strategies for MGFMs become crucial, mainly in terms of the factors that makes good fine-tuning in practice. 

To answer this question, we introduce \projj, a benchmark that evaluates existing fine-tuning methods across diverse molecular property prediction tasks. To explore factors influencing the fine-tuning (FT) performance of MGFMs, we categorize 8 FT methods into 3 distinct mechanisms: 1) \textit{weight-based FT}, which ensembles the weights from both pre-trained and fine-tuned models, 2) \textit{representation-based FT}, which regularizes the proximity between pre-trained and fine-tuned latent data representations, and 3) \textit{partial FT}, which optimizes only a subset of the pre-trained model’s weights while keeping the rest frozen. 
% To explore the connections of distinct fine-tuning mechanisms with different pre-training strategies and downstream task types, we experiment with 6 pre-trained models from both self-supervised and supervised pre-training including both pure graph-based, graph transformer based and multi-modal model with different model scales, then we evaluate on diverse molecular property prediction tasks, including 8 classification and 4 regression tasks. 
To derive generalizable insights into how different fine-tuning mechanisms interact with pre-training strategies and downstream task types, we evaluate six diverse pre-trained models, spanning self-supervised and supervised learning, with pure graph-based, graph transformer based and multi-modal models in varying scales, then evaluate on a broad set of molecular property prediction tasks, including 8 classification and 4 regression tasks.
To simulate the challenges encountered during the fine-tuning stages of MGFMs, we further consider the few-shot and out-of-distribution settings. Drawing from the broad range of pre-trained models and downstream tasks, we indeed find that the choice of best fine-tuning mechanism is highly determined by the \textit{pre-training objective} and the \textit{downstream task type}. We summarize high-level insights as follows, with further detailed results presented in Sec.~\ref{sec:result}. The bold text within brackets indicates the corresponding support in the experiment sections for clear cross-referencing:

\vspace{-0.5mm}
\begin{itemize}[leftmargin=*, nolistsep]
\item \textbf{Impact from Supervised vs. Self-supervised pre-trained models:}
Supervised pre-training learns domain-specific information with task supervision, while self-supervised pre-training captures general-purpose knowledge through training on generic synthetic tasks. We observe that, in few shot fine-tuning, supervised pre-training generally yields better fine-tuning performance than self-supervised pre-training even when the pre-training tasks do not align well with the fine-tuning tasks. In contrast, for non-few-shot settings, supervised pre-training performs better only when the supervised pre-training tasks closely align with the downstream tasks [\textbf{Q2}].
% \shikun{revise the language here}

\item \textbf{Impact from Classification vs. Regression tasks:}
Regression tasks need more precise numerical labels and finer molecule modeling. Therefore, MGFMs face less risk of overfitting in regression tasks compared to classification tasks, particularly in the few-shot setting [\textbf{Q1}].

\item \textbf{Correspondence with different fine-tuning methods:}
For self-supervised pre-trained models, \textit{weight-based fine-tuning} often results in better performance by effectively integrating general knowledge from pre-training with task-specific knowledge from fine-tuning [\textbf{Finding 1}]. On the other hand, \textit{partial fine-tuning} typically leads to underfitted molecular representations in few-shot fine-tuning, particularly for regression tasks [\textbf{Finding 2}]. For supervised pre-trained models, \textit{representation-based fine-tuning} performs well due to the preservation of domain-relevant pre-trained representations [\textbf{Finding 3}].
\end{itemize}
\vspace{-0.5mm}
% Inspired by Finding 1 and Q1, we propose a \textbf{new method, \proj}. We observe that simple post-hoc weight interpolation between pre-trained and fine-tuned model weights (WiSE-FT) performs well for classification tasks but struggles with regression tasks. In contrast, a more complex weight ensemble approach ($L^2$-SP) achieves better performance in regression tasks, though it comes with the cost of increased tuning complexity. \proj combines the strengths of WiSE-FT and $L^2$-SP, providing strong performance for both task types while maintaining the plug-and-play ease of post-hoc interpolation. The success of \proj illustrates how this benchmark can provide valuable insight for fine-tuning strategies for MGFMs. 

Based on the findings, we argue that the first step in selecting or designing an effective fine-tuning strategy is to consider the pre-training strategies. Then after finding the suitable fine-tuning mechanisms, we need to take the type of downstream tasks into account. For instance, weight-based fine-tuning methods generally work the best under self-supervised pre-trained model, while simple post-hoc weight interpolation between pre-trained and fine-tuned model weights (WiSE-FT) performs well for classification tasks but struggles with regression tasks. In contrast, a more complex weight ensemble approach ($L^2$-SP) achieves better performance in regression tasks, though it comes with the cost of increased tuning complexity. Therefore, inspired by the rule, we propose a \textbf{new method, \proj} that achieves strong performance for both regression and classification tasks as a weight-based solution for self-supervised pre-trained model. \proj combines the strengths of WiSE-FT and $L^2$-SP, providing strong performance for both task types while maintaining the plug-and-play ease of post-hoc interpolation. The success of \proj showcases that our benchmark identifies valuable insights in improving fine-tuning strategies given distinct MGFMs.  

\vspace{-2mm}
\section{Finetuning Methods for Evaluation} 
\vspace{-2mm}
%: Pre-training and Fine-tuning for Molecular Graph Foundation Models} 
\label{sec:prelim}
% \pan{Where are the  notations? As a section of preliminary rather than related works, we should write them in the more formal way.}
% \pan{write something to describe what is expected in this section if this is a long section and the content does not follow the tradition of the section caption. }
% \subsection{Graph Pre-training Methods \pan{why still graph?}}
% \label{subsec:pt_method}
% \begin{itemize}
%     \item unsupervised pre-training~\citep{you2020graph, hu2020strategies, hou2022graphmae, xiamole, suresh2021adversarial, wang2022molecular, xia2023systematic}
%     \item supervised pre-training \cite{gasteigergemnet, shoghimolecules, beainitowards}
% \end{itemize}
% \Shikun{trim to make it fit into about 1 page, put more discussion of graph pre-training methods and fine-tuning methods to appendix}
\vspace{-1mm}
In this section, we briefly introduce representative methodologies used in pre-training and fine-tuning for MGFMs.

\textbf{Self-supervised Pre-training} 
% Tremendous research was targeted on exploring the pre-training techniques for GNN, particularly for molecules~\citep{xia2022survey}. 
strategies have been proven to be effective in generating transferable molecular representations for  downstream tasks~\citep{zhao2024survey}.
In a high level, they can be divided into \textit{reconstruction} methods and \textit{contrastive} methods. The generative-based strategies adopt mask-based graph reconstruction by utilizing graph autoencoders~\citep{hou2022graphmae, tan2023s2gae, wang2017mgae, pan2018adversarially}, context predictions~\citep{hu2020strategies, rong2020self} and generative language
model pre-training~\citep{hu2020gpt, zhang2021motif}. On the other hand, contrastive-based methods aim for maximizing
the similarity between perturbed instance pairs~\citep{velivckovic2018deep, suresh2021adversarial, you2020graph, xia2023mole, wang2022molecular, zhu2022unified, you2021graph, qiu2020gcc, li2022let, xu2021self}. Moreover, the advancement of language models has prompted numerous studies to employ multi-modal frameworks. These approaches harness language models to enhance molecular understanding through techniques such as cross-modal contrastive learning and alignment~\citep{su2022molecular, liu2023multi, seidl2023enhancing, liu2023molca}.

In this work, we select \textit{GraphMAE}~\citep{hou2022graphmae} as the representative of the recontruction-based pre-trained model, which focuses on masked feature reconstruction with scaled cosine error that enabled robust training. Regarding the contrastive pre-trained model, we choose \textit{Mole-BERT}~\citep{xia2023mole} that combines the node-level masked atom modeling to predict the masked atom tokens and the graph-level contrastive learning through triplet loss and contrastive loss. Lastly, we choose \textit{MoleculeSTM}~\citep{liu2023multi} as the representative of multi-modal molecule structure-text model that jointly learning molecules’ chemical structures and textual descriptions via a contrastive learning strategy.

% \textbf{Supervised Pre-training}. Recently, with more diverse labeled molecular datasets and models with higher expressive power proposed \pan{not sure if the motivation for supervising pretraining is clear. It should be diversified dataset + diversified tasks}, researchers have started exploring the capability of supervised pre-training with multi-task learning for molecular representations~\citep{gasteigergemnet, shoghimolecules, beainitowards}.
% % \pan{year in reference}. 
% We utilize the datasets from the \textit{Graphium} library~\citep{beainitowards} and obtain the pre-trained model by training on multi-task labeled samples in the supervised manner \pan{is this your pretrained model or follow some established pretrained model? If it is your pretrained model, it is persuasive.}. 

\textbf{Supervised Pre-training.} Recently, to leverage more diverse datasets and tasks, researchers started exploring the ability of supervised pre-training with multi-task learning for molecular representations~\citep{gasteigergemnet, shoghimolecules, beainitowards}. We adopt pre-trained models trained on multi-task labeled samples in a supervised manner from the \textit{Graphium} library~\citep{beainitowards}. In addition to the GNN-based backbone, more expressive architectures like Graph Transformer~\citep{ying2021transformers, rampavsek2022recipe, mialon2021graphit} have been proposed and can be used as the pre-trained backbone with supervised labels, which we adopt \textit{GraphGPS}~\cite{rampavsek2022recipe} as a representative. % and obtain the pre-trained model by training on multi-task labeled samples in the supervised manner.

% \subsection{Fine-tuning Methods}
% \label{subsec:ft_method}
% \shikun{add here}

\begin{figure*}[t]
    % \vspace{-5mm}
     \centering
\includegraphics[trim={0.5cm 0cm 0.5cm 0cm},clip,width=0.75\linewidth]{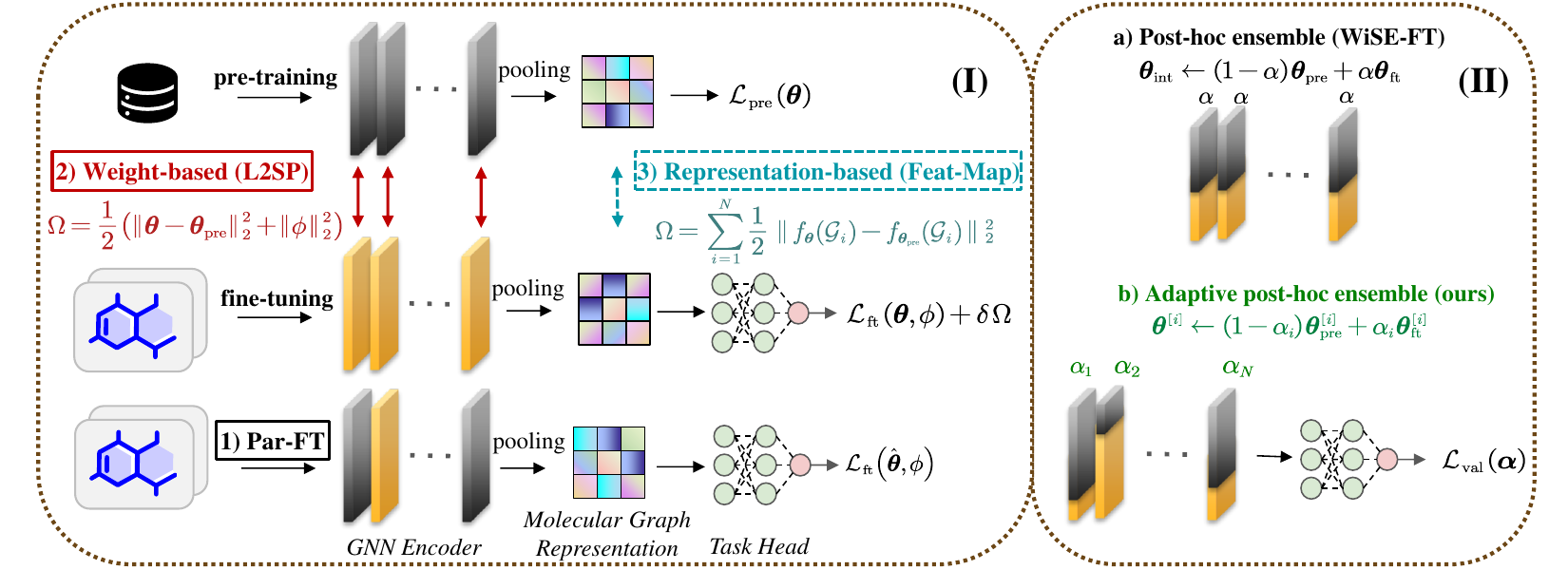}

     \caption{The overall framework of fine-tuning strategies evaluated in our benchmark, \projj, and the proposed novel method, \proj. \textbf{(I)} The GNN encoder is pre-trained on a large database by the objective $\mathcal{L}_{\text{pre}}$, and fine-tuned on the downstream dataset by $\mathcal{L}_\text{ft}$ (\textit{c.f.},  Eq.~\ref{eq:ft}). 1) Partial-FT, 2) Weight-based FT, and 3) Representation-based FT achieve robust fine-tuning by freezing partial pre-trained model weights, regularizing model weights and latent representations, respectively. \textbf{(II)} \proj combines the strength of simple post-hoc weight interpolation with more elaborate weight ensemble, showing the improved performance while maintaining easy usage.}
    \label{fig:pipeline}
    \vspace{-5mm}
\end{figure*}

% After the pre-training on large scale data, we have access to the pre-trained model and the goal for fine-tuning is to adapt this model to the downstream tasks. 
\textbf{Fine-tuning}'s overall goal is to adapt the pre-trained model to downstream applications.
Specifically, given a pre-trained GNN encoder $f_{\mtheta}$ with parameters $\mtheta$ initialized from the pretrained parameters $\mtheta_\text{pre}$, fine-tuning optimizes the encoder $f_{\mtheta}$ and an additional 
prediction head $g_{\mphi}$ with parameters $\mphi$ over downstream molecules  $\{(\dG_i, y_i)\}_{i=1}^N$. The vanilla version, \textbf{full-FT}, optimizes the entire model weights following:
\vspace{-1em}
\begin{align}
    \min_{\{\mtheta, \mphi\}} \sum_{i=1}^N \mathcal{L} (g_{\mphi}\circ f_{\mtheta}(\dG_i), y_i),   \label{eq:ft}
\end{align}
% \vspace{-0.1em}
where $\mtheta$ is initialized as $\mtheta_\text{pre}$ and $\mathcal{L}$ denotes the loss function for prediction tasks. As discussed, there are advanced fine-tuning strategies proposed on top of the full-FT framework. As shown in Fig.~\ref{fig:pipeline}, we group them into three categories based on their mechanisms and benchmark representative methods for each category.
More FT methods that fall into each category or others will be discussed in Appendix~\ref{appendix:related}.

% \derek{Seems that we need to delete from ``It's important...'' to ``...in future work''}
% It's important to note that our selection of fine-tuning methods for benchmarking is primarily based on exploring \textit{how various existing fine-tuning mechanisms perform in molecular FT} \pan{what is molecular FT, a new phrase created?}. Recently, several graph-specific fine-tuning techniques~\citep{huang2024measuring,sun2024fine,li2024adaptergnn} have been proposed. However, they have not exhibited systematic mechanism categorization \pan{I do not understand this sentence, do you want to say that these methods do not belong to one category? I do not think we even need to mention as they are very recent works. We can put them in the appendix.} and we plan to include in future work.

$\bullet$ \textbf{Partial FT}
strategies only optimizes partial weights of the pre-trained model, \textit{i.e.}, a subset of weights within $\{\mtheta, \mphi\}$ will be updated following the same objective as Eq.~\ref{eq:ft}. 
% \pan{Eq.~\ref{eq:ft}}.
% \vspace{-1mm}
% \begin{itemize}[leftmargin=*, nolistsep]
    % \item[$\dagger$] 
    \textit{Linear Probing (LP)} only trains the additional prediction head $g$ during the FT. 
    % which optimizes for $\mphi$ as $\min_{\mphi} \sum_{i=1}^N L(g\circ f(\dG_i), y_i)$ with fixed $\mtheta_0$ for the GNN encoder. 
    % \item[$\dagger$] 
    \textit{Surgical FT}~\citep{leesurgical} 
    % \pan{which year for this work? Also please check other references that miss the year.}  
    updates only partial layers within the encoder. For instance, we can update the weights for $k$-th layer of the GNN encoder as $\min_{\{[\mtheta]_k, \mphi\}} \sum_{i=1}^N \mathcal{L}(g_{\mphi}\circ f_{\mtheta}(\dG_i), y_i)$, where $k$ is the hyperparameter that can be tuned. 
    % \item[$\dagger$] 
    \textit{LP-FT}~\citep{kumar2022fine} aims to address the issue of pre-trained feature distortion during the full-FT process.   % when fine-tuning both the encoder and randomly initialized prediction head simultaneously.
    % identified that when we fine-tune the randomly initialized prediction head and the encoder simultaneously, the pre-trained features can be distorted. 
    It first performs the LP step to the prediction head $g_{\mphi}$ while keeping the encoder $f_{\mtheta}$ with fixed pre-trained parameters $\mtheta_{\text{pre}}$, followed by applying full-FT with the tuned prediction head. 
% \end{itemize}
% \vspace{-1mm}

$\bullet$ \textbf{Weight-based FT} strategies mainly update the entire model weights through combining pre-trained model weights and fine-tuned model weights.
% \vspace{-1mm}
% \begin{itemize}[leftmargin=*, nolistsep]
% \setlength{\itemsep}{0pt}
% \setlength{\parsep}{0pt}
% \setlength{\parskip}{0pt}
    % \item[$\dagger$] 
    \textit{WiSE-FT}~\citep{wortsman2022robust} linearly interpolates between pre-training parameters $\mtheta_{\text{pre}}$ and fine-tuning parameters $\mtheta_{\text{ft}}$ using a mixing coefficient $\alpha$, to get the interpolated GNN $f_{\mtheta_{\text{int}}}$ with weights $\mtheta_{\text{int}} = (1-\alpha)\cdot \mtheta_{\text{pre}} + \alpha \cdot \mtheta_{\text{ft}}$. We first perform full-FT to obtain the adapted encoder $f_{\mtheta_{\text{ft}}}$ and classifier $g_{\mphi}$, then apply post-hoc weight ensembling to get $f_{\mtheta_{\text{int}}}$, with final predictions given by $g_{\mphi}\circ f_{\mtheta_{\text{int}}}(\dG_i)$. $\alpha$, as a hyperparameter, controls the weight ensemble.
    % \item[$\dagger$] 
    \textit{$L^2$-SP}~\citep{xuhong2018explicit} regularizes the fine-tuning model weights $\mtheta$ closer to the pre-trained weights $\mtheta_{\text{pre}}$ by
    $\Omega(\mtheta, \mphi) = \frac{\delta}{2}\lVert\mtheta - \mtheta_{\text{pre}}\rVert^2_2$. We optimize for $\mtheta$ and $\mphi$ by combining the prediction loss from Eq.~\ref{eq:ft} and $\Omega(\mtheta, \mphi)$ with tunable trade-off coefficient $\delta$. 
% \end{itemize}
\vspace{-1mm}

$\bullet$ \textbf{Representation-based FT} methods
mainly regulate the latent representation space during FT.
% \vspace{-1mm}
% \begin{itemize}[leftmargin=*, nolistsep]
    % \item[$\dagger$] 
    \textit{Feature-map}~\citep{lidelta} adds distance regularization between the latent representations of pre-trained and fine-tuned models to the Full-FT loss. The regularization is defined as $\Omega(\mtheta) = \delta\sum_{i=1}^N \frac{1}{2}\lVert f_{\mtheta}(\dG_i) - f_{\mtheta_{\text{pre}}}(\dG_i)\rVert^2_2$, where $\delta$ controls the regularization strength.
    % \item[$\dagger$] 
    \textit{BSS}~\citep{chen2019catastrophic} aims at resolving the negative transfer issue through eliminating the spectral components corresponding to small singular values that are less transferable. The regularization is done as $\Omega(\mF) = \delta\sum_{i=1}^k \sigma^2_{-i}$, where $\mF = [f_{\mtheta}(\dG_0), \cdots, f_{\mtheta}(\dG_b)]$ is the feature matrix of a batch of graphs and $\sigma_{-i}$ are the $i$-th smallest singular values obtained from the SVD of $\mF$. We can tune $k$ and $\delta$ to determine the number of singular values to penalize and the degree of penalty. 

\vspace{-1mm}
\section{Experimental Settings in the Benchmark}
\vspace{-1mm}
\label{sec:exp_setting}

% \subsection{Datasets}
% \label{sec:exp_setting}
% \begin{itemize}
%     \item Introduce the datasets and their physical meaning.~\cite{wu2018moleculenet, liupre}
%     \item Include the table consisting of their size and evaluation metrics
%     \item In the appendix, having some plots indicating the overlapping molecules with the downstream datasets and the pretraining datasets
% \end{itemize}
% \derek{You may polish it later}.
In this section, we briefly introduce the experimental settings in this work. More detailed experimental settings can be found in Appendix~\ref{appendix:experiments}.

\textbf{Foundation Models}.
For self-supervised pre-training, we adopt three open-source pre-trained checkpoints: \textit{Mole-BERT}, \textit{GraphMAE}, and \textit{MoleculeSTM}. For supervised
pre-training, we use models from the \textit{Graphium}~\citep{beainitowards} library, which get pre-trained on the Toymix
and Largemix datasets provided in this library. To differentiate between them, we refer to these models as \textit{Graphium-Toy} and \textit{Graphium-Large}. For larger graph transformer based model, we adopt the pre-trained checkpoint of \textit{GraphGPS}~\cite{rampavsek2022recipe} pre-trained on the PCQM4MV2~\cite{hu2ogb}. 
For details of datasets used in pre-training are in Appendix~\ref{appendix:pt_data}. Furthermore, we include the traditional baseline XGBoost~\cite{chen2016xgboost} for Fewshot scenarios to better compare with the foundation model in Appendix~\ref{appendix:traditional}.

\textbf{Downstream Datasets}. We use 8 classification and 4 regression datasets for downstream task evaluation.
% , and categorize them into the following three types based on different aspects of molecular properties. 
Detailed statistics and references for these tasks are in Appendix~\ref{appendix:stat}.

$\dagger$ \textit{Classification.} The BBBP dataset
measures if a molecule will penetrate blood-brain barrier. The Tox21,
ToxCast, and ClinTox datasets are related to   toxicity qualitative measurements. The Sider dataset stores  qualitative results of different types of adverse drug reactions.  The MUV dataset is specifically designed for validation of virtual
screening techniques. The HIV dataset provides qualitative activity results of the molecular ability to inhibit HIV replication.  The BACE dataset contains qualitative binding results for a set of inhibitors of human $\beta$-secretase 1 (BACE-1).

$\dagger$ \textit{Regression.}  Esol is a dataset which measures aqueous solubility
of molecules. 
The Lipo dataset measures the
octanol-water partition coefficient. 
Cep is a subset of the Havard Clean Energy
Project (CEP), which estimates the organic photovoltaic efficiency. Malaria measures the drug efficacy against the parasite
that causes malaria.

\textbf{Dataset Splits}. For each downstream dataset, we experiment with \textit{random, scaffold}, and \textit{size} splits to create the Train/Val/Test subsets. Specifically, the random splitting shuffles the data, maintaining the Train/Val/Test sets as in-distribution (ID). The other two splitting methods simulate out-of-distribution (OOD) challenges in real-world applications. For scaffold splitting, we follow prior works~\citep{ramsundar2019deep}, ensuring structural differences in molecular scaffolds across splits. Size splitting, following \citet{zougess}, arranges molecules in ascending order by size, evaluating model generalization across different molecule sizes.

\textbf{Size of fine-tuning samples}. 
% \pan{Emphasize the motivation for few shot again, give more detailed descriptions with references}
In practice, molecular property prediction tasks can have very limited experimentally-validated data, e.g., with less than 100 samples~\citep{wijaya2024two}. Thus, we consider both \textit{Non-Fewshot} and \textit{Fewshot} settings to better simulate the label scarcity issue. In the Non-Fewshot setting, we use all available samples from the splitted train set. In the Fewshot settings, we sample subsets of 50, 100, and 500 molecules from the Train set for fine-tuning, while keeping the Val/Test sets unchanged to ensure a fair comparison. Note that we exclude MUV, Tox21, and ToxCast datasets for the Fewshot settings, as we cannot \textit{randomly} select training samples while ensuring that all tasks have a specified number of labels simultaneously, due to the severe label scarcity issues in these datasets.

\textbf{Evaluation Metrics}. We use AUC to evaluate the performance for classification datasets and RMSE for regression datasets. We report the model performance over 5 random seeds and the test performance are reported based on the best validation performance. The AVG, AVG-F, AVG-R denote the average metrics, average metrics without max and min values, and average rank over all the datasets for each evaluated method, respectively.
\begin{table}[htbp]
\vspace{-1em}
\caption{A summary of evaluated pre-trained models and their corresponding result tables for reference. ``\textsc{Clf}'' and ``\textsc{Rgs}'' represent classification and regression tasks, respectively, while ``\textsc{Non}'' and ``\textsc{Few}'' denote Non-Fewshot and Fewshot settings.
% \shikun{merge GraphMAE and GraphGPS together and add reference here. Will later update the table, currently it is a placeholder with the same format as other set of tables}
}
\label{tab:model-reference}
\centering
\resizebox{0.7\columnwidth}{!}{%
\begin{tabular}{cccccc}
\toprule[2pt]
\multirow{2}{*}{Objectives} &
  \multirow{2}{*}{Models} &
  \multicolumn{4}{c}{Reference Tables of Experimental Results} \\ \cline{3-6} 
 &
   &
  \textsc{Clf-Non} &
  \textsc{Clf-Few} &
  \textsc{Rgs-Non} &
  \textsc{Rgs-Few} \\ \midrule
\multirow{3}{*}{Self-Supervised} & Mole-BERT      &\ref{tab-main1}  &\ref{tab-clf-fewshot-bert-sup}  &\ref{tab-main2}  &\ref{tab-regs-fewshot-bert-sup}  \\
                                 & GraphMAE       &\ref{tab-mae-gps-clf-non}  &\ref{tab-mae-gps-clf-few}  &\ref{tab-mae-gps-rgs-non}  &\ref{tab-mae-gps-rgs-few}  \\
                                 & MoleculeSTM    &\ref{tab-stm-large-clf-non}  &\ref{tab-stm-large-clf-few}  &\ref{tab-stm-large-rgs-non}  &\ref{tab-stm-large-rgs-few}  \\\midrule
\multirow{2}{*}{Supervised}      & Graphium-Toy   &\ref{tab-main1}  &\ref{tab-clf-fewshot-bert-sup}  &\ref{tab-main2}  &\ref{tab-regs-fewshot-bert-sup}  \\
                                 & Graphium-Large &\ref{tab-stm-large-clf-non}  &\ref{tab-stm-large-clf-few}  &\ref{tab-stm-large-rgs-non}  &\ref{tab-stm-large-rgs-few}  \\
                                 & GraphGPS   &\ref{tab-mae-gps-clf-non}  &\ref{tab-mae-gps-clf-few}  &\ref{tab-mae-gps-rgs-non}  &\ref{tab-mae-gps-rgs-few}  \\
                                 \bottomrule[2pt]
\end{tabular}%
}
\vspace{-2em}
\end{table}
\section{Results and Analysis}

\label{sec:result}
% \Shikun{In main text, put the results of Mole-BERT and supervised pretrained model for classification and regression datasets. The few-shot learning keeps the 50 and 500. GraphMAE and few-shot 500 put into appendix. Total 7 tables, but can put the few-shot ones into plot maybe. If put 7 tables, it will take at least 1.5-2 pages}
\begin{table*}[t]
\vspace{-0.1cm}
\caption{Robust fine-tuning performance on 8 \hlb{\textbf{Classification}}{30}
datasets (AUC metrics) in the \hlr{\textbf{Non-Fewshot}}{30} setting, evaluated across 3 dataset splits (\textsc{Random, Scaffold, Size}), over \hlg{\textsc{\textbf{Mole-BERT}}}{30} and \hlg{\textsc{\textbf{Graphium-Toy}}}{30} models. \textsc{Avg, Avg-F, Avg-R} denote the average AUC, average AUC without max and min values, and average rank over all the  datasets for each method, respectively. Standard deviations across five replicates are shown. We \textbf{bold} and \underline{underline} the best and second-best performances in each scenario. 
}
\vspace{-1mm}
\label{tab-main1}
\begin{center}
\begin{adjustbox}{width = 1.0\textwidth}
\begin{small}
\begin{sc}
\begin{tabular}{ccccccccccccc}

\toprule[2pt]

 Split & Methods  &   ClinTox &  BBBP  &     Bace    & HIV     & MUV & Sider &  Tox21  &   ToxCast & Avg & Avg-F & Avg-R\\ \midrule
\multicolumn{13}{c}{{\textbf{Self-supervised Pre-training (Mole-BERT)}}}  \\ \midrule
\multirow{8}{*}{Scaffold}

&  Full-FT   & $\mathbf{ 77.70 \pm 1.50 }$ & $\underline{ 67.93 \pm 3.85 }$ &  $80.12 \pm 1.07$  &  $77.00 \pm 0.80$  &  $80.50 \pm 0.81$  &  $63.47 \pm 0.77$  &  $78.31 \pm 0.28$  &  $65.18 \pm 0.35$  & $73.78$ & $74.37$ & $3.75$ \\
 &  LP     &  $66.49 \pm 0.46$  &  $65.42 \pm 0.26$  &  $78.70 \pm 0.27$  & $\underline{ 77.15 \pm 0.12 }$ &  $79.27 \pm 0.48$  &  $62.01 \pm 0.60$  &  $78.12 \pm 0.15$  &  $64.75 \pm 0.17$ & $71.49$ & $71.77$ & $6.12$ \\
 &  Surgical-FT   &  $68.19 \pm 1.58$  &  $67.70 \pm 0.54$  & $\mathbf{ 84.24 \pm 0.37 }$ &  $76.65 \pm 0.46$  & $\underline{ 81.60 \pm 1.02 }$ & $\underline{ 64.61 \pm 0.31 }$ & $\underline{ 78.34 \pm 0.10 }$ &  $65.21 \pm 0.28$ & $73.32$ & $72.95$ & $3.62$ \\
 &  LP-FT   &  $70.35 \pm 0.99$  & $\mathbf{ 68.30 \pm 0.65 }$ &  $81.90 \pm 0.70$  &  $76.69 \pm 0.40$  &  $77.65 \pm 1.15$  &  $63.38 \pm 0.67$  &  $77.60 \pm 0.19$  &  $65.32 \pm 0.24$  & $72.65$ & $72.65$ & $4.88$ \\
 &  WiSE-FT  &  $73.59 \pm 3.74$  &  $66.52 \pm 3.29$  & $\underline{ 82.73 \pm 0.87 }$ & $\mathbf{ 77.21 \pm 0.69 }$ & $\mathbf{ 81.92 \pm 0.94 }$ &  $63.62 \pm 0.62$  &  $78.05 \pm 0.28$  & $\underline{ 65.41 \pm 0.25}$ & $73.63$ & $73.78$ & $3.38$ \\
 &  $L^2$-SP  &  $73.95 \pm 1.86$  &  $67.86 \pm 1.68$  &  $81.47 \pm 0.80$  &  $76.63 \pm 0.56$  &  $77.21 \pm 0.72$  & $\mathbf{ 65.27 \pm 0.45 }$ & $\mathbf{ 78.66 \pm 0.17 }$ &  $63.55 \pm 0.16$ & $73.07$ & $73.26$ & $4.50$ \\
 &  Feature-map  &  $70.65 \pm 0.76$  &  $65.41 \pm 2.37$  &  $73.44 \pm 0.23$  &  $76.71 \pm 0.26$  &  $80.03 \pm 0.47$  &  $64.35 \pm 0.17$  &  $76.61 \pm 0.39$  & $\mathbf{ 65.77 \pm 0.15}$ & $71.62$ & $71.43$ & $5.25$ \\
 &  BSS  & $\underline{ 76.07 \pm 3.23 }$ &  $67.47 \pm 3.80$  &  $80.98 \pm 1.27$  &  $77.12 \pm 0.86$  &  $77.35 \pm 1.76$  &  $63.88 \pm 0.80$  &  $78.19 \pm 0.40$  &  $65.00 \pm 0.27$  & $73.26$ & $73.53$ & $4.50$ \\
\midrule
\multirow{8}{*}{Size}

 &  Full-FT   &  $72.78 \pm 1.74$  &  $87.37 \pm 0.82$  &  $66.00 \pm 1.99$  &  $79.85 \pm 0.64$  &  $77.02 \pm 2.15$  &  $52.46 \pm 0.29$  &  $75.74 \pm 0.48$  &  $63.13 \pm 0.32$ & $71.79$ & $72.42$ & $4.88$ \\
 &  LP     & $\mathbf{ 76.07 \pm 0.32 }$ &  $82.73 \pm 0.76$  &  $47.18 \pm 0.45$  &  $78.16 \pm 0.24$  &  $78.52 \pm 1.60$  &  $51.25 \pm 0.22$  &  $74.92 \pm 0.22$  &  $63.33 \pm 0.20$ & $69.02$ & $70.37$ & $6.00$ \\
 &  Surgical-FT   &  $73.55 \pm 0.81$  & $\mathbf{ 88.82 \pm 0.53 }$ &  $66.43 \pm 0.88$  &  $79.30 \pm 0.87$  & $\mathbf{ 80.52 \pm 1.47 }$ &  $51.87 \pm 0.23$  &  $76.32 \pm 0.16$  & $\mathbf{ 64.51 \pm 0.20}$ & $72.66$ & $73.44$ & $3.50$ \\
 &  LP-FT   & $\underline{ 75.32 \pm 0.93 }$ &  $83.42 \pm 1.67$  &  $64.84 \pm 1.38$  &  $79.10 \pm 1.14$  & $\underline{ 79.38 \pm 1.86 }$ & $\underline{ 52.82 \pm 0.32 }$ &  $76.40 \pm 0.28$  &  $63.37 \pm 0.29$ & $71.83$ & $73.07$ & $3.88$ \\
 &  WiSE-FT  &  $73.45 \pm 1.08$  & $\underline{ 87.79 \pm 1.53 }$ & $\underline{ 66.58 \pm 1.11 }$ & $\underline{ 79.89 \pm 1.75 }$ &  $78.41 \pm 1.88$  &  $52.46 \pm 0.49$  & $\underline{ 76.46 \pm 0.46 }$ & $\underline{ 63.53 \pm 0.65 }$ & $72.32$ & $73.05$ & $3.00$ \\
 &  $L^2$-SP  &  $73.97 \pm 0.88$  &  $87.15 \pm 0.68$  &  $64.58 \pm 1.93$  & $\mathbf{ 80.05 \pm 0.53 }$ &  $74.83 \pm 1.06$  &  $52.37 \pm 0.22$  &  $75.84 \pm 0.28$  &  $60.63 \pm 0.36$ & $71.18$ & $71.65$ & $5.12$ \\
 &  Feature-map  &  $74.61 \pm 0.53$  &  $85.42 \pm 0.31$  &  $51.23 \pm 0.46$  &  $76.39 \pm 0.91$  &  $75.20 \pm 2.27$  &  $51.96 \pm 0.26$  & $\mathbf{ 76.81 \pm 0.25 }$ &  $63.42 \pm 0.76$  & $69.38$ & $69.73$ & $5.00$ \\
 &  BSS  &  $73.99 \pm 0.77$  &  $86.84 \pm 1.00$  & $\mathbf{ 66.97 \pm 1.58 }$ &  $79.64 \pm 1.44$  &  $73.42 \pm 2.60$  & $\mathbf{ 53.50 \pm 0.66 }$ &  $75.69 \pm 0.26$  &  $62.41 \pm 0.69$ & $71.56$ & $72.02$ & $4.62$ \\

\midrule
\multicolumn{13}{c}{{\textbf{Supervised Pre-training (Graphium-Toy)}}}  \\ \midrule
\multirow{8}{*}{Scaffold}

% & Full-FT  & $81.27\pm 3.88$ & $69.17\pm 1.32$ & $79.75\pm 1.07$ & $76.42\pm 0.72$ & $76.84\pm 1.80$ & $63.63\pm 0.06$ & $78.12\pm 0.46$ & $66.37\pm 0.26$\\
% & LP    & $80.48\pm 0.00$ & $66.90\pm 0.00$ & $80.44\pm 0.00$ & $75.83\pm 0.00$ & $73.35\pm 0.00$ & $62.03\pm 0.00$ & $79.02\pm 0.00$ & $66.09\pm 0.00$\\
% & Surgical-FT  & $86.17\pm $ & $73.71\pm $ & $84.16\pm $ & $77.47\pm $ & $78.87\pm $ & $64.02\pm $ & $78.23\pm $ & $67.34\pm $\\
% & LP-FT  & $83.67\pm 3.53$ & $69.98\pm 0.83$ & $79.28\pm 0.32$ & $76.17\pm 2.01$ & $77.82\pm 1.15$ & $61.20\pm 0.00$ & $76.94\pm 0.00$ & $66.28\pm 0.00$ \\
% & WiSE-FT & $85.40\pm $ & $71.89\pm $ & $78.13\pm $ & $76.69\pm $ & $74.37\pm $ & $63.58\pm $ & $77.98\pm $ & $66.48\pm $\\
% & $L^2$-SP & $76.83\pm $ & $67.35\pm $ & $78.17\pm $ & $73.69\pm $ & $62.35\pm $ & $62.21\pm $ & $76.27\pm $ & $62.75\pm $
% \\
% & Feature-map & $90.13 \pm $ & $70.99 \pm $ & $83.17 \pm $ & $73.61 \pm $ & $78.74 \pm $ & $62.12 \pm $ & $79.99 \pm $ & $65.03 \pm $
% \\
% & BSS & $79.99 \pm $ & $67.10 \pm $ & $78.12 \pm $ & $72.50 \pm $ & $61.20 \pm $ & $61.13 \pm $ & $76.69 \pm $ & $65.45 \pm $
% \\

&  Full-FT   &  $81.27\pm 3.88$  &  $69.17\pm 1.32$  &  $79.75\pm 1.07$  &  $76.42\pm 0.72$  &  $76.84\pm 1.80$  & $\underline{ 63.63\pm 0.06 }$ &  $78.12\pm 0.46$  &  $66.37\pm 0.26$ & $73.95$ & $74.45$ & $3.75$ \\
 &  LP     &  $80.48\pm 0.00$  &  $66.90\pm 0.00$  &  $80.44\pm 0.00$  &  $75.83\pm 0.00$  &  $73.35\pm 0.00$  &  $62.03\pm 0.00$  & $\underline{ 79.02\pm 0.00 }$ &  $66.09\pm 0.00$ & $73.02$ & $73.61$ & $5.12$ \\
 &  Surgical-FT   & $\underline{ 86.17\pm0.00  }$ & $\mathbf{ 73.71\pm0.00  }$ & $\mathbf{ 84.16\pm0.00  }$ & $\mathbf{ 77.47\pm0.00  }$ & $\mathbf{ 78.87\pm0.00  }$ & $\mathbf{ 64.02\pm0.00  }$ &  $78.23\pm0.00 $  & $\mathbf{ 67.34\pm0.00 }$ & $76.25$ & $76.63$ & $1.38$ \\
 &  LP-FT   &  $83.67\pm 3.53$  &  $69.98\pm 0.83$  &  $79.28\pm 0.32$  &  $76.17\pm 2.01$  &  $77.82\pm 1.15$  &  $61.20\pm 0.00$  &  $76.94\pm 0.00$  &  $66.28\pm 0.00$  & $73.92$ & $74.41$ & $4.62$ \\
 &  WiSE-FT  &  $85.40\pm1.61 $  & $\underline{ 71.89\pm1.79  }$ &  $78.13\pm2.92 $  & $\underline{ 76.69\pm1.76  }$ &  $74.37\pm1.79 $  &  $63.58\pm0.00 $  &  $77.98\pm0.33 $  & $\underline{ 66.48\pm0.43 }$ & $74.31$ & $74.26$ & $3.62$ \\
 &  $L^2$-SP  &  $76.83\pm8.87 $  &  $67.35\pm 0.82$  &  $78.17\pm0.02 $  &  $73.69\pm0.03 $  &  $62.35\pm0.15 $  &  $62.21\pm0.45 $  &  $76.27\pm0.32 $  &  $62.75\pm0.88 $ & $69.95$ & $69.87$ & $6.62$ \\
 &  Feature-map  & $\mathbf{ 90.13 \pm2.12  }$ &  $70.99 \pm 0.27$  & $\underline{ 83.17 \pm0.49  }$ &  $73.61 \pm0.03 $  & $\underline{ 78.74 \pm0.76  }$ &  $62.12 \pm0.02 $  & $\mathbf{ 79.99 \pm 0.12 }$ &  $65.03 \pm 0.08$ & $75.47$ & $75.25$ & $3.50$ \\
 &  BSS  &  $79.99 \pm5.89 $  &  $67.10 \pm0.93 $  &  $78.12 \pm2.32 $  &  $72.50 \pm0.51 $  &  $61.20 \pm 0.08$  &  $61.13 \pm0.95 $  &  $76.69 \pm 0.64$  &  $65.45 \pm0.89 $ & $70.27$ & $70.18$ & $7.38$ \\

\midrule
\multirow{8}{*}{Size}
% & Full-FT  & $85.96\pm 4.28$ & $87.62\pm 0.90$ & $67.41\pm 2.44$ & $81.47\pm 1.94$ & $72.03\pm 2.55$ & $54.72\pm 0.01$ & $69.71\pm 0.37$ & $61.31\pm 0.37$\\
% & LP    & $81.84\pm 0.02$ & $78.09\pm 0.00$ & $58.08\pm 0.01$ & $77.48\pm 0.00$ & $69.46\pm 0.00$ & $53.59\pm 0.00$ & $73.65\pm 0.00$ & $61.25\pm 0.00$\\
% & Surgical-FT  & $86.59\pm $ & $89.07\pm $ & $70.94\pm $ & $82.50\pm $ & $74.47\pm $ & $56.24\pm $ & $72.30\pm $ & $62.74\pm $\\
% & LP-FT  & $86.78\pm 2.69$ & $88.02\pm 1.50$ & $63.72\pm 1.85$ & $82.57\pm 0.46$ & $73.51\pm 1.77$ & $52.40\pm 0.00$  & $68.23\pm 0.87$ & $60.85\pm 0.00$\\
% & WiSE-FT & $82.44\pm $ & $87.76\pm $ & $72.89\pm $ & $81.37\pm $ & $73.67\pm $ & $55.87\pm $ & $68.85\pm $ & $60.61\pm $\\
% & $L^2$-SP & $71.03\pm $ & $81.32\pm $ & $68.82\pm $ & $70.66\pm $ & $64.69\pm $ & $52.08\pm $ & $70.91\pm $ & $56.50\pm $\\
% & Feature-map & $82.48\pm $ & $87.70\pm $ & $69.56\pm $ & $67.23\pm $ & $71.49\pm $ & $54.43\pm $ & $74.12\pm $ & $58.73\pm $\\
% & BSS & $72.42\pm $ & $82.92\pm $ & $62.76\pm $ & $72.81\pm $ & $65.79\pm $ & $52.89\pm $ & $71.91\pm $ & $57.79\pm $\\

&  Full-FT   &  $85.96\pm 4.28$  &  $87.62\pm 0.90$  &  $67.41\pm 2.44$  &  $81.47\pm 1.94$  &  $72.03\pm 2.55$  &  $54.72\pm 0.01$  &  $69.71\pm 0.37$  & $\underline{ 61.31\pm 0.37}$ & $72.53$ & $72.98$ & $3.88$ \\
 &  LP     &  $81.84\pm 0.02$  &  $78.09\pm 0.00$  &  $58.08\pm 0.01$  &  $77.48\pm 0.00$  &  $69.46\pm 0.00$  &  $53.59\pm 0.00$  & $\underline{ 73.65\pm 0.00 }$ &  $61.25\pm 0.00$ & $69.18$ & $69.67$ & $5.38$ \\
 &  Surgical-FT   & $\underline{ 86.59\pm 0.01 }$ & $\mathbf{ 89.07\pm 0.00 }$ & $\underline{ 70.94\pm0.01  }$ & $\underline{ 82.50\pm 0.00 }$ & $\mathbf{ 74.47\pm 0.00 }$ & $\mathbf{ 56.24\pm0.00  }$ &  $72.30\pm 0.00$  & $\mathbf{ 62.74\pm0.00 }$ & $74.36$ & $74.92$ & $1.62$ \\
 &  LP-FT   & $\mathbf{ 86.78\pm 2.69 }$ & $\underline{ 88.02\pm 1.50 }$ &  $63.72\pm 1.85$  & $\mathbf{ 82.57\pm 0.46 }$ &  $73.51\pm 1.77$  &  $52.40\pm 0.00$   &  $68.23\pm 0.87$  &  $60.85\pm 0.00$ & $72.01$ & $72.61$ & $4.00$ \\
 &  WiSE-FT  &  $82.44\pm3.02 $  &  $87.76\pm0.5 $  & $\mathbf{ 72.89\pm0.66  }$ &  $81.37\pm1.07 $  & $\underline{ 73.67\pm3.44  }$ & $\underline{ 55.87\pm0.01  }$ &  $68.85\pm0.84 $  &  $60.61\pm0.53 $ & $72.93$ & $73.31$ & $3.62$ \\
 &  $L^2$-SP  &  $71.03\pm3.67 $  &  $81.32\pm1.51 $  &  $68.82\pm0.06 $  &  $70.66\pm0.00 $  &  $64.69\pm0.32 $  &  $52.08\pm0.84 $  &  $70.91\pm0.34 $  &  $56.50\pm0.01 $ & $67.00$ & $67.10$ & $6.88$ \\
 &  Feature-map  &  $82.48\pm3.25 $  &  $87.70\pm0.64 $  &  $69.56\pm0.20 $  &  $67.23\pm1.93 $  &  $71.49\pm0.13 $  &  $54.43\pm0.03 $  & $\mathbf{ 74.12\pm0.09  }$ &  $58.73\pm0.04 $ & $70.72$ & $70.60$ & $4.38$ \\
 &  BSS  &  $72.42\pm0.03 $  &  $82.92\pm1.60 $  &  $62.76\pm4.23 $  &  $72.81\pm0.66 $  &  $65.79\pm 5.31$  &  $52.89\pm1.12 $  &  $71.91\pm0.44 $  &  $57.79\pm1.80 $ & $67.41$ & $67.25$ & $6.25$ \\
 \bottomrule[2pt]
\label{table:nonfew_cls_molebert}
\end{tabular}
\end{sc}
\end{small}
\end{adjustbox}
\end{center}
\vspace{-0.8cm}
\end{table*}

\begin{table*}[t]
% \vspace{-2mm}
\caption{Robust fine-tuning performance on 4 \hlb{\textbf{Regression}}{30}
datasets (RMSE metrics) in the \hlr{\textbf{Non-Fewshot}}{30} setting, evaluated across 3 dataset splits (\textsc{Random, Scaffold, Size}), over \hlg{\textsc{\textbf{Mole-BERT}}}{30} and \hlg{\textsc{\textbf{Graphium-Toy}}}{30} models. \textsc{Avg-R,Avg-R}$^*$ denote the average rank and the rank based on the average normalized performance over all the datasets for each method, respectively. Standard deviations across five replicates are shown. We \textbf{bold} and \underline{underline} the best and second-best performances in each scenario.}
% \pan{it seems that supervised PT is generally better, especially in few shot setting. I think this is a very important knowledge, which should be put in the intro. This means that masked or contrastive self-supervised PT is too generic, which works worse than domain-specific pretraining tasks. We should argue for using diversified domain-related supervised tasks for pretraining.}
\vspace{-3mm}\label{tab-main2}
\begin{center}
\begin{adjustbox}{width = 1\textwidth}
\begin{small}
\begin{sc}
\begin{tabular}{cccccccccccccc}
\toprule[2pt]
\multirow{2}{*}{Split} &
  \multirow{2}{*}{Methods} &
  \multicolumn{6}{c}{Self-supervised Pre-training (Mole-BERT)} &
  \multicolumn{6}{c}{Supervised Pre-training (Graphium-Toy)} \\ \cmidrule(lr){3-14}
 &
   &
  Esol &
  Lipo &
  Malaria &
  Cep &
  Avg-R &
  \multicolumn{1}{c|}{Avg-R$^*$} &
  Esol &
  Lipo &
  Malaria &
  Cep &
  Avg-R &
  \multicolumn{1}{c}{Avg-R$^*$} \\ \midrule
\multirow{8}{*}{Scaffold} &
  Full-FT &
  $1.126 \pm 0.014$ &
  $\underline{ 0.728 \pm 0.011 }$ &
  $1.152 \pm 0.015$ &
  $\mathbf{ 1.377 \pm 0.015 }$ &
  $3.75$ &
  \multicolumn{1}{c|}{5} &
  $0.911 \pm 0.041$ &
  $0.709 \pm 0.009$ &
  $1.110 \pm 0.009$ &
  $1.419 \pm 0.014$ &
  $4.00$ &
  4 \\
 &
  LP &
  $1.614 \pm 0.010$ &
  $0.870 \pm 0.003$ &
  $1.110 \pm 0.002$ &
  $2.006 \pm 0.002$ &
  $7.00$ &
  \multicolumn{1}{c|}{8} &
  $0.973 \pm 0.000$ &
  $0.881 \pm 0.000$ &
  $1.105 \pm 0.000$ &
  $1.826 \pm 0.000$ &
  $6.75$ &
  8 \\
 &
  Surgical-FT &
  $1.166 \pm 0.017$ &
  $0.783 \pm 0.003$ &
  $1.120 \pm 0.014$ &
  $1.601 \pm 0.006$ &
  $5.25$ &
  \multicolumn{1}{c|}{6} &
  $\underline{ 0.892 \pm0.000 }$ &
  $0.709 \pm0.000 $ &
  $1.105 \pm0.000 $ &
  $1.419 \pm0.000 $ &
  $3.50$ &
  2 \\
 &
  LP-FT &
  $\mathbf{ 1.070 \pm 0.021 }$ &
  $0.730 \pm 0.002$ &
  $1.144 \pm 0.022$ &
  $1.397 \pm 0.013$ &
  $3.50$ &
  \multicolumn{1}{c|}{4} &
  $0.922 \pm 0.004$ &
  $0.735 \pm 0.019$ &
  $\underline{ 1.080 \pm 0.005 }$ &
  $\mathbf{ 1.368 \pm 0.037}$ &
  $4.00$ &
  3 \\
 &
  WiSE-FT &
  $1.264 \pm 0.055$ &
  $0.768 \pm 0.010$ &
  $\mathbf{ 1.072 \pm 0.001 }$ &
  $1.470 \pm 0.029$ &
  $4.00$ &
  \multicolumn{1}{c|}{2} &
  $\mathbf{ 0.888 \pm0.014 }$ &
  $\underline{ 0.708 \pm0.008 }$ &
  $1.128 \pm0.021 $ &
  $1.490 \pm0.024 $ &
  $3.75$ &
  6 \\
 &
  $L^2$-SP &
  $\underline{ 1.099 \pm 0.030 }$ &
  $0.742 \pm 0.008$ &
  $1.101 \pm 0.001$ &
  $1.631 \pm 0.006$ &
  $3.75$ &
  \multicolumn{1}{c|}{3} &
  $0.948 \pm0.022 $ &
  $0.729 \pm0.015 $ &
  $1.141 \pm 0.015$ &
  $1.606 \pm0.013 $ &
  $7.00$ &
  7 \\
 &
  Feature-map &
  $1.403 \pm 0.012$ &
  $0.842 \pm 0.004$ &
  $\underline{ 1.083 \pm 0.002 }$ &
  $1.787 \pm 0.003$ &
  $5.75$ &
  \multicolumn{1}{c|}{7} &
  $0.895 \pm0.016 $ &
  $\mathbf{ 0.688 \pm 0.018 }$ &
  $\mathbf{ 1.074 \pm 0.000}$ &
  $1.472 \pm 0.010$ &
  $2.50$ &
  1 \\
 &
  BSS &
  $1.110 \pm 0.022$ &
  $\mathbf{ 0.726 \pm 0.004 }$ &
  $1.125 \pm 0.018$ &
  $\underline{ 1.385 \pm 0.018 }$ &
  $3.00$ &
  \multicolumn{1}{c|}{1} &
  $0.896 \pm0.018 $ &
  $0.718 \pm0.018 $ &
  $1.130 \pm 0.005$ &
  $\underline{ 1.408 \pm 0.039}$ &
  $4.50$ &
  5 \\ \midrule
\multirow{8}{*}{Size} &
  Full-FT &
  $\underline{ 1.419 \pm 0.044 }$ &
  $0.745 \pm 0.008$ &
  $0.896 \pm 0.007$ &
  $\mathbf{ 1.893 \pm 0.035 }$ &
  $3.25$ &
  \multicolumn{1}{c|}{3} &
  $1.070 \pm 0.082$ &
  $0.719 \pm 0.010$ &
  $0.886 \pm 0.007$ &
  $\underline{ 1.906 \pm 0.006}$ &
  $4.00$ &
  4 \\
 &
  LP &
  $2.073 \pm 0.012$ &
  $0.912 \pm 0.004$ &
  $0.921 \pm 0.008$ &
  $2.381 \pm 0.006$ &
  $8.00$ &
  \multicolumn{1}{c|}{8} &
  $1.115 \pm 0.000$ &
  $0.829 \pm 0.000$ &
  $0.907 \pm 0.000$ &
  $2.246 \pm 0.000$ &
  $8.00$ &
  8 \\
 &
  Surgical-FT &
  $1.685 \pm 0.060$ &
  $0.775 \pm 0.007$ &
  $0.890 \pm 0.005$ &
  $2.145 \pm 0.022$ &
  $5.00$ &
  \multicolumn{1}{c|}{6} &
  $\mathbf{ 0.993 \pm0.000 }$ &
  $0.719 \pm0.000 $ &
  $\mathbf{ 0.860 \pm 0.000 }$ &
  $\underline{ 1.906 \pm 0.000}$ &
  $2.50$ &
  1 \\
 &
  LP-FT &
  $1.440 \pm 0.081$ &
  $\underline{ 0.735 \pm 0.013 }$ &
  $0.893 \pm 0.007$ &
  $1.905 \pm 0.016$ &
  $3.50$ &
  \multicolumn{1}{c|}{2} &
  $1.038 \pm 0.038$ &
  $\underline{ 0.694 \pm 0.012 }$ &
  $\underline{ 0.883 \pm 0.005 }$ &
  $1.913 \pm 0.031$ &
  $2.75$ &
  2 \\
 &
  WiSE-FT &
  $1.814 \pm 0.092$ &
  $0.831 \pm 0.007$ &
  $\mathbf{ 0.873 \pm 0.005 }$ &
  $1.951 \pm 0.024$ &
  $4.50$ &
  \multicolumn{1}{c|}{5} &
  $1.100 \pm0.005 $ &
  $\mathbf{ 0.691 \pm0.015 }$ &
  $0.894 \pm0.007 $ &
  $1.943 \pm0.039 $ &
  $4.50$ &
  6 \\
 &
  $L^2$-SP &
  $1.438 \pm 0.046$ &
  $0.799 \pm 0.002$ &
  $0.888 \pm 0.005$ &
  $2.101 \pm 0.016$ &
  $4.00$ &
  \multicolumn{1}{c|}{4} &
  $1.053 \pm0.026 $ &
  $0.720 \pm0.015 $ &
  $0.904 \pm0.002 $ &
  $2.122 \pm0.018 $ &
  $6.00$ &
  7 \\
 &
  Feature-map &
  $1.656 \pm 0.025$ &
  $0.880 \pm 0.011$ &
  $0.893 \pm 0.002$ &
  $2.252 \pm 0.008$ &
  $6.25$ &
  \multicolumn{1}{c|}{7} &
  $\mathbf{ 0.993 \pm 0.034 }$ &
  $0.724 \pm0.009 $ &
  $0.884 \pm0.001 $ &
  $1.970 \pm0.013 $ &
  $4.50$ &
  3 \\
 &
  BSS &
  $\mathbf{ 1.375 \pm 0.019 }$ &
  $\mathbf{ 0.731 \pm 0.007 }$ &
  $\underline{ 0.887 \pm 0.010 }$ &
  $\underline{ 1.900 \pm 0.016 }$ &
  $1.50$ &
  \multicolumn{1}{c|}{1} &
  $1.043 \pm0.022 $ &
  $0.703 \pm0.016 $ &
  $0.905 \pm0.005 $ &
  $\mathbf{ 1.890 \pm0.071 }$ &
  $3.75$ &
  5 \\ \midrule[2pt]
\end{tabular}%
\end{sc}
\end{small}
\end{adjustbox}
\end{center}
\vspace{-0.8cm}
\end{table*}

\definecolor{steelblue}{RGB}{0, 191, 255}
\definecolor{firebrick}{RGB}{240, 128, 128}
\definecolor{forestgreen}{RGB}{46, 139, 87}
\setlength{\abovecaptionskip}{0.2cm}
% \Shikun{Further trim the result analysis to 2.5 pages}
    \vspace{-2mm}
\begin{figure}[t]
     \centering
     \subfloat[\label{fig:1a}Self-supervised pre-training (Classification)]{\includegraphics[trim={0.6cm 0.6cm 0.6cm 0.1cm},clip,width=0.48\linewidth]{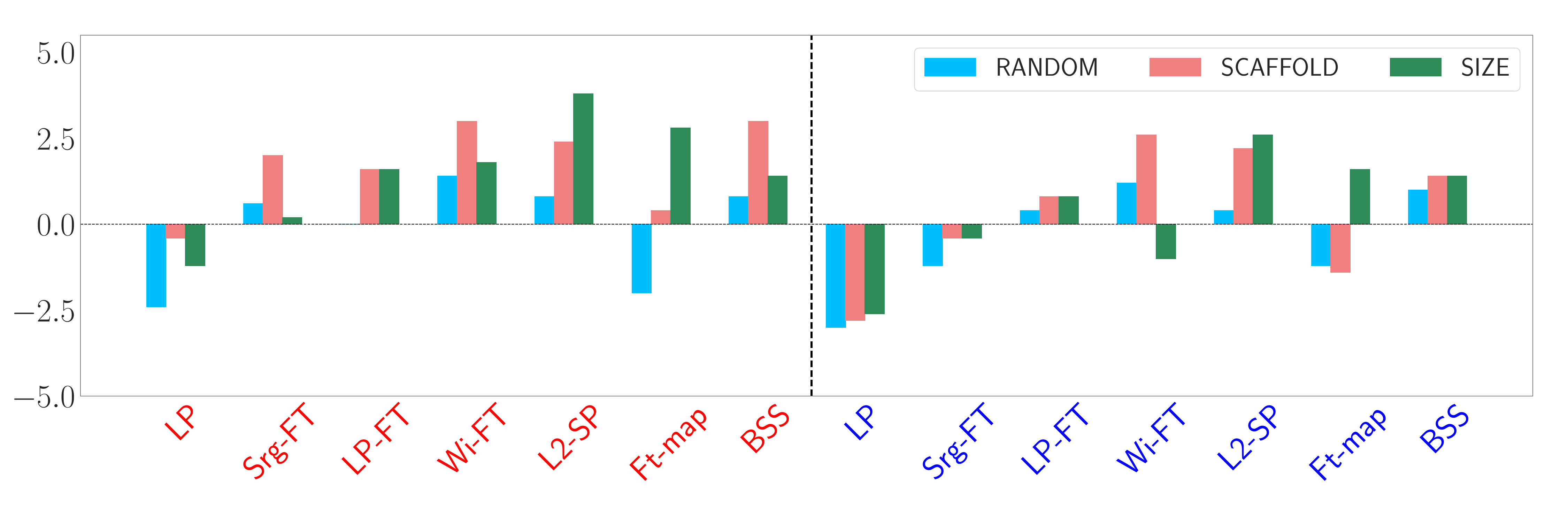}}
     \subfloat[\label{fig:1b}Supervised pre-training (Classification)]{\includegraphics[trim={0.6cm 0.6cm 0.6cm 0.1cm},clip,width=0.48\linewidth]{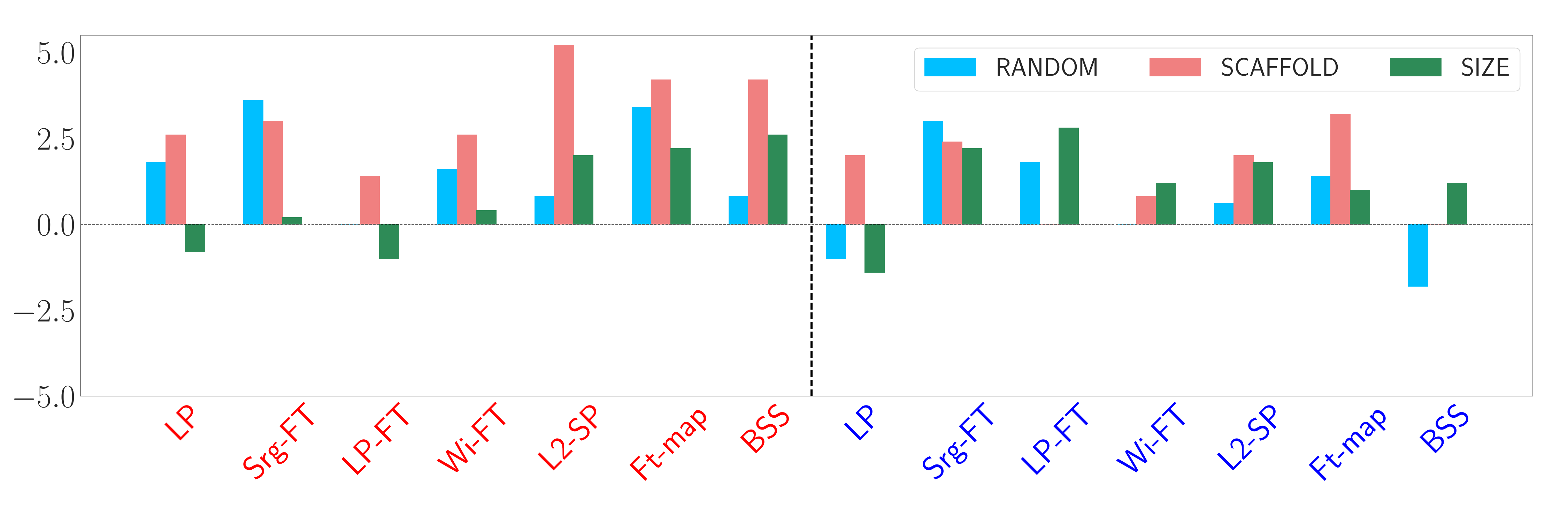}}\\
     \subfloat[\label{fig:2a}Self-supervised pre-training (Regression)]{\includegraphics[trim={0.6cm 0.6cm 0.6cm 0.1cm},clip,width=0.495\linewidth]{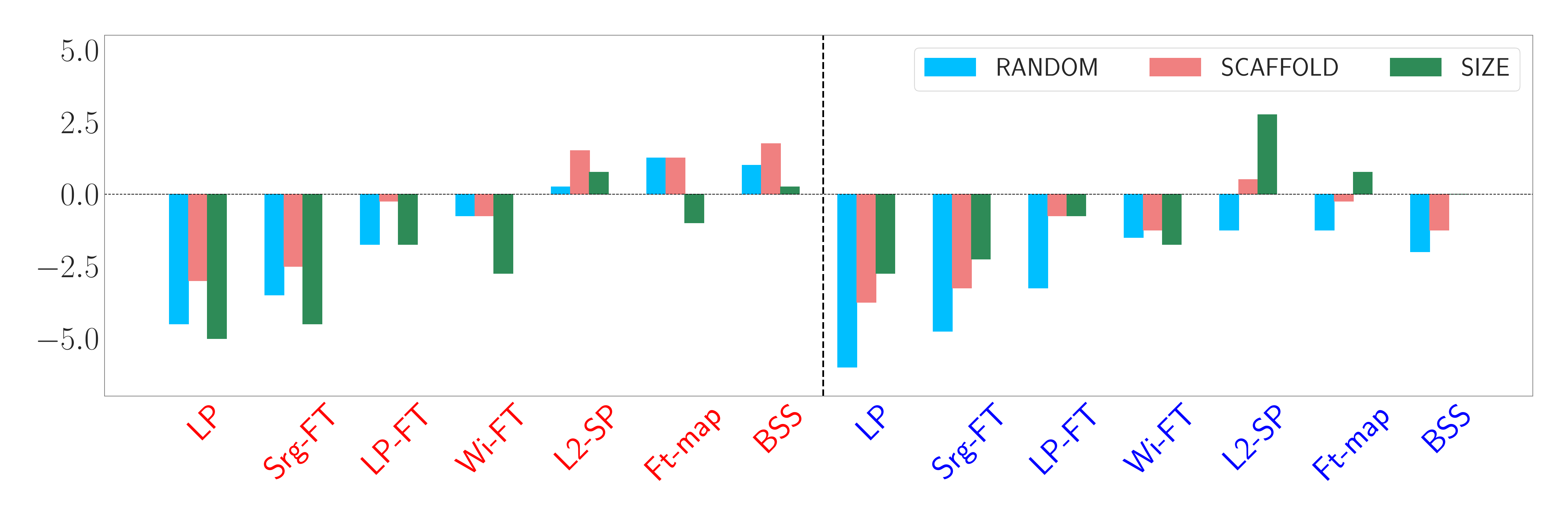}}
     \subfloat[\label{fig:2b}Supervised pre-training (Regression)]{\includegraphics[trim={0.6cm 0.6cm 0.6cm 0.1cm},clip,width=0.495\linewidth]{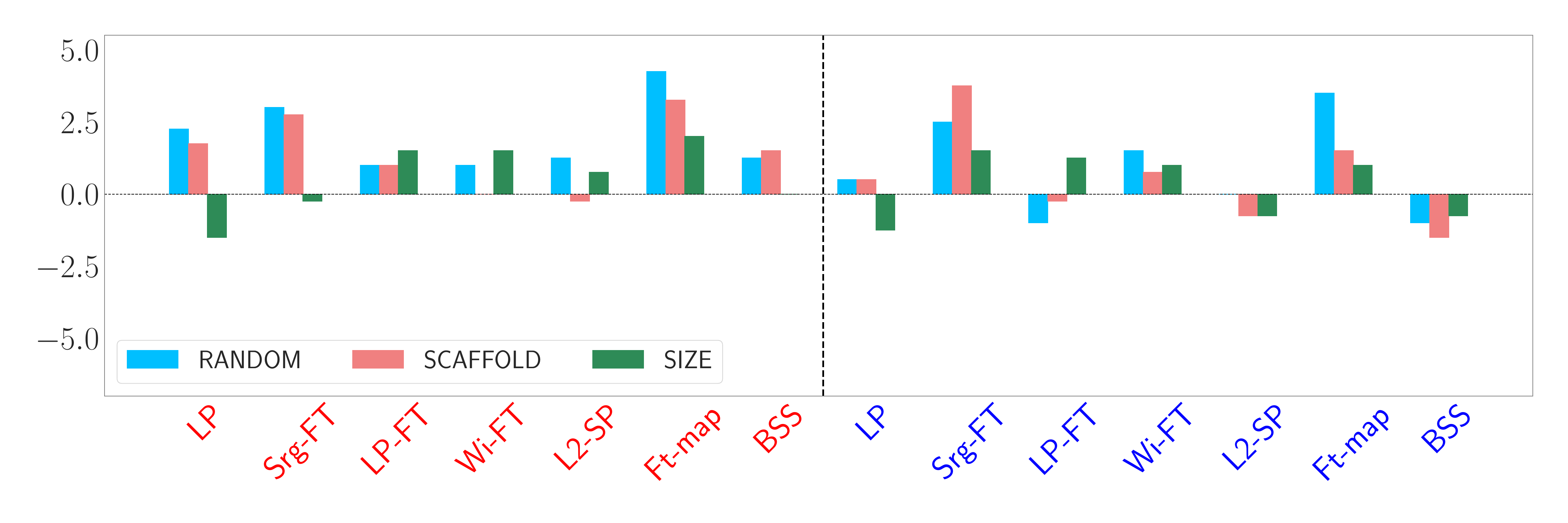}}
     \caption{Average Rank improvements over Full-fine-tuning for 7 robust fine-tuning methods in self-supervised and supervised pre-training
      across 8 \textit{classification} (\textit{a, b}) datasets and across 4 \textit{regression} (\textit{c, d}) datasets.
     Each subfigure presents {\color{red}few-shot-50} (left of the dashed line) and {\color{blue}few-shot-100} (right of the dashed line) settings, with {\color{steelblue}\textbf{random}}, {\color{firebrick}\textbf{scaffold}}, and {\color{forestgreen}\textbf{size}} splits.}
    % \caption{Average Test-AUC improvements (across 8 \textit{classification} datasets, \%) over Full-fine-tuning for 7 advanced fine-tuning methods in (a) self-supervised and (b) supervised pre-training scenarios. Each subfigure presents both {\color{red}few-shot-50} (right of the dashed line, colored in red) and {\color{blue}few-shot-100} (right of the dashed line, colored in blue) settings, with  {\color{steelblue}\textbf{random}}, {\color{firebrick}\textbf{scaffold}}, and {\color{forestgreen}\textbf{size}} splits.}

    \label{fig:maintext}
    \vspace{-6mm}
\end{figure}

This section mainly analyzes the experimental results from Mole-BERT and Graphium-Toy models as representatives of self-supervised and supervised pre-training.
Table~\ref{tab:model-reference} is a summary of all pre-trained models we test on and their corresponding result tables for reference. Since we observe similar trends from pre-trained models of the same category, we will refer to them in our result analysis and compare over different pre-trained models in Sec.~\ref{subsec:pretrain_comp}. Due to limited space, more findings with different fine-tuning methods and pre-trained models comparison can be found in Appendix~\ref{appendix:further}.  
% \shikun{revise here}
\vspace{-2mm}
\subsection{Self-supervised Pre-trained Models}
\vspace{-1mm}
% We benchmark the fine-tuning performance of Mole-BERT and GraphMAE as representative self-supervised pre-training models, and observe that Mole-BERT consistently outperforms GraphMAE across various fine-tuning methods under the same experimental conditions. 
% {\color{red}As a result, here we focus primarily on results of Mole-BERT in Table~\ref{}. The results of GraphMAE  are provided in Appendix~\ref{appendix:results}.}

% \Shikun{thinking about eliminating the non-few and few-shot subsections. We can combine the finding 3 to part of finding 6, combine finding 5 to part of finding 6 and combine finding 2 to part of finding 5. make it into 5 pieces by keeping 1 and 4 the way it is.  The new three should be 1). The reason why LP poor and compare to CV. 2). feature distortion less common in non-few-shot fine-tuning (so explains the consistent marginal improvement in finding 2) + more severe under non-few-shot, but different solutions lead to diverge results (finding 5, connect to finding 1 regarding the part with classification and regression) 3). few-shot model suffers the most from tuning part of the model, like the additional LP in LP-FT and surgical FT (connects finding 5 and 6)}

\textbf{\textit{Q1: How does self-supervised pre-training influence downstream prediction tasks?}}

% \begin{enumerate}[label=\textbf{(\alph*)}, leftmargin=*]
% \textbf{(1a) Molecular representations learned from self-supervised pre-training are likely to mismatch with downstream tasks and exhibit non-robust OOD performance. }

\textbf{(1a) Regression tasks require more task-specific knowledge from downstream fine-tuning compared to classification tasks.}

When checking the few-shot results in Fig.~\ref{fig:1a} and \ref{fig:2a}, full fine-tuning ranks the highest for regression tasks but only achieves mid-tier performance for classification tasks. This disparity likely arises from the distinct nature of these tasks. Classification tasks typically require coarser-grained features, as exemplified by the Tox21 dataset. In this case, determining toxicity may largely rely on recognizing certain functional groups, such as toxicophores or structural alerts~\citep{singh2016toxicophore}. In contrast, regression tasks demand finer-grained features. For example, predicting precise solubility involves factors such as partial charge distribution, conformational flexibility, and hydrogen bond patterns, among others~\citep{faller2007computational}. Consequently, models fine-tuned for regression tasks must acquire more downstream knowledge during the fine-tuning process and are generally less prone to overfitting compared to those used for classification tasks.

\textbf{(1b) Molecular representations learned from self-supervised pre-training are not informative enough for downstream tasks.}

As shown in Tables~\ref{tab-main1} and \ref{tab-main2}, LP is consistently the worst performing method for self-supervised pre-trained models across all data splits, even under the few-shot fine-tuning in Fig.~\ref{fig:1a} and \ref{fig:2a}. Furthermore, this behavior is widely observed across all tested self-supervised models as GraphMAE and MoleculeSTM, which contrasts the observations in CV where LP demonstrates robust OOD performance by preserving high quality and generalizable features from pre-trained embeddings~\citep{wortsman2022robust, kumar2022fine}. We attribute this to the misalignment between general-purpose representations produced by self-supervised pre-training and the features required by  the specific molecular tasks. Consequently, relying solely on tuning the classifier $g_{\mphi}$ is insufficient to extract meaningful predictions from these non-informative representations.

Below, we summarize insightful findings from the performance of different fine-tuning strategies.

$\bullet$ \textbf{Finding 1. Under few-shot fine-tuning, weight-based fine-tuning strategies stand out with WiSE-FT for classification tasks and $L^2$-SP for regression tasks.}

Among various fine-tuning methods, weight-based approaches consistently outperform others across a wide range of experiments, regardless of the few-shot sample sizes (Fig.~\ref{fig:1a} and \ref{fig:2a}). Self-supervised models are known to capture general-purpose knowledge for substructure discovery\citep{wang2024evaluating}. During fine-tuning, combining pre-trained and fine-tuned weights proves effective in extracting molecular patterns relevant to downstream tasks. Notably, WiSE-FT demonstrates superior performance on classification datasets, whereas $L^2$-SP excels in regression tasks. This finding is also supported by MoleculeSTM in table~\ref{tab-stm-large-rgs-few} where $L^2$-SP remains as top method under all few-shot regression tasks and WiSE-FT excels under Fewshot-50 classification. Essentially, WiSE-FT applies a straightforward post-hoc linear interpolation between pre-trained and fine-tuned models, governed by a single coefficient. In contrast, $L^2$-SP implicitly determines the weight combination through the training loss~\citep{lubana2022quadratic, xuhong2018explicit}, aligning with statement (1a) that regression tasks typically demand more nuanced modeling.

$\bullet$ \textbf{Finding 2. Partial FT results in underfitted molecular representations under Fewshot settings, which is more severe for regression tasks compared to classification.}
% \Shikun{thinking about how to alternate the finding summary to be more informative for 5 and 6. In particular, maybe try to express the feeling that few-shot fine-tuning suffers if some part of the model is biased like LP-FT and surgical FT.}

% We observe that LP and surgical FT performance are poor among table~\ref{}, which indicates the issue if we keep all the pre-trained model weights the same or only change a small portion of it. The inferior performance of LP even under few-shot fine-tuning further verifies our arguments in finding 3. The suboptimality with the pre-trained molecular representations obtained by self-supervised pre-training outweighs outweighs the potential drawbacks of few-shot fine-tuning, like over-adaptation and catastrophic forgetting. 
For the non-few-shot fine-tuning (Tables~\ref{tab-main1} and \ref{tab-main2}), surgical FT and LP-FT improve over full FT in both classification and regression tasks. However, in few-shot fine-tuning, both methods rank as the worst methods. This is likely because partial fine-tuning underfits and bias towards the the limited samples. This issue is more pronounced in regression tasks. %, which have more complex input-output relationships compared to binary classification tasks.

% However, we discover some different pattern with the supervised pre-trained model in sec~\ref{subsec:sup}. Besides, interestingly that surgical FT succeed in non-few-shot fine-tuning while fails in few-shot fine-tuning. This actually imply that surgical FT has the potential to alternate correct portion of weights when there is more samples to fine-tuning, but it will easily manipulate the weights in a wrong direction when we only unfreeze a subset of weights for fine-tuning. This is also the reason why weight-based and representation-based methods can perform better since they only add regularization on top of the full model fine-tuning instead of optimizing on a subset of parameters. 

% \subsubsection{Settings Comparison}
% \Shikun{maybe put into appendix due to limited space}

% \textbf{Different splits}
% \begin{itemize}
%     \item full FT works better under random splits and the benefits of other fine-tuning techniques comapred full FT larger in scaffold and size splits
%     \item BSS works well under scaffold few-shot cases under both Mole-BERT and GraphMAE
% \end{itemize}

% \textbf{Mole-BERT vs. GraphMAE}
% Under WiSE-FT, Mole-BERT prefers small alpha (more towards pretrained model) and GraphMAE prefers larger alpha (more towards full FT). This finding is consistent with how surgical FT is compared with full FT for both Mole-BERT and GraphMAE models.

\vspace{-3mm}
\subsection{Supervised Pre-trained Models}
\vspace{-2mm}
\label{subsec:sup}
% \Shikun{does not discuss separately for non-few-shot and few-shot since some of them are shared}

% Before diving into the detailed findings with different fine-tuning techniques for supervised pre-trained model, we first discuss the fundamental difference with the pre-training objectives between supervised and self-supervised pre-training. 

\textbf{\textit{Q2: How does supervised pre-training influence downstream tasks?}}

We first discuss the \textbf{task similarity} between the datasets used in the pre-training and downstream fine-tuning process. 
As introduced in Appendix.~\ref{appendix:pt_data}, the ToyMix dataset used for supervised pre-training contains QM9, Tox21 and Zinc12K. The predictions from QM9 are not directly related to our downstream tasks, but may involve indirect correlations, as the quantum chemical properties provided by QM9 are highly valuable for characterizing molecular features. \textbf{Tox21} is an overlapping dataset that also exists as one of the downstream datasets. Its tasks in predicting qualitative toxicity measurements are \textit{highly related} to the downstream \textbf{ClinTox} and \textbf{ToxCast} datasets, and also \textit{correlate} to the \textbf{Sider} dataset which contains evaluation in drug side effects. Lastly, Zinc12K, which is to predict the constrained
solubility, is relevant to the \textbf{Esol} and \textbf{Lipo} datasets that involve solubility predictions. Other downstream tasks \textit{do not share} the same tasks with pre-training \textit{directly}. Then we observe the following rules. 
% We compare the performance of self-supervised and supervised fine-tuning on downstream tasks given the task overlap.

% \begin{enumerate}[label=\textbf{(\alph*)}, leftmargin=*]

     % \textbf{(a) With non-few-shot fine-tuning, supervised pre-training provides more useful molecular representations when the pre-training and fine-tuning share similar tasks.}
\textbf{(2a) Under few-shot fine-tuning, supervised pre-training models generally yield higher fine-tuning performance compared to self-supervised pre-training,  regardless of the pre-training and fine-tuning task correlations.} 

Supervised pre-training brings more benefits to downstream tasks than self-supervised pre-training in few-shot situations when checking Tables~\ref{tab-clf-fewshot-bert-sup} and \ref{tab-regs-fewshot-bert-sup}. Besides, the benefits are less relevant to the task similarity in contrast to the non-few-shot cases. For example, the improvements are also observed in HIV and Cep datasets even their tasks do not share with pre-training tasks directly. This implies that learned domain-specific knowledge still offer better insights than generic knowledge when fine-tuning guidance is minimal. 
    
% \textbf{(2b) Under non-few-shot fine-tuning, supervised pre-training has better fine-tuning performances than self-supervised pre-training when its objectives align closely with downstream tasks. However, it may hurt downstream performance if the tasks do not align.}
\textbf{(2b) Under non-few-shot fine-tuning, fine-tuning performance given supervised pre-training outperforms self-supervised pre-training when its objectives closely align with downstream tasks, while task misalignment may harm performance. }

From Tables~\ref{tab-main1} and \ref{tab-main2}, we observe consistent fine-tuning performance  improvements over self-supervised pre-training on highly task-correlated downstream datasets including ClinTox, Esol, Lipo and Tox21. Even when pre-training involves regression tasks and downstream tasks are classification, performance gains occur if the physical meanings align. For datasets that do not directly share tasks with pre-training, we observe mixed performance on Sider, Malaria, and Cep datasets, and even worse performance on HIV and MUV datasets. This observation contrasts to few-shot cases in (2a), which entails that downstream task specific knowledge can be learned given sufficient guidance on top of generic knowledge from self-supervised pre-training.   
% \pan{How do you know Esol, Lipo are related while Malaria is not? wIll the decay for HIV and MUV be just regression pre-training v.s. classification fine-tuning?}
% This finding partially aligns with~\citep{sun2022does}, which concluded that naive supervised pre-training with aligned target labels only yields marginal improvements over self-supervised pre-training, while combining supervised objectives with self-supervised pre-training is better.
% The difference is that they pre-trained on single ChEMBL dataset~\citep{gaulton2012chembl} and did not evaluate for few-shot fine-tuning or on regression datasets. 
    
Below are some detailed findings with different fine-tuning methods given supervised pre-training.

$\bullet$ \textbf{Finding 3. Fine-tuning strategies that regularizes towards pre-trained molecular representations rank top, while weight-based methods are suboptimal.}

From non-few-shot (Tables~\ref{tab-main1} and \ref{tab-main2}) and few-shot fine-tuning (Figs.~\ref{fig:1b} and \ref{fig:2b}) in both supervised models with ToyMix and LargeMix, surgical FT and Feature-map tend to be the top-ranking methods. However, best performing weight-based methods for self-supervised pre-training, only show mediocre performance here. This can also be observed in the larger-scale GraphGPS model as discussed in Appendix~\ref{appendix:pretrain_discussion}. In addition, the other representation-based method BSS shows limited performance compared to Feature-map, which directly regularizes the distance to pre-trained representations. These observations suggest that given the task alignment between supervised pre-training and downstream fine-tuning, pre-trained representations tend to contain transferable features for downstream tasks. Consequently, controlling the degree to preserve pre-trained representations is the key to downstream fine-tuning performance. 

\vspace{-2mm}
\subsection{Discussions over Pre-trained Models}
\vspace{-2mm}
\label{subsec:pretrain_comp}
 Our extensive evaluation shows that the ranking of fine-tuning techniques remains consistent across pre-trained models within the same category, either supervised or self-supervised, regardless of model architecture, scale, or pre-training dataset. This suggests that our guidance for selecting fine-tuning methods based on the pre-training paradigm is broadly applicable and generalizable across diverse model designs. For instance, self-supervised models such as Mole-BERT and MoleculeSTM tend to benefit more from weight-based fine-tuning, while supervised models like Graphium and GraphGPS perform better with feature-based approaches.
% \shikun{revise language here}
% In addition, we find that supervised pre-trained models tend to have greater benefits when facing few-shot fine-tuning scenarios, especially on regression tasks.
% Regarding overall fine-tuning performance, we observe that results tend to be similar within each pre-training category but remain distinguishable from those of other techniques—for instance, Mole-BERT and the ToyMix model still underperform compared to MoleculeSTM \pan{do not under this sentence}. 

% However, minor performance variations do occur and are further explored in Appendix~\ref{appendix:further_discussion}.

% Regarding the overall fine-tuning performance, we find they tend to be close in a range and remain distinguiable with the perfomances of the other pre-training technique, like statement (2a) and (2b) also hold for LargeMix supervised model. However, the exact performance can have a small gap, and we will discuss the reason more in Appendix~\ref{appendix:further_discussion}.

\vspace{-1mm}
\section{Methodology Exploration}
\vspace{-1mm}
\label{sec:design}

\begin{table*}[t]
\vspace{-4mm}
\caption{\proj performance on 4 \hlb{\textbf{Regression}}{30}
datasets (RMSE metrics) in the \hlr{\textbf{Fewshot}}{30} setting with 50, 100 samples, evaluated across 3 dataset splits (\textsc{Random, Scaffold, Size}) given \hlg{\textbf{\textsc{Mole-BERT}}}{30} model. \textsc{Avg-R} denote the average rank. Standard deviations across five replicates are shown. We \textbf{bold} and \underline{underline} the best and second-best performances in each scenario. 
% \pan{what is TOP?; Why do we need to compare with top? If we cannot get consistently better than top, what is the sense to put it here? We can just leave the statement that our methods achieve the best average rank, and even achieve xxx top performance in xxx (how many over how many) datasets. } \Shikun{Since there is a reviewer questioning if we are comparing DWiSE-FT to all baselines, I'm wondering if we gonna keep the "best" row}
}
\vspace{-2mm}
\begin{center}
\begin{adjustbox}{width = 0.9\textwidth}
\begin{small}
\begin{sc}
\begin{tabular}{lcccccc|ccccc}
\toprule
& & \multicolumn{5}{c}{Fewshot 50} & \multicolumn{5}{c}{Fewshot 100} \\
\cmidrule(lr){3-7}\cmidrule(lr){8-12}
 Split & Methods  &   esol &  lipo  &     malaria    & cep  & AVG &   esol &  lipo  &     malaria    & cep  & AVG \\
\midrule
\multirow{4}{*}{Random} 

% & WiSE-FT & $1.384 \pm 0.047 $ & $1.212 \pm 0.020 $ & $1.276 \pm 0.007 $ & $2.410 \pm 0.051 $ & & $1.189 \pm 0.030 $ & $1.142 \pm 0.025 $ & $1.256 \pm 0.006 $ & $2.211 \pm 0.028 $ & & $0.995 \pm 0.010 $ & $0.855 \pm 0.011 $ & $1.193 \pm 0.003 $ & $1.893 \pm 0.021 $\\
% & $L^2$-SP & $1.372  \pm 0.029$ & $1.196  \pm 0.019$ & $1.277  \pm 0.006$ & $2.280  \pm 0.031$ & & $1.161  \pm 0.016$ & $1.149  \pm 0.007$ & $1.260  \pm 0.004$ & $2.131  \pm 0.014$ & & $0.878  \pm 0.026$ & $0.806  \pm 0.007$ & $1.192  \pm 0.004$ & $1.893  \pm 0.018$\\
% & Top & $1.329 \pm 0.021$ & $1.164  \pm 0.010$ & $1.271  \pm 0.007$ & $2.275  \pm 0.022$ & & $1.120 \pm 0.038$ & $1.139 \pm 0.017$ & $1.256 \pm 0.006$ & $2.131 \pm 0.014$ & & $ 0.878\pm 0.026$ & $0.806 \pm 0.007$ & $1.192 \pm 0.004$ & $1.862 \pm 0.010$\\
% & DWiSE-FT & $ 1.378 \pm 0.055$ & $ 1.189 \pm 0.020$ & $ 1.273 \pm 0.009$ & $ 2.222 \pm 0.059$ & & $ 1.132\pm 0.025$ & $ 1.138\pm 0.028$ & $ 1.256\pm 0.004$ & $ 2.129\pm 0.020$ & & $ 0.918\pm 0.012$ & $ 0.818\pm 0.013$ & $ 1.192\pm 0.004$ & $ 1.865\pm 0.030$\\

&  WiSE-FT  &  $1.384 \pm 0.047 $  &  $1.212 \pm 0.020 $  &  $1.276 \pm 0.007 $  &  $2.410 \pm 0.051 $  & $3.75$ &  $1.189 \pm 0.030 $  &  $1.142 \pm 0.025 $  & $\mathbf{ 1.256 \pm 0.006  }$ &  $2.211 \pm 0.028 $  & $3.00$ \\
 &  $L^2$-SP  & $\underline{ 1.372  \pm 0.029 }$ &  $1.196  \pm 0.019$  &  $1.277  \pm 0.006$  &  $2.280  \pm 0.031$  & $3.00$ &  $1.161  \pm 0.016$  &  $1.149  \pm 0.007$  &  $1.260  \pm 0.004$  & $\underline{ 2.131  \pm 0.014 }$ & $3.25$ \\
 &  Top  & $\mathbf{ 1.329 \pm 0.021 }$ & $\mathbf{ 1.164  \pm 0.010 }$ & $\mathbf{ 1.271  \pm 0.007 }$ & $\underline{ 2.275  \pm 0.022 }$ & $1.25$ & $\mathbf{ 1.120 \pm 0.038 }$ & $\underline{ 1.139 \pm 0.017 }$ & $\mathbf{ 1.256 \pm 0.006 }$ & $\underline{ 2.131 \pm 0.014 }$ & $1.50$ \\
 &  DWiSE-FT  &  $ 1.378 \pm 0.055$  & $\underline{  1.189 \pm 0.020 }$ & $\underline{  1.273 \pm 0.009 }$ & $\mathbf{  2.222 \pm 0.059 }$ & $2.00$ & $\underline{  1.132\pm 0.025 }$ & $\mathbf{  1.138\pm 0.028 }$ & $\mathbf{  1.256\pm 0.004 }$ & $\mathbf{  2.129\pm 0.020 }$ & $1.25$ \\

\midrule
\multirow{4}{*}{Scaffold} 

% & WiSE-FT & $1.842 \pm 0.056 $ & $1.177 \pm 0.009 $ & $1.162 \pm 0.004 $ & $2.454 \pm 0.043 $ & & $1.544 \pm 0.063 $ & $1.041 \pm 0.017 $ & $1.151 \pm 0.007 $ & $2.301 \pm 0.042 $ & & $1.388 \pm 0.023 $ & $0.834 \pm 0.012 $ & $1.114 \pm 0.002 $ & $1.936 \pm 0.037 $\\
% & $L^2$-SP & $1.699  \pm 0.049$ & $1.086  \pm 0.009$ & $1.162  \pm 0.002$ & $2.331  \pm 0.024$ & & $1.473  \pm 0.009$ & $0.961  \pm 0.003$ & $1.153  \pm 0.002$ & $2.201  \pm 0.038$ & & $1.163  \pm 0.026$ & $0.813  \pm 0.010$ & $1.126  \pm 0.011$ & $1.885  \pm 0.011$\\
% & Top & $1.680  \pm 0.042$ & $1.036  \pm 0.007$ & $1.159  \pm 0.000$ & $2.292  \pm 0.026$ & & $1.436 \pm 0.054$ & $0.937 \pm 0.008$ & $1.149 \pm 0.003$ & $2.187 \pm 0.034$ & & $ 1.112\pm 0.015$ & $0.802 \pm 0.003$ & $1.114 \pm 0.002$ & $1.881 \pm 0.010$\\
% & DWiSE-FT & $ 1.616 \pm 0.047$ & $ 1.110 \pm 0.013$ & $ 1.173 \pm 0.005$ & $ 2.306 \pm 0.030$ & & $ 1.485\pm 0.041$ & $ 0.979\pm 0.014$ & $ 1.158\pm 0.009$ & $2.149 \pm 0.040$ & & $ 1.266\pm 0.021$ & $ 0.823\pm 0.010$ & $ 1.121\pm 0.004$ & $ 1.900\pm 0.019$\\

&  WiSE-FT  &  $1.842 \pm 0.056 $  &  $1.177 \pm 0.009 $  & $\underline{ 1.162 \pm 0.004  }$ &  $2.454 \pm 0.043 $  & $3.50$ &  $1.544 \pm 0.063 $  &  $1.041 \pm 0.017 $  & $\underline{ 1.151 \pm 0.007  }$ &  $2.301 \pm 0.042 $  & $3.50$ \\
 &  $L^2$-SP  &  $1.699  \pm 0.049$  & $\underline{ 1.086  \pm 0.009 }$ & $\underline{ 1.162  \pm 0.002 }$ &  $2.331  \pm 0.024$  & $2.50$ & $\underline{ 1.473  \pm 0.009 }$ & $\underline{ 0.961  \pm 0.003 }$ &  $1.153  \pm 0.002$  &  $2.201  \pm 0.038$  & $2.50$ \\
 &  Top  & $\underline{ 1.680  \pm 0.042 }$ & $\mathbf{ 1.036  \pm 0.007 }$ & $\mathbf{ 1.159  \pm 0.000 }$ & $\mathbf{ 2.292  \pm 0.026 }$ & $1.25$ & $\mathbf{ 1.436 \pm 0.054 }$ & $\mathbf{ 0.937 \pm 0.008 }$ & $\mathbf{ 1.149 \pm 0.003 }$ & $\underline{ 2.187 \pm 0.034 }$ & $1.25$ \\
 &  DWiSE-FT  & $\mathbf{  1.616 \pm 0.047 }$ &  $ 1.110 \pm 0.013$  &  $ 1.173 \pm 0.005$  & $\underline{  2.306 \pm 0.030 }$ & $2.50$ &  $ 1.485\pm 0.041$  &  $ 0.979\pm 0.014$  &  $ 1.158\pm 0.009$  & $\mathbf{ 2.149 \pm 0.040 }$ & $2.75$ \\

\midrule
\multirow{4}{*}{Size} 

% & WiSE-FT & $2.615 \pm 0.072 $ & $1.391 \pm 0.042 $ & $0.929 \pm 0.004 $ & $2.762 \pm 0.053 $ & & $2.216 \pm 0.056 $ & $1.124 \pm 0.031 $ & $0.917 \pm 0.004 $ & $2.543 \pm 0.027 $ & & $2.071 \pm 0.078 $ & $0.902 \pm 0.016 $ & $0.912 \pm 0.003 $ & $2.379 \pm 0.086 $\\
% & $L^2$-SP & $2.393  \pm 0.068$ & $1.306  \pm 0.037$ & $0.915  \pm 0.002$ & $2.497  \pm 0.019$ & & $1.731  \pm 0.071$ & $1.025  \pm 0.028$ & $0.905  \pm 0.002$ & $2.424  \pm 0.024$ & & $1.629  \pm 0.084$ & $0.821  \pm 0.011$ & $0.904  \pm 0.003$ & $2.368  \pm 0.013$\\
% & Top & $2.369  \pm 0.075$ & $1.297  \pm 0.040$ & $0.911  \pm 0.002$ & $2.497  \pm 0.019$ & & $1.731 \pm 0.071$ & $1.025 \pm 0.028$ & $0.898 \pm 0.003$ & $2.424 \pm 0.024$ & & $1.629 \pm 0.084$ & $0.803 \pm 0.006$ & $0.895 \pm 0.002$ & $2.328 \pm 0.017$\\
% & DWiSE-FT & $ 1.488 \pm 0.101$ & $ 1.113 \pm 0.021$ & $ 0.913 \pm 0.007$ & $ 2.539 \pm 0.023$ & & $ 1.469\pm 0.052$ & $ 1.031\pm 0.022$ & $ 0.920\pm 0.006$ & $2.390 \pm 0.025$ & & $ 1.466\pm 0.040$ & $ 0.816\pm 0.022$ & $0.915 \pm 0.003 $ & $ 2.322\pm 0.031$\\

&  WiSE-FT  &  $2.615 \pm 0.072 $  &  $1.391 \pm 0.042 $  &  $0.929 \pm 0.004 $  &  $2.762 \pm 0.053 $  & $4.00$ &  $2.216 \pm 0.056 $  &  $1.124 \pm 0.031 $  &  $0.917 \pm 0.004 $  &  $2.543 \pm 0.027 $  & $3.75$ \\
 &  $L^2$-SP  &  $2.393  \pm 0.068$  &  $1.306  \pm 0.037$  &  $0.915  \pm 0.002$  & $\mathbf{ 2.497  \pm 0.019 }$ & $2.50$ & $\underline{ 1.731  \pm 0.071 }$ & $\mathbf{ 1.025  \pm 0.028 }$ & $\underline{ 0.905  \pm 0.002 }$ & $\underline{ 2.424  \pm 0.024 }$ & $1.75$ \\
 &  Top  & $\underline{ 2.369  \pm 0.075 }$ & $\underline{ 1.297  \pm 0.040 }$ & $\mathbf{ 0.911  \pm 0.002 }$ & $\mathbf{ 2.497  \pm 0.019 }$ & $1.50$ & $\underline{ 1.731 \pm 0.071 }$ & $\mathbf{ 1.025 \pm 0.028 }$ & $\mathbf{ 0.898 \pm 0.003 }$ & $\underline{ 2.424 \pm 0.024 }$ & $1.50$ \\
 &  DWiSE-FT  & $\mathbf{  1.488 \pm 0.101 }$ & $\mathbf{  1.113 \pm 0.021 }$ & $\underline{  0.913 \pm 0.007 }$ &  $ 2.539 \pm 0.023$  & $1.75$ & $\mathbf{  1.469\pm 0.052 }$ &  $ 1.031\pm 0.022$  &  $ 0.920\pm 0.006$  & $\mathbf{ 2.390 \pm 0.025 }$ & $2.25$ \\

\bottomrule
\label{table:new_method_reg}
\end{tabular}
\end{sc}
\end{small}
\end{adjustbox}
\end{center}
\vspace{-7mm}
\end{table*}
\vspace{-1.5mm}

Based on findings in Sec.~\ref{sec:result}, we observe that weight-based fine-tuning generally performs well under self-supervised pre-training. However, the top strategy varies: WiSE-FT excels in classification tasks, while $L^2$-SP is more effective for regression tasks. This motivates us to further explore the connections and trade-offs between these methods to identify potential improvements. In this section, we introduce \proj, an extension of the weight ensemble method unifying the strengths from WiSE-FT and $L^2$-SP. \proj demonstrates top-ranking results through efficient post-processing that better suits the practical fine-tuning needs. 
\vspace{-1mm}
\subsection{Motivation}
\vspace{-1mm}

As introduced in Sec.~\ref{sec:prelim}, WiSE-FT adopts the post-hoc linear interpolation between the pre-trained and fine-tuned model weights as $(1-\alpha)\cdot \mtheta_{\text{pre}} + \alpha \cdot \mtheta_{\text{ft}}$. Although $L^2$-SP does not explicitly have weight interpolation in the form, the optimal weight $\tilde{\mtheta}_{\text{ft}}$ from the weight-regularized loss $\tilde{\dL}(\mtheta)$ is indeed the linear interpolation of the optimal model from full FT $\mtheta^*_{\text{ft}}$ and the pre-trained model $\mtheta_{\text{pre}}$.

% $L^2$-SP fine-tunes over the loss $\tilde{\mathcal{L}}(\mtheta) = \mathcal{L}(\mtheta)+\frac{\delta}{2} \|{\mtheta - \mtheta_{\text{pre}}}\|^2_2$. Although $L^2$-SP does not explicitly have weight interpolation in the form, the optimal weight $\tilde{\mtheta}_{\text{ft}} = \mathrm{argmin}_{\mtheta} \tilde{\mathcal{L}}(\mtheta)$ is indeed the linear interpolation of the optimal model from full fine-tuning $\mtheta^*_{\text{ft}} = \mathrm{argmin}_{\mtheta} {\mathcal{L}}(\mtheta)$ and the pre-trained model $\mtheta_{\text{pre}}$ 

% with mixing coefficient $\frac{\lambda_i}{\lambda_i+\delta}$ in the direction defined by the $i$-th eigenvector of $\mathbf{H}$ with $\mathbf{H}$ being the Hessian matrix of $\mathcal{L}$ w.r.t. $\mtheta_{\text{ft}}$ as in~\citep{xuhong2018explicit} \pan{do you need to cite, can you just give a proposition to prove it, which enhances the depth of this work?}
\begin{proposition}
Given $\tilde{\mathcal{L}}(\mtheta) = \mathcal{L}(\mtheta)+\frac{\delta}{2} \|{\mtheta} - \mtheta_{\text{pre}}\|^2_2$, we define the optimal weights as $\tilde{\mtheta}_{\text{ft}} = \mathrm{argmin}_{\mtheta} \tilde{\mathcal{L}}(\mtheta)$ and $\mtheta^*_{\text{ft}} = \mathrm{argmin}_{\mtheta} {\mathcal{L}}(\mtheta)$.
\begin{equation}\label{eq:dwise}
\mathbf{Q}^T\tilde{\mtheta}_{\text{ft}} = (\mathbf{\Lambda}+\delta \mathbf{I})^{-1}\mathbf{\Lambda}\mathbf{Q}^T \mtheta^*_{\text{ft}} + \delta (\mathbf{\Lambda}+\delta \mathbf{I})^{-1}\mathbf{Q}^T\mtheta_{\text{pre}}
\enspace.
\end{equation}
where $\mH$ is the hessian matrix of $\mathcal{L}$ evaluated at $\mtheta^*_{\text{ft}}$ and $\mH = \mQ\mLambda \mQ^T$.
\end{proposition}

% \begin{equation}\label{eq:dwise}
% \mathbf{Q}^\top\tilde{\mtheta}_{\text{ft}} = (\mathbf{\Lambda}+\delta \mathbf{I})^{-1}\mathbf{\Lambda}\mathbf{Q}^\top \mtheta^*_{\text{ft}} + \delta (\mathbf{\Lambda}+\delta \mathbf{I})^{-1}\mathbf{Q}^\top\mtheta_{\text{pre}}
% \enspace.
% \end{equation}
% where $\mH$ is the hessian matrix of $\mathcal{L}$ evaluated at $\mtheta^*_{\text{ft}}$ and $\mH = \mQ\mLambda \mQ^T$.

Namely, $L^2$-SP can be seen as a more tailored weight ensemble method, employing variable mixing coefficients for different weights. This approach balances the influence of the prediction loss and the degree of weight regularization, unlike the fixed interpolation controlled by $\alpha$ across all weights in WiSE-FT. By accounting for subtle differences in loss values, $L^2$-SP is better suited for regression tasks, which are more sensitive to numerical variations. %This also explains why WiSE-FT suits better for classification tasks that demands less subtle knowledge while struggles with regression tasks that requires more complex molecular modeling. 

While $L^2$-SP excels on regression datasets, its regularization coefficient is less interpretable and necessitates retraining when experimenting with different values. In contrast, WiSE-FT offers a simpler and more flexible approach, performing post-hoc interpolation without additional training once the model is fine-tuned once. Furthermore, the mixing coefficient $\alpha$ is both easy to adjust and straightforward to interpret.
% Therefore, for regression datasets \pan{not only for regression dataset but for both reg and clas, can we have a method that take advantages of both fine-tuning approaches and apply to both cases?}, our goal is to inherit the format of WiSE-FT for the \textit{simplicity of post-hoc interpolation \pan{no need to say this again, redundency, just benefit from both sides and solve both tasks}}, while simultaneously incorporating the \textit{precision of nuanced weight ensemble} from $L^2$-SP \pan{no need to say this again, redundency,just benefit from both sides and solve both tasks}. 
Therefore, our goal is to find a method that benefits from both WiSE-FT and $L^2$-SP to accommodate regression and classification tasks at the same time.

\vspace{-1mm}
\subsection{Algorithm}
\vspace{-1mm}
We propose \proj that shares the framework of using the $\alpha$ to control the weight ensemble between the pre-trained model and fine-tuned model. The key idea, inspired by Eq.~\ref{eq:dwise} is to enable different $\alpha$ values when ensembling the weights for different encoder layers as shown in Fig.~\ref{fig:pipeline}. Given the pre-trained model with parameters $\mtheta_{\text{pre}}$ and model after full fine-tuning with parameters $\mtheta_{\text{ft}}$, 
% with the $i$-th layer parameters indexed by $[\mtheta_{\text{pre}}]_i$ and $[\mtheta_{\text{ft}}]_i$. 
The interpolated model has weights $\boldsymbol{\theta }^{\left[ i \right]}$ with mixing coefficient $\alpha_i$ for the $i$-th layer as:
\vspace{-1mm}
\begin{align}
    \boldsymbol{\theta }^{\left[ i \right]} = \left( 1-\alpha _i \right) \cdot \boldsymbol{\theta }_{\mathrm{pre}}^{\left[ i \right]}+\alpha _i \cdot \boldsymbol{\theta }_{\mathrm{ft}}^{\left[ i \right]}
\end{align}
\vspace{-5mm}

This approach naturally incorporates the characteristics of $L^2$-SP and even surgical FT: The weight ensemble in \proj offers the flexibility through varying mixing layer-wise coefficients between the pre-trained and fine-tuned models, addressing the limitations of WiSE-FT. Additionally, we enable the selection of $\boldsymbol{\alpha}$ through optimization via validation loss gradient inspired by the Gradient-based Neural Architecture Search (NAS)~\citep{dong2019searching}.

% \textbf{Automated hyperparameter tuning}
% Furthermore, the hyperparameter $\alpha$ that determines interpolation in WiSE-FT is easier to control and interpret in contrast to the regularization parameter $\delta$ in $L^2$-SP. $\alpha$ can be regarded as a combination of $\delta$ and the gradient information of the task prediction loss. With that being said, it can be hard to find an appropriate range of $\delta$ that mimics the desired effect of weight ensemble during fine-tuning in practice. It would be ideal to adopt the $\alpha$ as hyperparameter, but let it being tuned automatically to mimic the effect of $\delta$.

% The hyperparameter $\alpha$ in WiSE-FT, which determines interpolation, is easier to control and interpret compared to the regularization parameter $\delta$ in $L^2$-SP. An ideal $\alpha$ should reflect the combination of $\delta$ and the gradient information of the task prediction loss. Given this, identifying an appropriate range for $\delta$ that effectively mimics the desired weight ensemble effect during fine-tuning can be challenging in practice. Ideally, it would be beneficial to adopt $\alpha$ as the hyperparameter while enabling its automatic tuning to replicate the effects of $\delta$. This approach would combine the straightforwardness of WiSE-FT with the nuanced integration capabilities of $L^2$-SP.

\vspace{-1mm}
\subsection{Experiment results}
\vspace{-1mm}
Regarding the classification datasets, \proj should have the performance at least as good as WiSE-FT since WiSE-FT is the special case of \proj with one fixed mixing coefficient. We evaluate \proj to see how it improves upon WiSE-FT and matches the superior performance of $L^2$-SP for regression tasks under few-shot fine-tuning. Please note that, due to space constraints, we only present the experiments for few-shot fine-tuning with 50 and 100 samples in the main text. The complete table is available in Appendix E, Table~\ref{table:new_method_appendix}. In Table~\ref{table:new_method_reg}, we compare \proj's performance against WiSE-FT, $L^2$-SP, and the best-performing method in each setting. Specifically, we find that \proj consistently outperforms WiSE-FT. Furthermore, \proj often surpasses $L^2$-SP or at least maintains comparable results in most scenarios. Additionally, in some cases, \proj even exceeds the performance of the best-performing methods. Therefore, \proj can be a great candidate for fine-tuning on regression datasets in practice since it guarantees top performance with easier usage.

\vspace{-1mm}
\section{Conclusion}
\label{sec:conclusion}
This work benchmarks totally 8 fine-tuning methods, categorizing them into three groups, and evaluate them across 12 downstream datasets under 36 different experimental settings covering 3 dataset splits, 4 training sample sizes, and 6 molecular pre-trained models. 
% The design of these settings reflects diverse demands of molecular representation FT, namely label scarcity in downstream samples, distribution shifts between training and inference data, and the availability of diverse pre-trained molecular-graph foundation models \pan{diversified FMs (supervised, unsupervised), diversified tasks (in particular regression, not studied much in previou literatures), label sparsity, perhaps, no need to emphasize distribution shift.} . 
The design of these settings reflects practical demands of molecular representation fine-tuning under 1) diversified foundation model with both supervised and self-supervised pre-training, 2) wide range of downstream tasks in both classification and regression that has not been widely studied by previous literature and 3) scarcely labeled molecules for fine-tuning.
The study analyzes what is needed when facing classification vs. regression tasks and when given supervised vs. self-supervised pre-training. Then, we provide insights in best performing fine-tuning methods accordingly under aforementioned scenarios. 
% \pan{insights again: the need difference between regression and classification, supervised vs self-supervised, the best for supervised and the best for self-supervised (we have an improved approach) }. 
Additionally, we propose an extended fine-tuning method \proj, driven by our observations, that maintains top-ranking results through a more efficient and automated design for certain fine-tuning scenarios. This highlights the value of our benchmark in offering valuable insights for both fine-tuning methodology design and practical guidance in molecular representation learning.

\section{Acknowledgments}
\label{sec:ack}

S. Liu, D. Zou, N. Shoghi and P. Li are partially
supported by NSF awards PHY-2117997, IIS-2435957, IIS-2428777 and IIS-2239565; the JP-Morgan Chase Faculty Award, the NVIDIA Academic Grant Program and the IDEaS Cyberinfrastructure Awards.

% \newpage
\bibliographystyle{unsrtnat}
\bibliography{ref}

\clearpage
\appendix

\newpage
\section*{NeurIPS Paper Checklist}

\begin{enumerate}

\item {\bf Claims}
    \item[] Question: Do the main claims made in the abstract and introduction accurately reflect the paper's contributions and scope?
    \item[] Answer: \answerYes{} % Replace by \answerYes{}, \answerNo{}, or \answerNA{}.
    \item[] Justification: The main claims are justified by the experimental results and discussion in sec~\ref{sec:result}. Also, we refer the claims to the later findings in introduction. 
    \item[] Guidelines:
    \begin{itemize}
        \item The answer NA means that the abstract and introduction do not include the claims made in the paper.
        \item The abstract and/or introduction should clearly state the claims made, including the contributions made in the paper and important assumptions and limitations. A No or NA answer to this question will not be perceived well by the reviewers. 
        \item The claims made should match theoretical and experimental results, and reflect how much the results can be expected to generalize to other settings. 
        \item It is fine to include aspirational goals as motivation as long as it is clear that these goals are not attained by the paper. 
    \end{itemize}

\item {\bf Limitations}
    \item[] Question: Does the paper discuss the limitations of the work performed by the authors?
    \item[] Answer: \answerYes{} % Replace by \answerYes{}, \answerNo{}, or \answerNA{}.
    \item[] Justification: We include a section in appendix discussing our limitations and future works.
    \item[] Guidelines:
    \begin{itemize}
        \item The answer NA means that the paper has no limitation while the answer No means that the paper has limitations, but those are not discussed in the paper. 
        \item The authors are encouraged to create a separate "Limitations" section in their paper.
        \item The paper should point out any strong assumptions and how robust the results are to violations of these assumptions (e.g., independence assumptions, noiseless settings, model well-specification, asymptotic approximations only holding locally). The authors should reflect on how these assumptions might be violated in practice and what the implications would be.
        \item The authors should reflect on the scope of the claims made, e.g., if the approach was only tested on a few datasets or with a few runs. In general, empirical results often depend on implicit assumptions, which should be articulated.
        \item The authors should reflect on the factors that influence the performance of the approach. For example, a facial recognition algorithm may perform poorly when image resolution is low or images are taken in low lighting. Or a speech-to-text system might not be used reliably to provide closed captions for online lectures because it fails to handle technical jargon.
        \item The authors should discuss the computational efficiency of the proposed algorithms and how they scale with dataset size.
        \item If applicable, the authors should discuss possible limitations of their approach to address problems of privacy and fairness.
        \item While the authors might fear that complete honesty about limitations might be used by reviewers as grounds for rejection, a worse outcome might be that reviewers discover limitations that aren't acknowledged in the paper. The authors should use their best judgment and recognize that individual actions in favor of transparency play an important role in developing norms that preserve the integrity of the community. Reviewers will be specifically instructed to not penalize honesty concerning limitations.
    \end{itemize}

\item {\bf Theory assumptions and proofs}
    \item[] Question: For each theoretical result, does the paper provide the full set of assumptions and a complete (and correct) proof?
    \item[] Answer: \answerYes{}{} % Replace by \answerYes{}, \answerNo{}, or \answerNA{}.
    \item[] Justification: For the proposition 1 included in the paper, we have the complete proof in the appendix. 
    \item[] Guidelines:
    \begin{itemize}
        \item The answer NA means that the paper does not include theoretical results. 
        \item All the theorems, formulas, and proofs in the paper should be numbered and cross-referenced.
        \item All assumptions should be clearly stated or referenced in the statement of any theorems.
        \item The proofs can either appear in the main paper or the supplemental material, but if they appear in the supplemental material, the authors are encouraged to provide a short proof sketch to provide intuition. 
        \item Inversely, any informal proof provided in the core of the paper should be complemented by formal proofs provided in appendix or supplemental material.
        \item Theorems and Lemmas that the proof relies upon should be properly referenced. 
    \end{itemize}

    \item {\bf Experimental result reproducibility}
    \item[] Question: Does the paper fully disclose all the information needed to reproduce the main experimental results of the paper to the extent that it affects the main claims and/or conclusions of the paper (regardless of whether the code and data are provided or not)?
    \item[] Answer: \answerYes{} % Replace by \answerYes{}, \answerNo{}, or \answerNA{}.
    \item[] Justification: We detail the experimental settings in the sec~\ref{sec:exp_setting} and more hyperparameter tuning and dataset details in Appendix. 
    \item[] Guidelines:
    \begin{itemize}
        \item The answer NA means that the paper does not include experiments.
        \item If the paper includes experiments, a No answer to this question will not be perceived well by the reviewers: Making the paper reproducible is important, regardless of whether the code and data are provided or not.
        \item If the contribution is a dataset and/or model, the authors should describe the steps taken to make their results reproducible or verifiable. 
        \item Depending on the contribution, reproducibility can be accomplished in various ways. For example, if the contribution is a novel architecture, describing the architecture fully might suffice, or if the contribution is a specific model and empirical evaluation, it may be necessary to either make it possible for others to replicate the model with the same dataset, or provide access to the model. In general. releasing code and data is often one good way to accomplish this, but reproducibility can also be provided via detailed instructions for how to replicate the results, access to a hosted model (e.g., in the case of a large language model), releasing of a model checkpoint, or other means that are appropriate to the research performed.
        \item While NeurIPS does not require releasing code, the conference does require all submissions to provide some reasonable avenue for reproducibility, which may depend on the nature of the contribution. For example
        \begin{enumerate}
            \item If the contribution is primarily a new algorithm, the paper should make it clear how to reproduce that algorithm.
            \item If the contribution is primarily a new model architecture, the paper should describe the architecture clearly and fully.
            \item If the contribution is a new model (e.g., a large language model), then there should either be a way to access this model for reproducing the results or a way to reproduce the model (e.g., with an open-source dataset or instructions for how to construct the dataset).
            \item We recognize that reproducibility may be tricky in some cases, in which case authors are welcome to describe the particular way they provide for reproducibility. In the case of closed-source models, it may be that access to the model is limited in some way (e.g., to registered users), but it should be possible for other researchers to have some path to reproducing or verifying the results.
        \end{enumerate}
    \end{itemize}

\item {\bf Open access to data and code}
    \item[] Question: Does the paper provide open access to the data and code, with sufficient instructions to faithfully reproduce the main experimental results, as described in supplemental material?
    \item[] Answer: \answerYes{} % Replace by \answerYes{}, \answerNo{}, or \answerNA{}.
    \item[] Justification: The access to datasets and codes are provided and we include the detailed settings in the paper main text and appendix. 
    \item[] Guidelines:
    \begin{itemize}
        \item The answer NA means that paper does not include experiments requiring code.
        \item Please see the NeurIPS code and data submission guidelines (\url{https://nips.cc/public/guides/CodeSubmissionPolicy}) for more details.
        \item While we encourage the release of code and data, we understand that this might not be possible, so “No” is an acceptable answer. Papers cannot be rejected simply for not including code, unless this is central to the contribution (e.g., for a new open-source benchmark).
        \item The instructions should contain the exact command and environment needed to run to reproduce the results. See the NeurIPS code and data submission guidelines (\url{https://nips.cc/public/guides/CodeSubmissionPolicy}) for more details.
        \item The authors should provide instructions on data access and preparation, including how to access the raw data, preprocessed data, intermediate data, and generated data, etc.
        \item The authors should provide scripts to reproduce all experimental results for the new proposed method and baselines. If only a subset of experiments are reproducible, they should state which ones are omitted from the script and why.
        \item At submission time, to preserve anonymity, the authors should release anonymized versions (if applicable).
        \item Providing as much information as possible in supplemental material (appended to the paper) is recommended, but including URLs to data and code is permitted.
    \end{itemize}

\item {\bf Experimental setting/details}
    \item[] Question: Does the paper specify all the training and test details (e.g., data splits, hyperparameters, how they were chosen, type of optimizer, etc.) necessary to understand the results?
    \item[] Answer: \answerYes{} % Replace by \answerYes{}, \answerNo{}, or \answerNA{}.
    \item[] Justification: We clarify the dataset splits, hyperparameters and evaluation in the main text and appendix. 
    \item[] Guidelines:
    \begin{itemize}
        \item The answer NA means that the paper does not include experiments.
        \item The experimental setting should be presented in the core of the paper to a level of detail that is necessary to appreciate the results and make sense of them.
        \item The full details can be provided either with the code, in appendix, or as supplemental material.
    \end{itemize}

\item {\bf Experiment statistical significance}
    \item[] Question: Does the paper report error bars suitably and correctly defined or other appropriate information about the statistical significance of the experiments?
    \item[] Answer: \answerYes{} % Replace by \answerYes{}, \answerNo{}, or \answerNA{}.
    \item[] Justification: We include the standard deviation for all the experiments.
    \item[] Guidelines:
    \begin{itemize}
        \item The answer NA means that the paper does not include experiments.
        \item The authors should answer "Yes" if the results are accompanied by error bars, confidence intervals, or statistical significance tests, at least for the experiments that support the main claims of the paper.
        \item The factors of variability that the error bars are capturing should be clearly stated (for example, train/test split, initialization, random drawing of some parameter, or overall run with given experimental conditions).
        \item The method for calculating the error bars should be explained (closed form formula, call to a library function, bootstrap, etc.)
        \item The assumptions made should be given (e.g., Normally distributed errors).
        \item It should be clear whether the error bar is the standard deviation or the standard error of the mean.
        \item It is OK to report 1-sigma error bars, but one should state it. The authors should preferably report a 2-sigma error bar than state that they have a 96\% CI, if the hypothesis of Normality of errors is not verified.
        \item For asymmetric distributions, the authors should be careful not to show in tables or figures symmetric error bars that would yield results that are out of range (e.g. negative error rates).
        \item If error bars are reported in tables or plots, The authors should explain in the text how they were calculated and reference the corresponding figures or tables in the text.
    \end{itemize}

\item {\bf Experiments compute resources}
    \item[] Question: For each experiment, does the paper provide sufficient information on the computer resources (type of compute workers, memory, time of execution) needed to reproduce the experiments?
    \item[] Answer: \answerYes{} % Replace by \answerYes{}, \answerNo{}, or \answerNA{}.
    \item[] Justification: We include the computing resources in the appendix. 
    \item[] Guidelines:
    \begin{itemize}
        \item The answer NA means that the paper does not include experiments.
        \item The paper should indicate the type of compute workers CPU or GPU, internal cluster, or cloud provider, including relevant memory and storage.
        \item The paper should provide the amount of compute required for each of the individual experimental runs as well as estimate the total compute. 
        \item The paper should disclose whether the full research project required more compute than the experiments reported in the paper (e.g., preliminary or failed experiments that didn't make it into the paper). 
    \end{itemize}
    
\item {\bf Code of ethics}
    \item[] Question: Does the research conducted in the paper conform, in every respect, with the NeurIPS Code of Ethics \url{https://neurips.cc/public/EthicsGuidelines}?
    \item[] Answer: \answerYes{} % Replace by \answerYes{}, \answerNo{}, or \answerNA{}.
    \item[] Justification: We follow the code of ethics. 
    \item[] Guidelines:
    \begin{itemize}
        \item The answer NA means that the authors have not reviewed the NeurIPS Code of Ethics.
        \item If the authors answer No, they should explain the special circumstances that require a deviation from the Code of Ethics.
        \item The authors should make sure to preserve anonymity (e.g., if there is a special consideration due to laws or regulations in their jurisdiction).
    \end{itemize}

\item {\bf Broader impacts}
    \item[] Question: Does the paper discuss both potential positive societal impacts and negative societal impacts of the work performed?
    \item[] Answer: \answerYes{} % Replace by \answerYes{}, \answerNo{}, or \answerNA{}.
    \item[] Justification: We include the broader impact discussion in appendix.
    \item[] Guidelines:
    \begin{itemize}
        \item The answer NA means that there is no societal impact of the work performed.
        \item If the authors answer NA or No, they should explain why their work has no societal impact or why the paper does not address societal impact.
        \item Examples of negative societal impacts include potential malicious or unintended uses (e.g., disinformation, generating fake profiles, surveillance), fairness considerations (e.g., deployment of technologies that could make decisions that unfairly impact specific groups), privacy considerations, and security considerations.
        \item The conference expects that many papers will be foundational research and not tied to particular applications, let alone deployments. However, if there is a direct path to any negative applications, the authors should point it out. For example, it is legitimate to point out that an improvement in the quality of generative models could be used to generate deepfakes for disinformation. On the other hand, it is not needed to point out that a generic algorithm for optimizing neural networks could enable people to train models that generate Deepfakes faster.
        \item The authors should consider possible harms that could arise when the technology is being used as intended and functioning correctly, harms that could arise when the technology is being used as intended but gives incorrect results, and harms following from (intentional or unintentional) misuse of the technology.
        \item If there are negative societal impacts, the authors could also discuss possible mitigation strategies (e.g., gated release of models, providing defenses in addition to attacks, mechanisms for monitoring misuse, mechanisms to monitor how a system learns from feedback over time, improving the efficiency and accessibility of ML).
    \end{itemize}
    
\item {\bf Safeguards}
    \item[] Question: Does the paper describe safeguards that have been put in place for responsible release of data or models that have a high risk for misuse (e.g., pretrained language models, image generators, or scraped datasets)?
    \item[] Answer: \answerNA{} % Replace by \answerYes{}, \answerNo{}, or \answerNA{}.
    \item[] Justification: Our paper poses no such risks.
    \item[] Guidelines:
    \begin{itemize}
        \item The answer NA means that the paper poses no such risks.
        \item Released models that have a high risk for misuse or dual-use should be released with necessary safeguards to allow for controlled use of the model, for example by requiring that users adhere to usage guidelines or restrictions to access the model or implementing safety filters. 
        \item Datasets that have been scraped from the Internet could pose safety risks. The authors should describe how they avoided releasing unsafe images.
        \item We recognize that providing effective safeguards is challenging, and many papers do not require this, but we encourage authors to take this into account and make a best faith effort.
    \end{itemize}

\item {\bf Licenses for existing assets}
    \item[] Question: Are the creators or original owners of assets (e.g., code, data, models), used in the paper, properly credited and are the license and terms of use explicitly mentioned and properly respected?
    \item[] Answer: \answerYes{} % Replace by \answerYes{}, \answerNo{}, or \answerNA{}.
    \item[] Justification: We cite the sources of the datasets that are used. 
    \item[] Guidelines:
    \begin{itemize}
        \item The answer NA means that the paper does not use existing assets.
        \item The authors should cite the original paper that produced the code package or dataset.
        \item The authors should state which version of the asset is used and, if possible, include a URL.
        \item The name of the license (e.g., CC-BY 4.0) should be included for each asset.
        \item For scraped data from a particular source (e.g., website), the copyright and terms of service of that source should be provided.
        \item If assets are released, the license, copyright information, and terms of use in the package should be provided. For popular datasets, \url{paperswithcode.com/datasets} has curated licenses for some datasets. Their licensing guide can help determine the license of a dataset.
        \item For existing datasets that are re-packaged, both the original license and the license of the derived asset (if it has changed) should be provided.
        \item If this information is not available online, the authors are encouraged to reach out to the asset's creators.
    \end{itemize}

\item {\bf New assets}
    \item[] Question: Are new assets introduced in the paper well documented and is the documentation provided alongside the assets?
    \item[] Answer: \answerYes{} % Replace by \answerYes{}, \answerNo{}, or \answerNA{}.
    \item[] Justification: The new assets introduced in the paper are well documented. 
    \item[] Guidelines:
    \begin{itemize}
        \item The answer NA means that the paper does not release new assets.
        \item Researchers should communicate the details of the dataset/code/model as part of their submissions via structured templates. This includes details about training, license, limitations, etc. 
        \item The paper should discuss whether and how consent was obtained from people whose asset is used.
        \item At submission time, remember to anonymize your assets (if applicable). You can either create an anonymized URL or include an anonymized zip file.
    \end{itemize}

\item {\bf Crowdsourcing and research with human subjects}
    \item[] Question: For crowdsourcing experiments and research with human subjects, does the paper include the full text of instructions given to participants and screenshots, if applicable, as well as details about compensation (if any)? 
    \item[] Answer: \answerNA{} % Replace by \answerYes{}, \answerNo{}, or \answerNA{}.
    \item[] Justification: The paper does not involve crowdsourcing nor research with human subjects.
    \item[] Guidelines:
    \begin{itemize}
        \item The answer NA means that the paper does not involve crowdsourcing nor research with human subjects.
        \item Including this information in the supplemental material is fine, but if the main contribution of the paper involves human subjects, then as much detail as possible should be included in the main paper. 
        \item According to the NeurIPS Code of Ethics, workers involved in data collection, curation, or other labor should be paid at least the minimum wage in the country of the data collector. 
    \end{itemize}

\item {\bf Institutional review board (IRB) approvals or equivalent for research with human subjects}
    \item[] Question: Does the paper describe potential risks incurred by study participants, whether such risks were disclosed to the subjects, and whether Institutional Review Board (IRB) approvals (or an equivalent approval/review based on the requirements of your country or institution) were obtained?
    \item[] Answer: \answerNA{} % Replace by \answerYes{}, \answerNo{}, or \answerNA{}.
    \item[] Justification: The paper does not involve crowdsourcing nor research with human subjects.
    \item[] Guidelines:
    \begin{itemize}
        \item The answer NA means that the paper does not involve crowdsourcing nor research with human subjects.
        \item Depending on the country in which research is conducted, IRB approval (or equivalent) may be required for any human subjects research. If you obtained IRB approval, you should clearly state this in the paper. 
        \item We recognize that the procedures for this may vary significantly between institutions and locations, and we expect authors to adhere to the NeurIPS Code of Ethics and the guidelines for their institution. 
        \item For initial submissions, do not include any information that would break anonymity (if applicable), such as the institution conducting the review.
    \end{itemize}

\item {\bf Declaration of LLM usage}
    \item[] Question: Does the paper describe the usage of LLMs if it is an important, original, or non-standard component of the core methods in this research? Note that if the LLM is used only for writing, editing, or formatting purposes and does not impact the core methodology, scientific rigorousness, or originality of the research, declaration is not required.
    %this research? 
    \item[] Answer: \answerNA{} % Replace by \answerYes{}, \answerNo{}, or \answerNA{}.
    \item[] Justification: The core method development in this research does not involve LLMs as any important, original, or non-standard components.
    \item[] Guidelines:
    \begin{itemize}
        \item The answer NA means that the core method development in this research does not involve LLMs as any important, original, or non-standard components.
        \item Please refer to our LLM policy (\url{https://neurips.cc/Conferences/2025/LLM}) for what should or should not be described.
    \end{itemize}

\end{enumerate}

\clearpage
\appendix

\section{Proof of proposition 1}
\begin{proposition}
Given $\tilde{\mathcal{L}}(\mtheta) = \mathcal{L}(\mtheta)+\frac{\delta}{2} \|{\mtheta} - \mtheta_{\text{pre}}\|^2_2$, we define the optimal weights as $\tilde{\mtheta}_{\text{ft}} = \mathrm{argmin}_{\mtheta} \tilde{\mathcal{L}}(\mtheta)$ and $\mtheta^*_{\text{ft}} = \mathrm{argmin}_{\mtheta} {\mathcal{L}}(\mtheta)$.
\begin{equation}\label{eq:dwise}
\mathbf{Q}^T\tilde{\mtheta}_{\text{ft}} = (\mathbf{\Lambda}+\delta \mathbf{I})^{-1}\mathbf{\Lambda}\mathbf{Q}^T \mtheta^*_{\text{ft}} + \delta (\mathbf{\Lambda}+\delta \mathbf{I})^{-1}\mathbf{Q}^T\mtheta_{\text{pre}}
\enspace.
\end{equation}
where $\mH$ is the hessian matrix of $\mathcal{L}$ evaluated at $\mtheta^*_{\text{ft}}$ and $\mH = \mQ\mLambda \mQ^T$.
\end{proposition}

\begin{proof}
Based on the quadratic approximation, we can approximate $\mathcal{L}(\mtheta)$ as follows:
\begin{align*}
    \dL(\mtheta) &= \dL(\mtheta^*_{\text{ft}}) + \dL'(\mtheta^*_{\text{ft}})(\mtheta - \mtheta^*_{\text{ft}}) + \frac{1}{2}(\mtheta - \mtheta^*_{\text{ft}})^T\mH(\mtheta - \mtheta^*_{\text{ft}})\\
    &= \dL(\mtheta^*_{\text{ft}}) + \frac{1}{2}(\mtheta - \mtheta^*_{\text{ft}})^T\mH(\mtheta - \mtheta^*_{\text{ft}})
\end{align*}
since $\dL'(\mtheta^*_{\text{ft}}) = 0$ as $\mtheta^*_{\text{ft}}$ is the minimum. 
Then, we add the weight regularization term, such that
\begin{align*}
    \tilde{\dL}(\mtheta) = \dL(\mtheta^*_{\text{ft}}) + \frac{1}{2}(\mtheta - \mtheta^*_{\text{ft}})^T\mH(\mtheta - \mtheta^*_{\text{ft}}) + \delta \|{\mtheta_{\text{ft}} - \mtheta_{\text{pre}}}\|^2_2
\end{align*}
Then, we solve for $\tilde{\mtheta}_{\text{ft}}$ by setting $\nabla \tilde{\dL}(\mtheta) = 0$
\begin{gather*}
    \mH(\tilde{\mtheta}_\text{ft} - \mtheta^*_{\text{ft}}) + \delta(\tilde{\mtheta}_\text{ft} - \mtheta_{\text{pre}}) = 0\\
    (\mH + \delta \mI)\tilde{\mtheta}_{\text{ft}} = \mH\mtheta^*_{\text{ft}} + \delta \mtheta_{\text{pre}}\\
    \tilde{\mtheta}_{\text{ft}} = (\mH+\delta \mI)^{-1}(\mH\mtheta^*_{\text{ft}} + \delta \mtheta_{\text{pre}})\\
    \tilde{\mtheta}_{\text{ft}} = (\mQ\mLambda\mQ^T + \delta\mI)^{-1}(\mQ\mLambda\mQ^T \mtheta^*_{\text{ft}} + \delta \mtheta_{\text{pre}})\\
    \tilde{\mtheta}_{\text{ft}} = (\mQ(\mLambda + \delta \mI)\mQ^T)^{-1}(\mQ\mLambda\mQ^T \mtheta^*_{\text{ft}} + \delta \mtheta_{\text{pre}})\\
    \mathbf{Q}^T\tilde{\mtheta}_{\text{ft}} = (\mathbf{\Lambda}+\delta \mathbf{I})^{-1}\mathbf{\Lambda}\mathbf{Q}^T \mtheta^*_{\text{ft}} + \delta (\mathbf{\Lambda}+\delta \mathbf{I})^{-1}\mathbf{Q}^T\mtheta_{\text{pre}}
\end{gather*}
\end{proof}

\section{Limitations and Future Works}
\label{sec:future}

We acknowledge certain limitations in this current work and highlight potential improvements for future research. Firstly, this study primarily focuses on the \textit{property prediction tasks} of \textit{small molecules} using \textit{2D-graph} based foundation models. 
% while emerging research involves materials datasets or large-molecule data with 3D information. 
Exploring a broader array of foundation models across a wider range of applications--such as covering more areas like DNA, proteins, and materials, addressing various scientific tasks like linker design and chemical reactions, and incorporating diverse data formats like 3D geometric data--is highly worthwhile.
% Exploring more diverse sets of foundation models on a wider range of applications is certainly worthwhile. 
Secondly, although we attempt to include many representative fine-tuning methods from various categories in this study, additional fine-tuning methods from different categories, as discussed in Appendix~\ref{appendix:related}, deserve investigation. For instance, future research could explore whether graph-specific fine-tuning methods offer additional benefits over non-graph fine-tuning approaches across various settings we design.
Thirdly, the method \proj introduced here is an extension and combination of existing methods directly motivated by our benchmark findings for specific fine-tuning scenarios. Future work may involve more thorough exploration into fine-tuning methodology design inspired by our current findings, and aiming to develop approaches effective across a broader range of fine-tuning scenarios.

Regarding the broader impact, we recognize our work can be beneficial to the drug discovery and material science, but people should be aware of the misuse of molecular property prediction tasks to harmful chemical production. 

\section{Additional Discussions of Related Works}\label{appendix:related}
In this section, we additionally discuss more related works about fine-tuning (FT) techniques. Designing advanced fine-tuning  strategies first gained attention in the computer vision (CV) and natural language
processing (NLP) domains, leading to the development of various research directions. We categorize the mainstream approaches into the following groups.

\textbf{Partial model FT.} Numerous studies demonstrate that freezing certain parameters while fine-tuning only specific components of the pre-trained model can help mitigate overfitting during the fine-tuning process~\citep{kirkpatrick2017overcoming,lee2019would,ramasesh2020anatomy,eastwood2021source,evci2022head2toe,cohen2022my}. Specifically, Linear Probing (LP) only trains the additional prediction head during FT. Surgical FT~\citep{leesurgical} selectively fine-tunes a subset of layers based on the specific mechanism of distribution shifts. Partial FT is similar to the concept of parameter efficient fine-tuning methods like LoRA~\citep{hu2022lora}, Prefix tuning~\cite{li2021prefix} and IA3~\cite{liu2022few}. We also include an additional study on LoRA performance in App.~\ref{app:lora}. 

\textbf{Weight-based FT} strategies mainly control the model weights during the FT. Specifically, 
WiSE-FT~\citep{wortsman2022robust}, grounded on the  linear mode connectivity~\citep{frankle2020linear}, linearly interpolates between pre-training parameters and fine-tuning parameters by a mixing coefficient. $L^2$-SP~\citep{xuhong2018explicit} regularizes the fine-tuning model weights using $L^2$ distance to constrain the parameters around pre-trained ones.
REGSL~\citep{li2021improved} further introduces a layer-wise parameter regularization, where the constraint strength gradually reduces
from the top to bottom layers.
MARS-SP~\citep{gouk2020distance} adopts the projected gradient method (PGM) to constrain the fine-tuning model weights
within a small sphere centered on the pre-trained ones. More recently, TPGM~\citep{tian2023trainable} further incorporates trainable weight projection radii constraint for each layer, inspired by MARS-SP, to support layer-wise regularization optimization.

\textbf{Representation-based FT} methods mainly regulate the latent representation space during FT.  Feature-map~\citep{lidelta} adds distance regularization between the latent representations of pre-trained and fine-tuned models to the Full-FT loss. DELTA~\citep{li2019delta} specifically constrains feature maps with the pre-trained activations selected by channel-wise attention.
BSS~\citep{chen2019catastrophic} penalizes the spectral
components corresponding to small singular values that are less transferable to prevent negative transfer. 
\citet{li2020representation} proposes to transfer
representations by encouraging small deviations from the reference one through an regularizer based on optimal transport. Inspired by this,  GTOT-Tuning~\citep{zhang2022fine} presents optimal transport-based fine-tuning framework. 
LP-FT~\citep{kumar2022fine} first performs LP to prediction head while keeping the pre-trained encoder fixed, followed by applying full-FT with the tuned prediction head. 

\textbf{Architecture Refinement.}
Besides the weight and representation based FT, StochNorm~\citep{kou2020stochastic} refactors the widely used Batch Normalization (BN) module and proposes Stochastic
Normalization, to transfer more pre-trained knowledge during the fine-tuning process and mitigate over-fitting.

\textbf{Contrastive-based FT.} As discussed in Sec.~\ref{sec:prelim}, contrastive-based strategies have been widely demonstrated to be effective in the pre-training stage. 
There are other works which explore its effectiveness in the fine-tuning process. \citet{gunel2020supervised},
Bi-tuning~\citep{zhong2020bi}, Core-tuning~\citep{zhang2021unleashing} and
COIN~\citep{pan2023improving} introduce supervised contrastive learning~\citep{khosla2020supervised}
to better leverage the label information in the target datasets
with more discriminative representations as a result. More recently, FLYP~\citep{goyal2023finetune} shows that simply finetuning a classifier via the same contrastive loss as pre-training leads to superior performance in finetuning image-text models.
\citet{oh2024geodesic} fine-tunes the model with contrastive loss on additional hard negative samples, which are generated by
 geodesic multi-modal Mixup,  for robust fine-tuning in multi-modal models.

\textbf{Graph-specific fine-tuning techniques.} 
Apart from the CV and NLP domains, several fine-tuning techniques specifically designed for the Graph-ML domain have recently been proposed. GTOT-Tuning~\citep{zhang2022fine} achieves efficient knowledge transfer from the
pre-trained models by an optimal transport-based FT
framework.
Bridge-Tune~\citep{huang2024measuring} introduces an
intermediate step that bridges pre-training and downstream
tasks by considering the task similarity between them. 
G-tuning~\citep{sun2024fine} tunes
the pre-trained GNN so that it can reconstruct the generative
patterns (graphons) of the downstream graphs. \citet{li2024adaptergnn} leverages expressive adapters for GNNs, to boost adaptation to the downstream tasks.

\section{Pre-training Datasets Detail}
\label{appendix:pt_data}
For self-supervised pre-training,  \textit{Mole-BERT} and \textit{GraphMAE} %, both of which use 5-layer Graph Isomorphism Networks (GINs)~\citep{xu2018powerful} as the encoder and 
are pre-trained over 2M molecules sampled from the ZINC15 database~\citep{sterling2015zinc}, following previous works~\citep{hu2019strategies}. 
 \textit{MoleculeSTM} is initially trained on PubChemSTM, a large multimodal dataset comprising over 280,000 chemical structure–text pairs contructed from the PubChem database~\citep{kim2021pubchem}. 
 
 For supervised pre-training, we use the models from the \textit{Graphium}~\citep{beainitowards} library, which %adopts the same GIN encoder, 
get pre-trained on the Toymix and Largemix datasets provided in this library. The ToyMix dataset~\citep{beainitowards},
totally 2.61M graph-level data points, contains QM9~\citep{ramakrishnan2014quantum}, Tox21~\citep{wu2018moleculenet} and Zinc12K~\citep{dwivedi2023benchmarking}. Specifically, QM9 consists of 19 graph-level quantum properties associated to an energy-minimized 3D conformation of the molecules. Zinc12K is to predict the constrained solubility which is the term $\rm logP - SA - cycle$ (octanol-water partition coefficients, logP,
penalized by the synthetic accessibility score, SA, and number of long cycles, cycle). 
The Largemix dataset, totally 343.4M graph-level data points and 197.7M node-level data points, contains four different datasets with tasks
taken from quantum chemistry (PCQM4M\_G25\_N4), bio-assays (PCBA1328) and transcriptomics (L1000 VCAP and MCF7). Specifically, L1000 VCAP and MCF7 are from the LINCS L1000 database~\citep{subramanian2017next}, which is generated using high-throughput
transcriptomics. VCAP and MCF7 are, respectively, prostate cancer
and human breast cancer cell lines. The PCQM4M\_G25\_N4 dataset is sourced from the PubChemQC project~\citep{nakata2017pubchemqc} that computed
DFT properties on the energy-minimized conformation of 3.8M small molecules from PubChem. The PCBA1328 dataset, originally sourced from \citet{wang2017pubchem}, comprises 1,328 assays and 1.56M molecules and contains information about a molecule's biological activity across various assay settings. The pretraining dataset for GraphGPS is PCQM4Mv2, which is a large-scale molecular dataset containing 3.75M graphs curated from PubChemQC. The task is to regress the HOMO-LUMO gap, a quantum
physical property originally calculated using Density Functional Theory. 

\section{Dataset Statistics}\label{appendix:stat}
The statistics and references of the downstream datasets included in this work are shown in Table~\ref{tab-stat}.
% Please add the following required packages to your document preamble:
% \usepackage{graphicx}
\begin{table}[ht]
\caption{Summary for the molecular datasets used for downstream FT, where ``\textsc{\# Tasks}'' and ``\textsc{\# Molecules}'' denote the number of tasks and molecules of each dataset, respectively.}\label{tab-stat}
\begin{center}
\begin{adjustbox}{width = 1.0\textwidth}\begin{small}
\begin{sc}
{%
\begin{tabular}{ccccc}
\toprule[1.5pt]
Dataset & Evaluation Metrics & Task           & \# Tasks & \# Molecules \\ \midrule
BBBP~\citep{martins2012bayesian}    & AUC                & Classification & 1       & 2,039       \\
Tox21   & AUC                & Classification & 12      & 7,831       \\
ToxCast~\citep{richard2016toxcast} & AUC                & Classification & 617     & 8,576       \\
Sider~\citep{kuhn2016sider}   & AUC                & Classification & 27      & 1,427       \\
ClinTox~\citep{gayvert2016data} & AUC                & Classification & 2       & 1,478       \\
MUV~\citep{rohrer2009maximum}     & AUC                & Classification & 17      & 93,087      \\
HIV~\citet{zaharevitz2015aids}     & AUC                & Classification & 1       & 41,127      \\
Bace~\citep{subramanian2016computational}    & AUC                & Classification & 1       & 1,513       \\ \midrule
Esol~\citep{delaney2004esol}    & RMSE               & Regression     & 1       & 1,128       \\
Lipo~\citep{gaulton2012chembl}     & RMSE               & Regression     & 1       & 4,200       \\
Malaria~\citep{gamo2010thousands} & RMSE               & Regression     & 1       & 9,999       \\
CEP~\citep{hachmann2011harvard}     & RMSE               & Regression     & 1       & 29,978      \\ \bottomrule[1.5pt]
\end{tabular}%
}
\end{sc}
\end{small}
\end{adjustbox}
\end{center}
\end{table}

\section{Details of Experimental Implementation}\label{appendix:experiments}

\textbf{Pre-training Implementations.} 
For self-supervised pre-training, we use the open-source pre-trained checkpoints of Mole-BERT\footnote{\url{https://github.com/junxia97/Mole-BERT}} and GraphMAE\footnote{\url{https://github.com/THUDM/GraphMAE}}. For supervised pre-training, we follow the same training pipeline as proposed in the Graphium\footnote{\url{https://github.com/datamol-io/graphium}}. We drop out the task head MLPs used for supervised pre-training during the downstream fine-tuning process, keeping only the graph encoder component. 
Note that we keep the architecture of the GNN encoder and the graph pooling strategy the same across the three pre-training models. Specifically, we use  a 5-layer Graph Isomorphism Networks (GINs) with 300 hidden dimension and mean pooling as the readout function.

\textbf{Fine-tuning Implementations.} We keep the same training configurations across all the downstream datasets, pre-training models, and fine-tuning strategies, following~\citet{hu2020strategies}. Specifically, for each distinct setting, we fine-tune the pre-training models with 5 random seeds (0-4). We use a batch size of 32 and a dropout rate of 0.5. For each dataset, We train models for 100 epochs and report the test performance when the optimal validation performance is achieved.

\textbf{Hyperparameter Tuning.} We set learning rate to be 0.001 for all the methods and train for 100 epochs. Below is the detailed sets of hyperparameters we tuned for each fine-tuning strategy.
\begin{itemize}
    \item \textit{Surgical FT:} We tune $k$ as which layer in GNN encoder to be updated from $\{0, 1, 2, 3, 4\}$ since our backbone architecture is a 5-layer GIN.

    \item \textit{WiSE-FT:} We tune the mixing coefficient $\alpha$ from $\{0.1, 0.2, 0.3, 0.4, 0.5, 0.6, 0.7, 0.8, 0.9\}$ to control the weight ensemble from pre-trained model and fine-tuned model. A larger $\alpha$ indicates the weights are adopted more from the fine-tuned model.

    \item \textit{$L^2$-SP/ BSS/ Feature-map:} For these three methods that involve an additional regularization term in the loss, we tune the regularization coefficient $\delta$ from $\{1, 0.1, 0.01, 0.001, 0.0001\}$ to control the degree of regularization. For BSS, we follow the original paper and set $k$ to be 1 meaning that we are regularizing the smallest singular value. 

    \item \textit{LP-FT:} We train the LP step before full fine-tuning for 100 epochs and then use the updated prediction head as initilization for the full-FT afterwards for 100 epochs. The training all use the default learning rate 0.001.

    \item \textit{Full FT/ LP:} There is no additional hyperparameter tuning, where we use the default fine-tuning setting.

    \item \textit{\proj:} We tune the initialization of $\alpha_i$ for each layer $i$, where we use the same value to initialize for all layers from $\{0.9, 0.7, 0.5\}$ and the learning rate for validation loss descent from $\{0.001, 0.005, 0.01\}$. We tune $\malpha$ over validation sets over 200 epochs. 
\end{itemize}
Indeed, from the \proj experiments with different starting points of mixing coefficients, the variance of final results is small since it will converge towards the optimal value of mixing coefficients regardless of the initial starting point given a reasonable training time.

\textbf{Computing Resources} The experiments are run on NVIDIA RTX A6000 with 48G memory. 

\section{Further Result Discussions}
\label{appendix:further}

\subsection{Comparisons over pre-trained models}
\label{appendix:pretrain_discussion}

We mainly select the pretrained models based on their pre-training objective divided as supervised and self-supervised learning as discussed in \ref{sec:prelim}. Then, among each category of pretrained models, we diversify with different architecture, model size and detailed training objective or pretraining dataset to discover the effect to the downstream finetuning method selection. 

In the following, we will briefly discuss some more results that are not included in the main text with more pretrained models we tried. Detailed tables can be found in Appendix~\ref{appendix:results}

In general, we found the trend discussed in the main text about the difference of supervised pretrained model and self-supervised pretrained model hold in most cases. Especially, how they prefer over the representation based finetuning techniques or the weight based finetuning techniques remain consistent. However, some small variations may happen regarding the model size and architecture. For instance, for smaller model like 5 layer base GIN model, it is less likely to overfit on fewshot dataset compared to the larger scale graph transformer model. Also, the model expressiveness and capability will vary with different model scale. Therefore, we can compare the rank of different finetuning methods under pretrained models with the same scale, while it is not directly comparable if the model scale is significantly different. 

For instance, both the Graphium model and the GraphGPS demonstrate superior performance from the representation based method like feature-map and BSS compared to other techniques. However, in contrast to the Graphium-Toy model results in the main text that feature-map perform better than BSS especially under the very few shot scenarios. In the GraphGPS results, we find that feature-map tend to be better with more finetuning samples and BSS tends to be better than feature-map in the fewshot cases. This might be due to the variation in the model size that leads to more overfitting, where BSS regularize over noisy feature space through penalizing smaller eigenvalues can be more crucial in reducing overfitting compared to feature-map. Also, we experience a change in pretrained dataset compared to the ToyMix and LargeMix in the Graphium model, where the PCQM4Mv2 is less diversed. This might also cause the degraded performance of feature-map under GraphGPS with fewshot scenario since the learned representation from pretraining might not directly fit the downstream task. When there are more samples available, there might be a larger overlap with the learned representation space. Furthermore, we also observe the worse performance of LP and LP-FT under the larger model which coincides with findings in the main text from Graphium models. 

The conclusions presented in Section 4 generalize well to models pre-trained on large-scale datasets, such as GraphGPS (pre-trained on PCQM4MV2) and Graphium-Large (trained on the LargeMix dataset containing hundreds of millions of labeled molecular graphs). In Section 4.3 and Appendix G.1, we analyze the consistency of trends across all six pre-trained models in our benchmark. Below, we summarize key observations that hold true for models trained on large-scale data:
\begin{itemize}[leftmargin=*]
\item Supervised pre-training on large datasets leads to stronger downstream performance, particularly in few-shot settings. This aligns with our main conclusion in Section 4.2 (Q2), where we compare supervised and self-supervised pre-training. Models like GraphGPS and Graphium-Large consistently outperform self-supervised models such as Mole-BERT and GraphMAE under the same fine-tuning protocols.

\item Representation-based fine-tuning methods (e.g., Feature-map and BSS) remain top-performing strategies for supervised pre-trained models on large-scale datasets, consistent with Finding 3. This trend holds across both classification and regression tasks, and across different dataset splits.

\item Partial fine-tuning methods (e.g., LP and Surgical-FT) continue to underperform in few-shot settings. This observation supports Finding 2, and reflects their tendency to underfit in data-scarce regimes, even when the underlying pre-trained model is strong.

\end{itemize}

% Lastly, note that in the few-shot setting, MoleculeSTM underperforms other evaluated models in self-supervised pretraining. This is mainly because MoleculeSTM’s GNN encoder was specifically pretrained with graph-text alignment to enhance multi-modal tasks like structure-text retrieval. Therefore, the encoder would retain features optimized for cross-modal alignment rather than purely graph structural information.
% Since the downstream tasks in our benchmark do not involve text, the randomly initialized task head struggles to effectively utilize these features with limited data, whereas other models provide a more direct and task-relevant representation, leading to better performance in low-data scenarios.

\subsection{Comparisons over traditional method}
To further understand the effect from foundation model pre-training and fine-tuning process, we include the XGBoost algorithm as a representative for the traditional method. Specifically, we tested the XGBoost algorithm under the Fewshot setting with 50, 100 and 500 samples to see whether it can surpass the performance of foundation model when the training data is scarce. The featurizer being used for the XGBoost model is the Extended Connectivity Circular Fingerprints adopted from the MoleculeNet paper. Then, we keep the exact same splits with the other experiments under random, scaffold and size split. From the result in table~\ref{tab:xgboost}, we can conclude that foundation model result (e.g.) from Mole-BERT surpass the performance in XGBoost on almost all the settings. This indicates the benefit from the pretraining and finetuning framework and the value of our work in selecting the best finetuning technique given different pretraining situation. 
\label{appendix:traditional}
% From \textbf{Q1}, Feature-map proves ineffective, especially in the non-few-shot setting (\textit{c.f.}, Tables~\ref{tab-main1} and \ref{tab-main2}), as it aims to preserve mismatched pre-trained molecular representations. 
% In addition, under few-shot fine-tuning (\textit{c.f.}, Fig.~\ref{fig:1a} and \ref{fig:2a}), BSS consistently outperforms full FT across various splits and prediction tasks. However, under non-few-shot fine-tuning, it shows only marginal benefits or none at all. 
% These findings can also be exlained by the statement \textbf{(1a)}: In the non-few-shot setting, the semantic gap in the representation space between pre-training and downstream tasks \pan{what do you mean by ``the semantic gap in the representation space between pre-training and downstream tasks''? Do not create your words.} is bridged by sufficient training samples, which reduce feature distortion. 
% Also, BSS's compact representation space prevents overfitting, particularly when fine-tuning with limited samples \pan{BSS performance is just so-so, no need to spend much space to explain; Focus more on Feature-map. }.

\subsection{Study on parameter efficient fine-tuning methods}
\label{app:lora}
As an additional study over parameter efficient fine-tuning method, we incorporate the LoRA~\cite{hu2022lora} results for GraphGPS under the scaffold split across three regimes: Fewshot-50, Fewshot-500, and non-fewshot. The results are shown in the table~\ref{table:lora}.

Across both classification and regression tasks, LoRA falls short of full fine-tuning in roughly two-thirds of cases, with the gap widening for regression when more samples are available. This pattern is unsurprising since more challenging tasks and larger downstream datasets generally require updating a greater number of parameters. In the instances where LoRA does outperform full-FT, its results typically lie between standard full fine-tuning and the strongest fine-tuning baselines. Notably, under the Fewshot-50 regression setting, LoRA occasionally matches or even exceeds the best benchmarked fine-tuning methods, highlighting its potential in extremely low-data scenarios.

\subsection{Additional study of \proj on other pretrained models}
We additionally test \proj on other pretrained model like GraphGPS. As shown in Table~\ref{table:dwise-gps}, we report the results of fewshot fine-tuning with 50 samples under scaffold and size splits. These results show that \proj not only significantly improves over WiSE-FT and $L^2$-SP, but also matches or exceeds the best-performing method (TOP) in some cases. This demonstrates that \proj remains effective even under supervised pre-training, including on models like GraphGPS where traditional weight-based methods struggle.

\subsection{Additional findings}
\label{appendix:further_discussion}
$\bullet$ \textbf{Finding 4. LP with pre-trained molecular representations from supervised pre-training surpasses full FT under few-shot fine-tuning, except for size splits.}
% \derek{\textbf{\underline{In contrast to} self-supervised pre-training (Finding x), LP can show advantages  in the few-shot fine-tuning settings with supervised pre-training, but only under  certain types of shifts}}

For few-shot fine-tuning with 50 and 100 samples (\textit{c.f.}, Fig.~\ref{fig:1b} and \ref{fig:2b}), LP surpasses full FT in random and scaffold splits, differing from self-supervised pre-training discussed in \textbf{(1a)}. 
This again supports the claim that directly adopting molecular representations from supervised pre-training retain useful knowledge for downstream tasks. But interestingly, this does not hold for size splits. We believe it is due to the susceptibility of graph level tasks under size shift, as noted in prior OOD studies~\citep{zougess}. Namely, the prediction head tends to overfit to the mapping from representations to output labels with molecules in a specific range of sizes, and thus cannot generalize to OOD molecules of different sizes. 
% \pan{finding 4 should be trimmed and put after finding 5, as partial fine-tuning still performs just so so.}
% This points to the same conclusion above that molecular representations obtained from supervised pre-training comprises more useful knowledge than representations from self-supervised to make LP succeed in few-shot fine-tuning. 

$\bullet$ \textbf{Finding 5.} \textbf{Regulating feature representations brings significant benefits under few-shot fine-tuning but has only a marginal impact in non-few-shot fine-tuning.}

% % $\bullet$ \textbf{Finding 5.} \textbf{Different regularization objectives on molecular representation leads to diverged performances under few-shot fine-tuning and rank differently with classification and regression tasks}

Representation-based methods incorporates additional representation regularization in addition to full FT. BSS aims to eliminate noisy or non-transferable dimensions by regularizing small singular values of representations and Feature-map enforces a close distance of the fine-tuned representations to the pre-trained representations. Since the baseline full FT performs well under non-few-shot settings (\textit{c.f.}, Tables~\ref{tab-main1} and \ref{tab-main2}), and pre-trained molecular representations are unsatisfying as discussed in \textbf{Q1}, having fine-tuned representations to unsatisfying pre-trained representations does not lead to any benefits. While under few-shot fine-tuning, representation regularization prevents overfitting with limited samples on top of full FT to some extend.

\section{Additional Experimental Results}
\label{appendix:results}

In this section, we present complementary baseline results over all pretrained models that are not shown in the main text due to space limit. Table~\ref{tab:model-reference} is a summary of all pre-trained models we test on and their corresponding result tables for reference. Also, a complete table including all few-shot fine-tuning results for \proj (including Fewshot 500 case omitted in the main text) are in Table~\ref{table:new_method_appendix}.

% Specifically, the results on classification tasks in the Fewshot settings over the Mole-BERT (self-supervised pre-training) and Graphium (supervised pre-training) models are in Table~\ref{tab-clf-fewshot-bert-sup} \pan{why not follow the format of table 1 and 2?} \pan{please reorganize the allocation of these tables: having the same format table 1 and 2, use table caption with bold font to directly distinguish the different settings and models.}. The results on regression tasks in the Fewshot settings over the Mole-BERT and Graphium models are in Table~\ref{tab-regs-fewshot-bert-sup}. The results on classification tasks in the Non-Fewshot setting over the Graph-MAE (self-supervised pre-training) model are in Table~\ref{tab-clf-mae-non-fewshot}.
% The results on classification tasks in the Fewshot settings over the Graph-MAE  model are in Table~\ref{tab-clf-mae-fewshot}.
% The results on regression tasks over the Graph-MAE model, including both Non-Fewshot and Fewshot settings, are in Table~\ref{tab-regs-mae}.

% The results of classification datasets over the MoleculeSTM model are in Tables~\ref{table:nonfew_cls_moleSTM}-\ref{table:few500_cls_moleSTM}. The results of regression datasets over the MoleculeSTM model are in Tables~\ref{table:nonfew_reg_moleSTM}-\ref{table:few500_reg_moleSTM}.

% \subsubsection{Hyperparameter analysis}

% You may include other additional sections here.

% More tables

% Hyperparameter Tuning

% More related work discussion

% \input{tab-main1}
% \input{tab-main2}
% \input{tab-main3}

\begin{table*}[t]
\caption{Robust fine-tuning performance on 5 \hlb{\textbf{Classification}}{30}
datasets (AUC metrics) in the \hlr{\textbf{Fewshot}}{30} setting (covering \textsc{Fewshot-50, Fewshot-100, Fewshot-500}), evaluated across 3 dataset splits (\textsc{Random, Scaffold, Size}), over \hlg{\textsc{\textbf{Mole-BERT}}}{30} and \hlg{\textsc{\textbf{Graphium-Toy}}}{30} models.  We \textbf{bold} and \underline{underline} the best and second-best performances in each scenario.}
\label{tab-clf-fewshot-bert-sup}
\begin{center}
\begin{adjustbox}{width = 1\textwidth}
\begin{small}
\begin{sc}
\begin{tabular}{cccccccccc|cccccccc}
\toprule[2pt]
 \multirow{2}{*}{Split}&\multirow{2}{*}{Methods} & \multicolumn{8}{c}{Self-supervised Pre-training (Mole-BERT)} & \multicolumn{8}{c}{Supervised Pre-training (Graphium-Toy)} \\
\cmidrule(lr){3-10}\cmidrule(lr){10-18}        
  &   &   ClinTox &  BBBP  &     Bace    & HIV & Sider & avg & avg-F & avg-R &   ClinTox &  BBBP  &     Bace    & HIV & Sider & avg & avg-F & avg-R  \\
\midrule
\multicolumn{18}{c}{{\textbf{Fewshot-50}}}\\
\midrule
\multirow{8}{*}{Random}

&  Full-FT   &  $74.45 \pm 2.10$  &  $88.56 \pm 0.83$  & $\underline{ 75.80 \pm 0.43 }$ &  $57.41 \pm 0.69$  &  $52.22 \pm 0.48$  & $69.69$ & $69.22$ & $4.40$ &  $70.14\pm 0.52$  &  $77.57\pm 0.01$  &  $80.45\pm 0.00$  &  $63.57\pm 0.00$  &  $55.57\pm 0.00$ & $69.46$ & $70.43$ & $6.00$ \\
 &  LP     &  $77.50 \pm 1.31$  &  $82.05 \pm 0.37$  &  $75.04 \pm 0.58$  &  $53.34 \pm 2.39$  &  $51.40 \pm 0.11$  & $67.87$ & $68.63$ & $6.80$ & $\underline{ 84.09\pm 0.00 }$ &  $81.04\pm 0.00$  & $\underline{ 81.57\pm 0.00 }$ &  $49.05\pm 0.00$  &  $55.62\pm 0.00$ & $70.27$ & $72.74$ & $4.20$ \\
 &  Surgical-FT   & $\mathbf{ 77.91 \pm 1.25 }$ &  $85.41 \pm 0.66$  & $\mathbf{ 75.94 \pm 0.40 }$ &  $57.90 \pm 0.40$  &  $51.99 \pm 0.18$  & $69.83$ & $70.58$ & $3.80$ &  $77.64\pm 0.00$  & $\underline{ 84.99\pm0.00  }$ & $\mathbf{ 81.93\pm0.00  }$ & $\underline{ 64.72\pm0.00  }$ &  $56.40\pm0.00 $ & $73.14$ & $74.76$ & $2.40$ \\
 &  LP-FT   & $\underline{ 77.66 \pm 0.74 }$ & $\mathbf{ 88.99 \pm 0.14 }$ &  $75.18 \pm 0.48$  &  $57.38 \pm 0.37$  &  $51.68 \pm 0.16$  & $70.18$ & $70.07$ & $4.40$ &  $69.84\pm 0.00$  &  $80.15\pm 0.00$  &  $78.64\pm 0.00$  & $\mathbf{ 65.82\pm 0.00 }$ &  $53.56\pm 0.00$  & $69.60$ & $71.43$ & $6.00$ \\
\end{tabular}
\end{sc}
\end{small}
\end{adjustbox}
\end{center}
\end{table*}

\begin{table*}[t]
\caption{Robust fine-tuning performance on 4 \hlb{\textbf{Regression}}{30}
datasets (RMSE metrics) in the \hlr{\textbf{Fewshot}}{30}  setting (covering \textsc{Fewshot-50, Fewshot-100}, and \textsc{Fewshot-500}), evaluated across 3 dataset splits (\textsc{Random, Scaffold, Size}), over \hlg{\textsc{\textbf{Mole-BERT}}}{30} and \hlg{\textsc{\textbf{Graphium-Toy}}}{30} models.  \textsc{Avg-R, Avg-R}$^*$ denote the average rank and the rank based on the average normalized performance over all the datasets for each evavluated method, respectively. Standard deviations across five replicates are shown in parentheses. We \textbf{bold} and \underline{underline} the best and second-best performances in each scenario. }
\label{tab-regs-fewshot-bert-sup}
\begin{center}
\begin{adjustbox}{width = 1\textwidth}
\begin{small}
\begin{sc}
\begin{tabular}{cccccccc|cccccc}
\midrule[2pt]
\multirow{2}{*}{Split} &
  \multirow{2}{*}{Methods} &
  \multicolumn{6}{c}{Self-supervised Pre-training (Mole-BERT)} &
  \multicolumn{6}{c}{Supervised Pre-training (Graphium-Toy)} \\ \cmidrule(lr){3-14} 
 &
   &
  Esol &
  Lipo &
  Malaria &
  CEP &
  Avg-R &
  Avg-R$^*$ &
  Esol &
  Lipo &
  Malaria &
  CEP &
  Avg-R &
  Avg-R$^*$ \\ \midrule
  \multicolumn{14}{c}{{\textbf{Fewshot-50}}}\\
  \midrule
\multirow{8}{*}{Random} &
  Full-FT &
  $1.390 \pm 0.051$ &
  $1.189 \pm 0.016$ &
  $\underline{ 1.276 \pm 0.006 }$ &
  $2.383 \pm 0.046$ &
  $3.50$ &
  4 &
  $1.223 \pm 0.000$ &
  $1.062 \pm 0.000$ &
  $1.284 \pm 0.000$ &
  $2.359 \pm 0.000$ &
  $6.25$ &
  7 \\
 &
  LP &
  $2.654 \pm 0.016$ &
  $1.825 \pm 0.011$ &
  $1.296 \pm 0.005$ &
  $3.736 \pm 0.020$ &
  $8.00$ &
  8 &
  $\mathbf{ 1.085 \pm 0.000 }$ &
  $1.072 \pm 0.000$ &
  $\mathbf{ 1.272 \pm 0.000 }$ &
  $2.571 \pm 0.000$ &
  $4.00$ &
  3 \\
 &
  Surgical-FT &
  $2.647 \pm 0.022$ &
  $1.618 \pm 0.014$ &
  $1.295 \pm 0.004$ &
  $3.596 \pm 0.037$ &
  $7.00$ &
  7 &
  $1.174 \pm0.000 $ &
  $\mathbf{ 1.009 \pm 0.000 }$ &
  $1.277 \pm0.000 $ &
  $2.355 \pm0.000 $ &
  $3.25$ &
  2 \\
 &
  LP-FT &
  $1.422 \pm 0.027$ &
  $1.237 \pm 0.027$ &
  $1.291 \pm 0.005$ &
  $2.296 \pm 0.012$ &
  $5.25$ &
  6 &
  $1.386 \pm 0.000$ &
  $\underline{ 1.019 \pm 0.000 }$ &
  $1.286 \pm 0.000$ &
  $2.287 \pm 0.000$ &
  $5.25$ &
  8 \\
 &
  WiSE-FT &
  $1.384 \pm0.047 $ &
  $1.212 \pm0.020 $ &
  $\underline{ 1.276 \pm0.007 }$ &
  $2.410 \pm0.051 $ &
  $4.25$ &
  5 &
  $1.219 \pm0.000 $ &
  $1.060 \pm0.000 $ &
  $1.280 \pm0.000 $ &
  $2.366 \pm0.000 $ &
  $5.25$ &
  4 \\
 &
  L2-SP &
  $1.372 \pm 0.029$ &
  $1.196 \pm 0.019$ &
  $1.277 \pm 0.006$ &
  $\underline{ 2.280 \pm 0.031 }$ &
  $3.25$ &
  3 &
  $1.147 \pm0.026 $ &
  $1.092 \pm0.001 $ &
  $1.283 \pm0.000 $ &
  $2.312 \pm0.020 $ &
  $5.00$ &
  5 \\
 &
  Feature-map &
  $\mathbf{ 1.329 \pm 0.021 }$ &
  $\mathbf{ 1.164 \pm 0.010 }$ &
  $\mathbf{ 1.271 \pm 0.007 }$ &
  $2.448 \pm 0.010$ &
  $2.25$ &
  1 &
  $\underline{ 1.089 \pm0.001 }$ &
  $1.046 \pm0.000 $ &
  $\underline{ 1.276 \pm0.000 }$ &
  $\mathbf{ 2.191 \pm0.017 }$ &
  $2.00$ &
  1 \\
 &
  BSS &
  $\underline{ 1.365 \pm 0.028 }$ &
  $\underline{ 1.186 \pm 0.017 }$ &
  $1.277 \pm 0.006$ &
  $\mathbf{ 2.275 \pm 0.022 }$ &
  $2.50$ &
  2 &
  $1.175 \pm0.011 $ &
  $1.128 \pm0.035 $ &
  $1.281 \pm0.000 $ &
  $\underline{ 2.262 \pm0.064 }$ &
  $5.00$ &
  6 \\ \midrule
\multirow{8}{*}{Scaffold} &
  Full-FT &
  $\underline{ 1.696 \pm 0.058 }$ &
  $1.124 \pm 0.006$ &
  $1.178 \pm 0.005$ &
  $2.356 \pm 0.033$ &
  $4.25$ &
  5 &
  $1.353 \pm 0.000$ &
  $1.071 \pm 0.000$ &
  $1.168 \pm 0.000$ &
  $2.001 \pm 0.000$ &
  $5.75$ &
  8 \\
 &
  LP &
  $3.754 \pm 0.020$ &
  $1.858 \pm 0.005$ &
  $1.167 \pm 0.002$ &
  $3.849 \pm 0.009$ &
  $7.25$ &
  8 &
  $1.226 \pm0.000 $ &
  $\underline{ 1.013 \pm0.000 }$ &
  $1.166 \pm0.000 $ &
  $2.450 \pm 0.000$ &
  $4.00$ &
  6 \\
 &
  Surgical-FT &
  $3.599 \pm 0.039$ &
  $1.843 \pm 0.006$ &
  $1.167 \pm 0.003$ &
  $3.819 \pm 0.017$ &
  $6.75$ &
  7 &
  $1.239 \pm0.000 $ &
  $1.019 \pm0.000 $ &
  $\mathbf{ 1.162 \pm0.000 }$ &
  $2.083 \pm0.000 $ &
  $3.00$ &
  2 \\
 &
  LP-FT &
  $1.822 \pm 0.014$ &
  $1.134 \pm 0.012$ &
  $1.184 \pm 0.004$ &
  $\mathbf{ 2.292 \pm 0.026 }$ &
  $4.50$ &
  6 &
  $1.283 \pm 0.000$ &
  $1.033 \pm 0.000$ &
  $1.169 \pm 0.000$ &
  $\mathbf{ 1.949 \pm 0.000}$ &
  $4.75$ &
  5 \\
 &
  WiSE-FT &
  $1.842 \pm0.056 $ &
  $1.177 \pm0.009 $ &
  $\underline{ 1.162 \pm0.004 }$ &
  $2.454 \pm0.043 $ &
  $5.00$ &
  4 &
  $1.320 \pm0.000 $ &
  $1.071 \pm0.000 $ &
  $1.168 \pm0.000 $ &
  $\underline{ 1.992 \pm0.000 }$ &
  $5.75$ &
  7 \\
 &
  L2-SP &
  $1.699 \pm 0.049$ &
  $\underline{ 1.086 \pm 0.009 }$ &
  $\underline{ 1.162 \pm 0.002 }$ &
  $2.331 \pm 0.024$ &
  $2.75$ &
  2 &
  $1.273 \pm0.047 $ &
  $1.015 \pm0.007 $ &
  $1.166 \pm0.000 $ &
  $2.132 \pm0.048 $ &
  $6.00$ &
  4 \\
 &
  Feature-map &
  $1.823 \pm 0.028$ &
  $\mathbf{ 1.036 \pm 0.007 }$ &
  $\mathbf{ 1.159 \pm 0.000 }$ &
  $2.425 \pm 0.012$ &
  $3.00$ &
  1 &
  $\mathbf{ 1.213 \pm0.001 }$ &
  $\mathbf{ 0.991 \pm0.000 }$ &
  $\underline{ 1.164 \pm0.000 }$ &
  $2.128 \pm0.006 $ &
  $2.50$ &
  1 \\
 &
  BSS &
  $\mathbf{ 1.680 \pm 0.042 }$ &
  $1.114 \pm 0.008$ &
  $1.165 \pm 0.001$ &
  $\underline{ 2.319 \pm 0.025 }$ &
  $2.50$ &
  3 &
  $\underline{ 1.222 \pm0.012 }$ &
  $1.039 \pm0.000 $ &
  $1.166 \pm0.000 $ &
  $2.121 \pm0.029 $ &
  $4.25$ &
  3 \\ \midrule
\multirow{8}{*}{Size} &
  Full-FT &
  $\underline{2.382 \pm 0.079 }$ &
  $\mathbf{ 1.297 \pm 0.040 }$ &
  $0.929 \pm 0.004$ &
  $2.656 \pm 0.039$ &
  $2.75$ &
  4 &
  $1.441 \pm 0.000$ &
  $1.055 \pm 0.000$ &
  $0.914 \pm 0.000$ &
  $2.329 \pm 0.000$ &
  $5.00$ &
  7 \\
 &
  LP &
  $4.534 \pm 0.021$ &
  $2.157 \pm 0.012$ &
  $0.941 \pm 0.004$ &
  $4.706 \pm 0.022$ &
  $7.75$ &
  8 &
  $1.443 \pm 0.000$ &
  $1.003 \pm 0.000$ &
  $0.936 \pm 0.000$ &
  $2.688 \pm 0.000$ &
  $6.50$ &
  8 \\
 &
  Surgical-FT &
  $4.344 \pm 0.026$ &
  $2.111 \pm 0.021$ &
  $0.943 \pm 0.004$ &
  $4.265 \pm 0.028$ &
  $7.25$ &
  7 &
  $1.469 \pm0.000 $ &
  $1.015 \pm0.000 $ &
  $0.914 \pm0.000 $ &
  $\underline{ 2.313 \pm0.000 }$ &
  $5.25$ &
  5 \\
 &
  LP-FT &
  $2.421 \pm 0.060$ &
  $1.395 \pm 0.018$ &
  $0.939 \pm 0.007$ &
  $\underline{ 2.525 \pm 0.013 }$ &
  $4.50$ &
  6 &
  $\mathbf{ 1.395 \pm 0.000 }$ &
  $0.999 \pm 0.000$ &
  $\underline{ 0.907 \pm 0.000 }$ &
  $2.410 \pm 0.000$ &
  $3.50$ &
  1 \\
 &
  WiSE-FT &
  $2.615 \pm0.072 $ &
  $1.391 \pm0.042 $ &
  $0.929 \pm0.004 $ &
  $2.762 \pm0.053 $ &
  $5.50$ &
  5 &
  $\underline{ 1.411 \pm0.000 }$ &
  $1.071 \pm0.000 $ &
  $\mathbf{ 0.905 \pm0.000 }$ &
  $2.324 \pm0.000 $ &
  $3.50$ &
  4 \\
 &
  L2-SP &
  $2.393 \pm 0.068$ &
  $\underline{ 1.306 \pm 0.037 }$ &
  $\underline{ 0.915 \pm 0.002 }$ &
  $\mathbf{ 2.497 \pm 0.019 }$ &
  $2.00$ &
  2 &
  $1.446 \pm0.055 $ &
  $\underline{ 0.997 \pm0.000 }$ &
  $0.908 \pm0.000 $ &
  $2.340 \pm0.020 $ &
  $4.25$ &
  3 \\
 &
  Feature-map &
  $2.422 \pm 0.021$ &
  $1.327 \pm 0.022$ &
  $\mathbf{ 0.911 \pm 0.002 }$ &
  $2.659 \pm 0.021$ &
  $3.75$ &
  1 &
  $1.415 \pm0.005 $ &
  $\mathbf{ 0.989 \pm0.027 }$ &
  $0.921 \pm0.002 $ &
  $\mathbf{ 2.254 \pm0.001 }$ &
  $3.00$ &
  2 \\
 &
  BSS &
  $\mathbf{ 2.369 \pm 0.075 }$ &
  $1.319 \pm 0.050$ &
  $0.925 \pm 0.003$ &
  $2.563 \pm 0.022$ &
  $2.50$ &
  3 &
  $1.499 \pm0.028 $ &
  $\underline{ 0.997 \pm0.000 }$ &
  $\underline{ 0.907 \pm 0.000 }$ &
  $2.381 \pm 0.006$ &
  $5.00$ &
  6 \\ \midrule
    \multicolumn{14}{c}{{\textbf{Fewshot-100}}}\\
  \midrule
\multirow{8}{*}{Random} &
  Full-FT &
  $\underline{ 1.141 \pm 0.030 }$ &
  $\underline{ 1.141 \pm 0.023 }$ &
  $\mathbf{ 1.256 \pm 0.006 }$ &
  $2.150 \pm 0.021$ &
  $2.00$ &
  1 &
  $1.191 \pm 0.000$ &
  $1.103 \pm 0.000$ &
  $\underline{ 1.258 \pm 0.000 }$ &
  $2.076 \pm 0.118$ &
  $5.25$ &
  4 \\
 &
  LP &
  $2.273 \pm 0.029$ &
  $1.569 \pm 0.008$ &
  $1.280 \pm 0.003$ &
  $3.235 \pm 0.019$ &
  $8.00$ &
  8 &
  $1.066 \pm 0.000$ &
  $1.045 \pm 0.000$ &
  $1.267 \pm 0.000$ &
  $2.383 \pm 0.000$ &
  $4.75$ &
  5 \\
 &
  Surgical-FT &
  $1.953 \pm 0.039$ &
  $1.281 \pm 0.020$ &
  $1.270 \pm 0.006$ &
  $3.019 \pm 0.047$ &
  $6.75$ &
  7 &
  $1.075 \pm 0.000$ &
  $\underline{ 1.030 \pm 0.000 }$ &
  $1.266 \pm0.000 $ &
  $\mathbf{ 1.935 \pm 0.000}$ &
  $2.75$ &
  2 \\
 &
  LP-FT &
  $1.244 \pm 0.057$ &
  $1.147 \pm 0.018$ &
  $1.277 \pm 0.003$ &
  $2.156 \pm 0.019$ &
  $5.25$ &
  6 &
  $1.689 \pm 0.000$ &
  $1.097 \pm 0.000$ &
  $1.273 \pm 0.000$ &
  $2.044 \pm 0.015$ &
  $6.25$ &
  8 \\
 &
  WiSE-FT &
  $1.189 \pm0.030 $ &
  $1.142 \pm0.025 $ &
  $\mathbf{ 1.256 \pm0.006 }$ &
  $2.211 \pm0.028 $ &
  $3.50$ &
  2 &
  $1.131 \pm0.000 $ &
  $1.078 \pm0.000 $ &
  $\mathbf{ 1.256 \pm0.000 }$ &
  $2.001 \pm0.071 $ &
  $3.75$ &
  3 \\
 &
  L2-SP &
  $1.161 \pm 0.016$ &
  $1.149 \pm 0.007$ &
  $1.260 \pm 0.004$ &
  $\mathbf{ 2.131 \pm 0.014 }$ &
  $3.25$ &
  4 &
  $1.098 \pm0.012 $ &
  $1.077 \pm0.001 $ &
  $1.270 \pm0.001 $ &
  $2.261 \pm0.008 $ &
  $5.25$ &
  6 \\
 &
  Feature-map &
  $\mathbf{ 1.120 \pm 0.038 }$ &
  $\mathbf{ 1.139 \pm 0.017 }$ &
  $1.266 \pm 0.004$ &
  $2.283 \pm 0.011$ &
  $3.25$ &
  5 &
  $\mathbf{ 0.995 \pm0.018 }$ &
  $\mathbf{ 1.025 \pm0.000 }$ &
  $\underline{ 1.258 \pm0.003 }$ &
  $\underline{ 1.937 \pm0.023 }$ &
  $1.75$ &
  1 \\
 &
  BSS &
  $1.199 \pm 0.033$ &
  $1.149 \pm 0.023$ &
  $1.259 \pm 0.006$ &
  $\underline{ 2.132 \pm 0.019 }$ &
  $4.00$ &
  3 &
  $\underline{ 1.055 \pm0.009 }$ &
  $1.136 \pm0.000 $ &
  $1.274 \pm0.000 $ &
  $2.269 \pm0.010 $ &
  $6.25$ &
  7 \\ \midrule
\multirow{8}{*}{Scaffold} &
  Full-FT &
  $\mathbf{ 1.436 \pm 0.054 }$ &
  $1.026 \pm 0.009$ &
  $1.160 \pm 0.011$ &
  $\underline{ 2.198 \pm 0.034 }$ &
  $3.25$ &
  4 &
  $1.111 \pm 0.000$ &
  $1.037 \pm 0.000$ &
  $1.172 \pm 0.000$ &
  $\underline{ 1.965 \pm 0.023}$ &
  $5.00$ &
  6 \\
 &
  LP &
  $3.255 \pm 0.025$ &
  $1.503 \pm 0.008$ &
  $1.154 \pm 0.003$ &
  $3.350 \pm 0.007$ &
  $7.00$ &
  8 &
  $1.228 \pm 0.000$ &
  $\mathbf{ 0.960 \pm 0.000 }$ &
  $1.162 \pm 0.000$ &
  $2.423 \pm 0.000$ &
  $4.50$ &
  5 \\
 &
  Surgical-FT &
  $2.587 \pm 0.076$ &
  $1.192 \pm 0.015$ &
  $1.156 \pm 0.003$ &
  $2.914 \pm 0.066$ &
  $6.50$ &
  7 &
  $\mathbf{ 1.087 \pm0.000 }$ &
  $\underline{ 0.966 \pm 0.000 }$ &
  $\mathbf{ 1.156 \pm0.000 }$ &
  $\mathbf{ 1.959 \pm0.000 }$ &
  $1.25$ &
  1 \\
 &
  LP-FT &
  $1.544 \pm 0.042$ &
  $1.010 \pm 0.011$ &
  $1.163 \pm 0.004$ &
  $\mathbf{ 2.187 \pm 0.034 }$ &
  $4.00$ &
  6 &
  $1.111 \pm 0.000$ &
  $0.984 \pm 0.000$ &
  $1.173 \pm 0.000$ &
  $2.149 \pm 0.012$ &
  $5.25$ &
  4 \\
 &
  WiSE-FT &
  $1.544 \pm0.063 $ &
  $1.041 \pm 0.017 $ &
  $\underline{ 1.151 \pm0.007 }$ &
  $2.301 \pm0.042 $ &
  $4.50$ &
  3 &
  $\underline{ 1.110 \pm0.000 }$ &
  $1.027 \pm0.000 $ &
  $1.169 \pm0.000 $ &
  $2.013 \pm0.049 $ &
  $4.25$ &
  3 \\
 &
  L2-SP &
  $1.473 \pm 0.009$ &
  $\underline{ 0.961 \pm 0.003 }$ &
  $1.153 \pm 0.002$ &
  $2.201 \pm 0.038$ &
  $2.75$ &
  2 &
  $1.252 \pm0.021 $ &
  $0.994 \pm0.013 $ &
  $1.163 \pm0.000 $ &
  $2.367 \pm0.052 $ &
  $5.75$ &
  7 \\
 &
  Feature-map &
  $1.677 \pm 0.020$ &
  $\mathbf{ 0.937 \pm 0.008 }$ &
  $\mathbf{ 1.149 \pm 0.003 }$ &
  $2.356 \pm 0.018$ &
  $3.50$ &
  1 &
  $1.158 \pm0.020 $ &
  $\underline{ 0.966 \pm0.010 }$ &
  $\underline{ 1.161 \pm0.000 }$ &
  $2.024 \pm0.019 $ &
  $3.50$ &
  2 \\
 &
  BSS &
  $\underline{ 1.463 \pm 0.008 }$ &
  $1.040 \pm 0.018$ &
  $1.160 \pm 0.006$ &
  $2.210 \pm 0.018$ &
  $4.50$ &
  5 &
  $1.253 \pm0.027 $ &
  $1.033 \pm 0.015$ &
  $1.167 \pm0.000 $ &
  $2.333 \pm0.022 $ &
  $6.50$ &
  8 \\ \midrule
\multirow{8}{*}{Size} &
  Full-FT &
  $1.889 \pm 0.065$ &
  $1.077 \pm 0.028$ &
  $0.918 \pm 0.005$ &
  $\underline{ 2.425 \pm 0.024 }$ &
  $4.00$ &
  3 &
  $1.411 \pm 0.000$ &
  $\mathbf{ 0.962 \pm 0.000 }$ &
  $0.921 \pm 0.006$ &
  $2.328 \pm 0.015$ &
  $4.75$ &
  5 \\
 &
  LP &
  $3.851 \pm 0.033$ &
  $1.676 \pm 0.025$ &
  $0.911 \pm 0.003$ &
  $4.115 \pm 0.038$ &
  $6.75$ &
  8 &
  $\underline{ 1.253 \pm 0.000 }$ &
  $0.981 \pm 0.000$ &
  $0.924 \pm 0.000$ &
  $2.635 \pm 0.000$ &
  $6.00$ &
  8 \\
 &
  Surgical-FT &
  $3.237 \pm 0.085$ &
  $1.374 \pm 0.031$ &
  $0.912 \pm 0.002$ &
  $3.174 \pm 0.048$ &
  $6.25$ &
  7 &
  $1.329 \pm0.000 $ &
  $0.965 \pm0.000 $ &
  $0.910 \pm0.000 $ &
  $\underline{ 2.283 \pm0.000 }$ &
  $3.25$ &
  2 \\
 &
  LP-FT &
  $1.831 \pm 0.066$ &
  $1.085 \pm 0.014$ &
  $0.920 \pm 0.008$ &
  $2.468 \pm 0.021$ &
  $4.75$ &
  4 &
  $\mathbf{ 1.242 \pm 0.000 }$ &
  $\mathbf{ 0.962 \pm 0.000 }$ &
  $0.912 \pm 0.000$ &
  $2.375 \pm 0.000$ &
  $3.50$ &
  1 \\
 &
  WiSE-FT &
  $2.216 \pm0.056 $ &
  $1.124 \pm0.031 $ &
  $0.917 \pm0.004 $ &
  $2.543 \pm0.027 $ &
  $5.75$ &
  5 &
  $1.398 \pm0.000 $ &
  $0.963 \pm0.000 $ &
  $0.907 \pm0.002 $ &
  $2.319 \pm0.014 $ &
  $3.75$ &
  4 \\
 &
  L2-SP &
  $\mathbf{ 1.731 \pm 0.071 }$ &
  $\mathbf{ 1.025 \pm 0.028 }$ &
  $\underline{ 0.905 \pm 0.002 }$ &
  $\mathbf{ 2.424 \pm 0.024 }$ &
  $1.25$ &
  1 &
  $1.418 \pm0.035 $ &
  $0.998 \pm0.038 $ &
  $\mathbf{ 0.906 \pm0.000 }$ &
  $2.436 \pm0.072 $ &
  $5.50$ &
  6 \\
 &
  Feature-map &
  $2.135 \pm 0.077$ &
  $\underline{ 1.049 \pm 0.013 }$ &
  $\mathbf{ 0.898 \pm 0.003 }$ &
  $2.500 \pm 0.017$ &
  $3.25$ &
  2 &
  $1.335 \pm0.005 $ &
  $0.967 \pm0.008 $ &
  $0.911 \pm0.001 $ &
  $\mathbf{ 2.265 \pm0.020 }$ &
  $3.75$ &
  3 \\
 &
  BSS &
  $\underline{ 1.734 \pm 0.060 }$ &
  $1.073 \pm 0.024$ &
  $0.931 \pm 0.008$ &
  $2.439 \pm 0.015$ &
  $4.00$ &
  6 &
  $1.387 \pm0.039 $ &
  $0.998 \pm0.006 $ &
  $\mathbf{ 0.906 \pm0.000 }$ &
  $2.518 \pm0.137 $ &
  $5.50$ &
  7 \\ \midrule
  \multicolumn{14}{c}{{\textbf{Fewshot-500}}}\\
  \midrule
\multirow{8}{*}{Random} &
  Full-FT &
  $\underline{ 0.883 \pm 0.032 }$ &
  $0.817 \pm 0.012$ &
  $1.194 \pm 0.003$ &
  $\underline{ 1.891 \pm 0.026 }$ &
  $2.50$ &
  3 &
  $0.753 \pm 0.000$ &
  $0.842 \pm 0.000$ &
  $1.221 \pm 0.012$ &
  $1.806 \pm 0.005$ &
  $4.75$ &
  4 \\
 &
  LP &
  $1.274 \pm 0.011$ &
  $1.036 \pm 0.004$ &
  $1.216 \pm 0.002$ &
  $2.285 \pm 0.004$ &
  $8.00$ &
  8 &
  $1.007 \pm 0.000$ &
  $0.972 \pm 0.000$ &
  $1.223 \pm 0.000$ &
  $2.117 \pm 0.000$ &
  $7.25$ &
  8 \\
 &
  Surgical-FT &
  $0.961 \pm 0.013$ &
  $0.888 \pm 0.005$ &
  $1.201 \pm 0.005$ &
  $1.962 \pm 0.009$ &
  $5.75$ &
  6 &
  $0.748 \pm0.000 $ &
  $\mathbf{ 0.825 \pm0.000 }$ &
  $\underline{ 1.210 \pm0.000 }$ &
  $1.795 \pm0.000 $ &
  $3.00$ &
  2 \\
 &
  LP-FT &
  $0.884 \pm 0.035$ &
  $0.842 \pm 0.013$ &
  $1.215 \pm 0.002$ &
  $1.904 \pm 0.011$ &
  $4.75$ &
  5 &
  $\mathbf{ 0.697 \pm 0.000 }$ &
  $\underline{ 0.835 \pm 0.016 }$ &
  $1.220 \pm 0.008$ &
  $\underline{ 1.794 \pm 0.004}$ &
  $2.00$ &
  3 \\
 &
  WiSE-FT &
  $0.995 \pm0.010 $ &
  $0.855 \pm0.011 $ &
  $\underline{ 1.193 \pm 0.003 }$ &
  $1.893 \pm0.021 $ &
  $4.00$ &
  4 &
  $0.742 \pm0.000 $ &
  $0.852 \pm0.001 $ &
  $1.228 \pm0.004 $ &
  $1.809 \pm0.006 $ &
  $5.25$ &
  5 \\
 &
  L2-SP &
  $\mathbf{ 0.878 \pm 0.026 }$ &
  $\mathbf{ 0.806 \pm 0.007 }$ &
  $\mathbf{ 1.192 \pm 0.004 }$ &
  $1.893 \pm 0.018$ &
  $1.75$ &
  1 &
  $0.741 \pm0.029 $ &
  $0.907 \pm0.020 $ &
  $1.243 \pm0.006 $ &
  $1.822 \pm 0.003$ &
  $6.00$ &
  7 \\
 &
  Feature-map &
  $1.057 \pm 0.008$ &
  $0.894 \pm 0.009$ &
  $1.196 \pm 0.002$ &
  $2.019 \pm 0.004$ &
  $6.50$ &
  7 &
  $\underline{ 0.706 \pm0.005 }$ &
  $0.840 \pm0.013 $ &
  $\mathbf{ 1.200 \pm0.014 }$ &
  $\mathbf{ 1.773 \pm0.008 }$ &
  $1.75$ &
  1 \\
 &
  BSS &
  $0.886 \pm 0.010$ &
  $\underline{ 0.809 \pm 0.005 }$ &
  $1.194 \pm 0.006$ &
  $\mathbf{ 1.862 \pm 0.010 }$ &
  $2.75$ &
  2 &
  $0.715 \pm0.024 $ &
  $0.892 \pm0.014 $ &
  $1.248 \pm0.006 $ &
  $1.824 \pm 0.006$ &
  $6.00$ &
  6 \\ \midrule
\multirow{8}{*}{Scaffold} &
  Full-FT &
  $1.196 \pm 0.013$ &
  $0.819 \pm 0.009$ &
  $1.137 \pm 0.016$ &
  $1.892 \pm 0.017$ &
  $4.25$ &
  4 &
  $0.956 \pm 0.000$ &
  $0.888 \pm 0.011$ &
  $1.149 \pm 0.014$ &
  $\mathbf{ 1.787 \pm 0.020}$ &
  $4.50$ &
  5 \\
 &
  LP &
  $1.867 \pm 0.006$ &
  $0.937 \pm 0.004$ &
  $1.140 \pm 0.002$ &
  $2.338 \pm 0.005$ &
  $7.75$ &
  8 &
  $1.006 \pm 0.000$ &
  $0.921 \pm 0.000$ &
  $1.162 \pm 0.000$ &
  $2.183 \pm 0.000$ &
  $8.00$ &
  8 \\
 &
  Surgical-FT &
  $1.221 \pm 0.011$ &
  $0.883 \pm 0.010$ &
  $1.130 \pm 0.005$ &
  $1.953 \pm 0.007$ &
  $5.75$ &
  6 &
  $0.955 \pm0.000 $ &
  $0.887 \pm0.000 $ &
  $1.138 \pm0.000 $ &
  $\mathbf{ 1.787 \pm0.000 }$ &
  $3.75$ &
  3 \\
 &
  LP-FT &
  $\mathbf{ 1.112 \pm 0.015 }$ &
  $\mathbf{ 0.802 \pm 0.003 }$ &
  $1.153 \pm 0.005$ &
  $1.895 \pm 0.013$ &
  $3.50$ &
  5 &
  $\underline{ 0.951 \pm 0.000 }$ &
  $0.883 \pm 0.025$ &
  $1.143 \pm 0.000$ &
  $1.791 \pm 0.008$ &
  $3.50$ &
  4 \\
 &
  WiSE-FT &
  $1.388 \pm 0.023 $ &
  $0.834 \pm0.012 $ &
  $\mathbf{ 1.114 \pm0.002 }$ &
  $1.936 \pm0.037 $ &
  $4.25$ &
  3 &
  $\mathbf{ 0.947 \pm0.000 }$ &
  $0.893 \pm0.007 $ &
  $1.134 \pm0.011 $ &
  $1.800 \pm0.006 $ &
  $4.00$ &
  2 \\
 &
  L2-SP &
  $\underline{ 1.163 \pm 0.026 }$ &
  $\underline{ 0.813 \pm 0.010 }$ &
  $1.126 \pm 0.011$ &
  $\underline{ 1.885 \pm 0.011 }$ &
  $2.50$ &
  2 &
  $0.991 \pm0.018 $ &
  $\underline{ 0.878 \pm0.012 }$ &
  $\underline{ 1.128 \pm 0.002 }$ &
  $2.017 \pm0.179 $ &
  $4.50$ &
  7 \\
 &
  Feature-map &
  $1.495 \pm 0.016$ &
  $0.863 \pm 0.005$ &
  $\underline{ 1.118 \pm 0.001 }$ &
  $2.008 \pm 0.010$ &
  $5.50$ &
  7 &
  $0.966 \pm0.014 $ &
  $\mathbf{ 0.826 \pm0.017 }$ &
  $1.136 \pm0.003 $ &
  $1.792 \pm 0.011$ &
  $3.50$ &
  1 \\
 &
  BSS &
  $1.188 \pm 0.026$ &
  $0.814 \pm 0.007$ &
  $1.123 \pm 0.005$ &
  $\mathbf{ 1.881 \pm 0.010 }$ &
  $2.50$ &
  1 &
  $0.977 \pm0.021 $ &
  $0.885 \pm0.014 $ &
  $\mathbf{ 1.126 \pm0.007 }$ &
  $1.949 \pm0.127 $ &
  $4.25$ &
  6 \\ \midrule
\multirow{8}{*}{Size} &
  Full-FT &
  $1.692 \pm 0.070$ &
  $0.838 \pm 0.023$ &
  $0.922 \pm 0.013$ &
  $\underline{ 2.364 \pm 0.030 }$ &
  $4.00$ &
  3 &
  $1.115 \pm 0.019$ &
  $0.848 \pm 0.038$ &
  $0.915 \pm 0.000$ &
  $2.230 \pm 0.009$ &
  $5.25$ &
  5 \\
 &
  LP &
  $2.290 \pm 0.017$ &
  $1.039 \pm 0.005$ &
  $0.908 \pm 0.002$ &
  $2.749 \pm 0.018$ &
  $6.75$ &
  8 &
  $\mathbf{ 1.073 \pm 0.000 }$ &
  $0.871 \pm 0.000$ &
  $0.904 \pm 0.000$ &
  $2.435 \pm 0.000$ &
  $5.25$ &
  8 \\
 &
  Surgical-FT &
  $1.928 \pm 0.039$ &
  $0.895 \pm 0.007$ &
  $0.919 \pm 0.007$ &
  $2.397 \pm 0.014$ &
  $5.50$ &
  6 &
  $1.094 \pm0.000 $ &
  $\underline{ 0.807 \pm0.000 }$ &
  $0.904 \pm0.000 $ &
  $\mathbf{ 2.200 \pm0.000 }$ &
  $2.75$ &
  1 \\
 &
  LP-FT &
  $1.674 \pm 0.030$ &
  $\mathbf{ 0.803 \pm 0.006 }$ &
  $0.954 \pm 0.011$ &
  $\mathbf{ 2.328 \pm 0.017 }$ &
  $3.25$ &
  5 &
  $\underline{ 1.081 \pm 0.024 }$ &
  $0.842 \pm 0.021$ &
  $0.925 \pm 0.000$ &
  $2.280 \pm 0.000$ &
  $5.25$ &
  7 \\
 &
  WiSE-FT &
  $2.071 \pm0.078 $ &
  $0.902 \pm0.016 $ &
  $0.912 \pm0.003 $ &
  $2.379 \pm 0.086 $ &
  $5.75$ &
  7 &
  $1.116 \pm0.023 $ &
  $\mathbf{ 0.805 \pm0.015 }$ &
  $0.907 \pm0.001 $ &
  $2.228 \pm0.010 $ &
  $4.00$ &
  2 \\
 &
  L2-SP &
  $\mathbf{ 1.629 \pm 0.084 }$ &
  $0.821 \pm 0.011$ &
  $\underline{ 0.904 \pm 0.003 }$ &
  $2.368 \pm 0.013$ &
  $2.50$ &
  1 &
  $1.183 \pm0.055 $ &
  $0.853 \pm0.031 $ &
  $0.903 \pm0.004 $ &
  $2.227 \pm0.038 $ &
  $5.00$ &
  6 \\
 &
  Feature-map &
  $1.963 \pm 0.035$ &
  $0.910 \pm 0.009$ &
  $\mathbf{ 0.895 \pm 0.002 }$ &
  $2.366 \pm 0.006$ &
  $4.25$ &
  4 &
  $1.193 \pm0.058 $ &
  $0.850 \pm0.021 $ &
  $\underline{ 0.901 \pm0.025 }$ &
  $\underline{ 2.203 \pm0.023 }$ &
  $4.50$ &
  4 \\
 &
  BSS &
  $\underline{ 1.630 \pm 0.035 }$ &
  $\underline{ 0.818 \pm 0.005 }$ &
  $0.925 \pm 0.019$ &
  $2.370 \pm 0.013$ &
  $4.00$ &
  2 &
  $1.142 \pm0.049$ &
  $0.834 \pm0.018 $ &
  $\mathbf{ 0.900 \pm 0.003 }$ &
  $2.245 \pm0.027 $ &
  $4.00$ &
  3 \\ \midrule[2pt]
\end{tabular}%
\end{sc}
\end{small}
\end{adjustbox}
\end{center}

\end{table*}

\begin{table*}[t]
\caption{Robust fine-tuning performance on 8 \hlb{\textbf{Classification}}{30}
datasets (AUC metrics) in the \hlr{\textbf{Non-Fewshot}}{30} setting, evaluated across 3 dataset splits (\textsc{Random, Scaffold, Size}), over \hlg{\textsc{\textbf{MoleculeSTM}}}{30} and \hlg{\textsc{\textbf{Graphium-Large}}}{30} models. \textsc{Avg, Avg-F, Avg-R} denote the average AUC, average AUC without max and min values, and average rank over all the  datasets for each method, respectively. Standard deviations across five replicates are shown in parentheses. We \textbf{bold} and \underline{underline} the best and second-best performances in each scenario. 
}\label{tab-stm-large-clf-non}
\begin{center}
\begin{adjustbox}{width = 1.0\textwidth}
\begin{small}
\begin{sc}
\begin{tabular}{ccccccccccccc}

\toprule[2pt]

 Split & Methods  &   ClinTox &  BBBP  &     Bace    & HIV     & MUV & Sider &  Tox21  &   ToxCast & Avg & Avg-F & Avg-R\\ \midrule
\multicolumn{13}{c}{{\textbf{Self-supervised Pre-training (MoleculeSTM)}}}  \\ \midrule
\multirow{8}{*}{Random}  
% & Full-FT      & $89.90 \pm 1.49$ & $93.43 \pm 0.99$ & $89.82 \pm 1.08$ & $84.72 \pm 1.11$ & $77.82 \pm 3.46$ & $62.12 \pm 1.15$ & $82.49 \pm 0.41$ & $72.95 \pm 0.31$ \\  
% & LP           & $74.32 \pm 1.90$ & $84.76 \pm 0.29$ & $74.85 \pm 0.27$ & $74.15 \pm 0.69$ & $76.86 \pm 1.07$ & $59.69 \pm 0.24$ & $73.72 \pm 0.20$ & $66.19 \pm 0.14$ \\  
% & Surgical-FT  & $86.04 \pm 0.89$ & $93.68 \pm 0.51$ & $89.99 \pm 0.46$ & $85.68 \pm 0.84$ & $79.59 \pm 2.47$ & $63.64 \pm 0.78$ & $81.84 \pm 0.66$ & $71.83 \pm 0.55$ \\  
% & LP-FT        & $86.39 \pm 1.85$ & $93.72 \pm 0.93$ & $89.82 \pm 0.57$ & $84.17 \pm 1.41$ & $76.87 \pm 2.38$ & $62.19 \pm 1.00$ & $82.54 \pm 0.51$ & $72.19 \pm 0.52$ \\  
% & WiSE-FT      & $90.35 \pm 1.26$ & $92.93 \pm 0.80$ & $90.41 \pm 0.86$ & $84.38 \pm 1.05$ & $77.23 \pm 3.08$ & $62.17 \pm 1.25$ & $82.67 \pm 0.32$ & $73.08 \pm 0.32$ \\  
% & $L^2$-SP     & $89.69 \pm 1.39$ & $93.77 \pm 0.37$ & $89.21 \pm 0.92$ & $81.94 \pm 1.20$ & $50.21\pm4.41$               & $61.07 \pm 1.22$ & $82.97 \pm 0.39$ & $71.02 \pm 0.57$ \\  
% & Feature-map  & $79.93 \pm 1.54$ & $90.59 \pm 0.39$ & $83.69 \pm 0.24$ & $77.66 \pm 0.46$ & $80.03 \pm 1.01$ & $59.93 \pm 0.14$ & $75.32 \pm 0.19$ & $67.51 \pm 0.30$ \\  
% & BSS          & $90.17 \pm 2.84$ & $94.16 \pm 0.55$ & $89.74 \pm 1.12$ & $83.96 \pm 1.29$ & $76.64\pm1.29$               & $61.87 \pm 0.69$ & $83.26 \pm 0.57$ & $74.55 \pm 0.31$ \\
&  Full-FT       &  $89.90 \pm 1.49$  &  $93.43 \pm 0.99$  &  $89.82 \pm 1.08$  & $\underline{ 84.72 \pm 1.11 }$ &  $77.82 \pm 3.46$  &  $62.12 \pm 1.15$  &  $82.49 \pm 0.41$  &  $72.95 \pm 0.31$  & $81.66$ & $82.95$ & $3.62$ \\
   &  LP            &  $74.32 \pm 1.90$  &  $84.76 \pm 0.29$  &  $74.85 \pm 0.27$  &  $74.15 \pm 0.69$  &  $76.86 \pm 1.07$  &  $59.69 \pm 0.24$  &  $73.72 \pm 0.20$  &  $66.19 \pm 0.14$  & $73.07$ & $73.35$ & $7.75$ \\
   &  Surgical-FT   &  $86.04 \pm 0.89$  &  $93.68 \pm 0.51$  & $\underline{ 89.99 \pm 0.46 }$ & $\mathbf{ 85.68 \pm 0.84 }$ & $\underline{ 79.59 \pm 2.47 }$ & $\mathbf{ 63.64 \pm 0.78 }$ &  $81.84 \pm 0.66$  &  $71.83 \pm 0.55$  & $81.54$ & $82.50$ & $3.38$ \\
   &  LP-FT         &  $86.39 \pm 1.85$  &  $93.72 \pm 0.93$  &  $89.82 \pm 0.57$  &  $84.17 \pm 1.41$  &  $76.87 \pm 2.38$  & $\underline{ 62.19 \pm 1.00 }$ &  $82.54 \pm 0.51$  &  $72.19 \pm 0.52$  & $80.99$ & $82.00$ & $3.75$ \\
   &  WiSE-FT       & $\mathbf{ 90.35 \pm 1.26 }$ &  $92.93 \pm 0.80$  & $\mathbf{ 90.41 \pm 0.86 }$ &  $84.38 \pm 1.05$  &  $77.23 \pm 3.08$  &  $62.17 \pm 1.25$  &  $82.67 \pm 0.32$  & $\underline{ 73.08 \pm 0.32 }$ & $81.65$ & $83.02$ & $2.88$ \\
   &  $L^2$-SP      &  $89.69 \pm 1.39$  & $\underline{ 93.77 \pm 0.37 }$ &  $89.21 \pm 0.92$  &  $81.94 \pm 1.20$  &  $50.21\pm4.41$                &  $61.07 \pm 1.22$  & $\underline{ 82.97 \pm 0.39 }$ &  $71.02 \pm 0.57$  & $77.48$ & $79.32$ & $5.00$ \\
   &  Feature-map   &  $79.93 \pm 1.54$  &  $90.59 \pm 0.39$  &  $83.69 \pm 0.24$  &  $77.66 \pm 0.46$  & $\mathbf{ 80.03 \pm 1.01 }$ &  $59.93 \pm 0.14$  &  $75.32 \pm 0.19$  &  $67.51 \pm 0.30$  & $76.83$ & $77.36$ & $6.25$ \\
   &  BSS           & $\underline{ 90.17 \pm 2.84 }$ & $\mathbf{ 94.16 \pm 0.55 }$ &  $89.74 \pm 1.12$  &  $83.96 \pm 1.29$  &  $76.64\pm1.29$                &  $61.87 \pm 0.69$  & $\mathbf{ 83.26 \pm 0.57 }$ & $\mathbf{ 74.55 \pm 0.31 }$ & $81.79$ & $83.05$ & $3.38$ \\
\midrule
\multirow{8}{*}{Scaffold}

% & Full-FT      & $74.94 \pm 7.23$ & $68.62 \pm 0.80$ & $75.35 \pm 2.06$ & $76.03 \pm 0.91$ & $73.43 \pm 2.50$ & $57.88 \pm 1.18$ & $76.67 \pm 0.68$ & $63.62 \pm 0.27$ \\  
% & LP           & $65.07 \pm 1.08$ & $59.39 \pm 0.35$ & $69.24 \pm 0.16$ & $69.97 \pm 0.57$ & $71.81 \pm 2.40$ & $59.93 \pm 0.37$ & $69.87 \pm 0.28$ & $60.05 \pm 0.25$ \\  
% & Surgical-FT  & $71.07 \pm 4.16$ & $67.78 \pm 0.60$ & $80.16 \pm 2.36$ & $76.80 \pm 1.06$ & $75.87 \pm 0.82$ & $59.24 \pm 1.22$ & $75.54 \pm 0.64$ & $63.27 \pm 0.70$ \\  
% & LP-FT        & $75.07 \pm 2.24$ & $67.05 \pm 1.42$ & $75.33 \pm 1.14$ & $76.68 \pm 0.82$ & $71.36 \pm 1.39$ & $58.51 \pm 1.15$ & $76.85 \pm 0.63$ & $62.98 \pm 0.51$ \\  
% & WiSE-FT      & $77.27 \pm 4.28$ & $68.72 \pm 0.75$ & $77.37 \pm 1.44$ & $75.91 \pm 0.74$ & $74.38 \pm 2.20$ & $58.19 \pm 1.26$ & $76.89 \pm 0.69$ & $64.05 \pm 0.34$ \\  
% & $L^2$-SP     & $74.62 \pm 4.99$ & $68.30 \pm 1.19$ & $79.91 \pm 2.29$ & $73.97 \pm 0.78$ & $61.62\pm2.07$               & $59.78 \pm 0.33$ & $75.39 \pm 0.51$ & $62.34 \pm 0.82$ \\  
% & Feature-map  & $61.06 \pm 2.00$ & $65.12 \pm 1.98$ & $82.66 \pm 0.62$ & $74.54 \pm 1.00$ & $72.81 \pm 1.16$ & $60.47 \pm 0.45$ & $70.39 \pm 0.11$ & $60.10 \pm 0.19$ \\  
% & BSS          & $73.89 \pm 6.04$ & $70.04 \pm 2.00$ & $77.94 \pm 2.04$ & $76.28 \pm 1.28$ & $76.20\pm1.33$               & $59.99 \pm 1.39$ & $75.86 \pm 1.08$ & $63.62 \pm 0.50$ \\
&  Full-FT       &  $74.94 \pm 7.23$  &  $68.62 \pm 0.80$  &  $75.35 \pm 2.06$  &  $76.03 \pm 0.91$  &  $73.43 \pm 2.50$  &  $57.88 \pm 1.18$  &  $76.67 \pm 0.68$  & $\underline{ 63.62 \pm 0.27 }$ & $70.82$ & $72.00$ & $4.25$ \\
   &  LP            &  $65.07 \pm 1.08$  &  $59.39 \pm 0.35$  &  $69.24 \pm 0.16$  &  $69.97 \pm 0.57$  &  $71.81 \pm 2.40$  &  $59.93 \pm 0.37$  &  $69.87 \pm 0.28$  &  $60.05 \pm 0.25$  & $65.67$ & $65.69$ & $7.00$ \\
   &  Surgical-FT   &  $71.07 \pm 4.16$  &  $67.78 \pm 0.60$  & $\underline{ 80.16 \pm 2.36 }$ & $\mathbf{ 76.80 \pm 1.06 }$ & $\underline{ 75.87 \pm 0.82 }$ &  $59.24 \pm 1.22$  &  $75.54 \pm 0.64$  &  $63.27 \pm 0.70$  & $71.22$ & $71.72$ & $3.75$ \\
   &  LP-FT         & $\underline{ 75.07 \pm 2.24 }$ &  $67.05 \pm 1.42$  &  $75.33 \pm 1.14$  & $\underline{ 76.68 \pm 0.82 }$ &  $71.36 \pm 1.39$  &  $58.51 \pm 1.15$  & $\underline{ 76.85 \pm 0.63 }$ &  $62.98 \pm 0.51$  & $70.48$ & $71.41$ & $4.62$ \\
   &  WiSE-FT       & $\mathbf{ 77.27 \pm 4.28 }$ & $\underline{ 68.72 \pm 0.75 }$ &  $77.37 \pm 1.44$  &  $75.91 \pm 0.74$  &  $74.38 \pm 2.20$  &  $58.19 \pm 1.26$  & $\mathbf{ 76.89 \pm 0.69 }$ & $\mathbf{ 64.05 \pm 0.34 }$ & $71.60$ & $72.87$ & $3.12$ \\
   &  $L^2$-SP      &  $74.62 \pm 4.99$  &  $68.30 \pm 1.19$  &  $79.91 \pm 2.29$  &  $73.97 \pm 0.78$  &  $61.62\pm2.07$                &  $59.78 \pm 0.33$  &  $75.39 \pm 0.51$  &  $62.34 \pm 0.82$  & $69.49$ & $69.37$ & $5.25$ \\
   &  Feature-map   &  $61.06 \pm 2.00$  &  $65.12 \pm 1.98$  & $\mathbf{ 82.66 \pm 0.62 }$ &  $74.54 \pm 1.00$  &  $72.81 \pm 1.16$  & $\mathbf{ 60.47 \pm 0.45 }$ &  $70.39 \pm 0.11$  &  $60.10 \pm 0.19$  & $68.39$ & $67.40$ & $5.25$ \\
   &  BSS           &  $73.89 \pm 6.04$  & $\mathbf{ 70.04 \pm 2.00 }$ &  $77.94 \pm 2.04$  &  $76.28 \pm 1.28$  & $\mathbf{ 76.20\pm1.33               }$ & $\underline{ 59.99 \pm 1.39 }$ &  $75.86 \pm 1.08$  & $\underline{ 63.62 \pm 0.50 }$ & $71.73$ & $72.65$ & $2.75$ \\
\midrule
\multirow{8}{*}{Size}

%  & Full-FT      & $61.94 \pm 2.67$ & $82.80 \pm 2.31$ & $63.62 \pm 1.19$ & $77.81 \pm 2.99$ & $72.05 \pm 2.96$ & $54.92 \pm 0.79$ & $71.08 \pm 0.77$ & $62.47 \pm 0.83$ \\  
% & LP           & $55.54 \pm 0.65$ & $75.89 \pm 0.90$ & $42.31 \pm 0.48$ & $67.54 \pm 1.27$ & $69.87 \pm 1.51$ & $53.74 \pm 0.43$ & $68.10 \pm 0.39$ & $57.50 \pm 0.19$ \\  
% & Surgical-FT  & $64.54 \pm 8.03$ & $88.90 \pm 0.74$ & $61.99 \pm 2.13$ & $78.10 \pm 0.96$ & $76.07 \pm 0.57$ & $57.13 \pm 1.87$ & $72.24 \pm 0.28$ & $60.52 \pm 0.95$ \\  
% & LP-FT        & $63.79 \pm 3.29$ & $83.12 \pm 5.20$ & $65.48 \pm 0.70$ & $76.47 \pm 3.53$ & $72.24 \pm 2.79$ & $56.31 \pm 0.72$ & $72.65 \pm 0.59$ & $61.71 \pm 0.63$ \\  
% & WiSE-FT      & $63.85 \pm 3.69$ & $81.81 \pm 2.80$ & $62.71 \pm 1.26$ & $77.83 \pm 2.02$ & $73.40 \pm 2.08$ & $56.63 \pm 0.63$ & $71.27 \pm 0.77$ & $62.70 \pm 0.87$ \\  
% & $L^2$-SP     & $63.67 \pm 1.79$ & $88.00 \pm 1.00$ & $63.98 \pm 1.51$ & $77.38 \pm 1.25$ & $58.29\pm3.74$               & $56.23 \pm 1.70$ & $71.93 \pm 0.21$ & $59.29 \pm 0.72$ \\  
% & Feature-map  & $64.41 \pm 1.38$ & $86.82 \pm 0.76$ & $59.62 \pm 1.17$ & $70.71 \pm 0.99$ & $76.01 \pm 0.60$ & $55.03 \pm 0.30$ & $67.98 \pm 0.41$ & $57.91 \pm 0.31$ \\  
% & BSS          & $67.80 \pm 4.60$ & $84.90 \pm 2.20$ & $62.77 \pm 3.69$ & $78.13 \pm 2.21$ & $74.58\pm1.13$               & $54.91 \pm 1.34$ & $71.40 \pm 0.44$ & $63.04 \pm 0.35$ \\
&  Full-FT       &  $61.94 \pm 2.67$  &  $82.80 \pm 2.31$  &  $63.62 \pm 1.19$  &  $77.81 \pm 2.99$  &  $72.05 \pm 2.96$  &  $54.92 \pm 0.79$  &  $71.08 \pm 0.77$  &  $62.47 \pm 0.83$  & $68.34$ & $68.16$ & $5.12$ \\
   &  LP            &  $55.54 \pm 0.65$  &  $75.89 \pm 0.90$  &  $42.31 \pm 0.48$  &  $67.54 \pm 1.27$  &  $69.87 \pm 1.51$  &  $53.74 \pm 0.43$  &  $68.10 \pm 0.39$  &  $57.50 \pm 0.19$  & $61.31$ & $62.05$ & $7.75$ \\
   &  Surgical-FT   & $\underline{ 64.54 \pm 8.03 }$ & $\mathbf{ 88.90 \pm 0.74 }$ &  $61.99 \pm 2.13$  & $\underline{ 78.10 \pm 0.96 }$ & $\mathbf{ 76.07 \pm 0.57 }$ & $\mathbf{ 57.13 \pm 1.87 }$ & $\underline{ 72.24 \pm 0.28 }$ &  $60.52 \pm 0.95$  & $69.94$ & $68.91$ & $2.50$ \\
   &  LP-FT         &  $63.79 \pm 3.29$  &  $83.12 \pm 5.20$  & $\mathbf{ 65.48 \pm 0.70 }$ &  $76.47 \pm 3.53$  &  $72.24 \pm 2.79$  &  $56.31 \pm 0.72$  & $\mathbf{ 72.65 \pm 0.59 }$ &  $61.71 \pm 0.63$  & $68.97$ & $68.72$ & $3.75$ \\
   &  WiSE-FT       &  $63.85 \pm 3.69$  &  $81.81 \pm 2.80$  &  $62.71 \pm 1.26$  &  $77.83 \pm 2.02$  &  $73.40 \pm 2.08$  & $\underline{ 56.63 \pm 0.63 }$ &  $71.27 \pm 0.77$  & $\underline{ 62.70 \pm 0.87 }$ & $68.78$ & $68.63$ & $4.00$ \\
   &  $L^2$-SP      &  $63.67 \pm 1.79$  & $\underline{ 88.00 \pm 1.00 }$ & $\underline{ 63.98 \pm 1.51 }$ &  $77.38 \pm 1.25$  &  $58.29\pm3.74$                &  $56.23 \pm 1.70$  &  $71.93 \pm 0.21$  &  $59.29 \pm 0.72$  & $67.35$ & $65.76$ & $4.50$ \\
   &  Feature-map   &  $64.41 \pm 1.38$  &  $86.82 \pm 0.76$  &  $59.62 \pm 1.17$  &  $70.71 \pm 0.99$  & $\underline{ 76.01 \pm 0.60 }$ &  $55.03 \pm 0.30$  &  $67.98 \pm 0.41$  &  $57.91 \pm 0.31$  & $67.31$ & $66.11$ & $5.25$ \\
   &  BSS           & $\mathbf{ 67.80 \pm 4.60 }$ &  $84.90 \pm 2.20$  &  $62.77 \pm 3.69$  & $\mathbf{ 78.13 \pm 2.21 }$ &  $74.58\pm1.13$                &  $54.91 \pm 1.34$  &  $71.40 \pm 0.44$  & $\mathbf{ 63.04 \pm 0.35 }$ & $69.69$ & $69.62$ & $3.12$ \\

\midrule
\multicolumn{13}{c}{{\textbf{Supervised Pre-training (Graphium-Large)}}}  \\ \midrule

\multirow{8}{*}{Scaffold}

&  Full-FT   &  $81.27\pm 3.88$  &  $69.17\pm 1.32$  &  $79.75\pm 1.07$  &  $76.42\pm 0.72$  &  $76.84\pm 1.80$  & $\underline{ 63.63\pm 0.06 }$ &  $78.12\pm 0.46$  &  $66.37\pm 0.26$ & $73.95$ & $74.45$ & $3.75$ \\
 &  LP     &  $80.48\pm 0.00$  &  $66.90\pm 0.00$  &  $80.44\pm 0.00$  &  $75.83\pm 0.00$  &  $73.35\pm 0.00$  &  $62.03\pm 0.00$  & $\underline{ 79.02\pm 0.00 }$ &  $66.09\pm 0.00$ & $73.02$ & $73.61$ & $5.12$ \\
 &  Surgical-FT   & $\underline{ 86.17\pm0.00  }$ & $\mathbf{ 73.71\pm0.00  }$ & $\mathbf{ 84.16\pm0.00  }$ & $\mathbf{ 77.47\pm0.00  }$ & $\mathbf{ 78.87\pm0.00  }$ & $\mathbf{ 64.02\pm0.00  }$ &  $78.23\pm0.00 $  & $\mathbf{ 67.34\pm0.00 }$ & $76.25$ & $76.63$ & $1.38$ \\
 &  LP-FT   &  $83.67\pm 3.53$  &  $69.98\pm 0.83$  &  $79.28\pm 0.32$  &  $76.17\pm 2.01$  &  $77.82\pm 1.15$  &  $61.20\pm 0.00$  &  $76.94\pm 0.00$  &  $66.28\pm 0.00$  & $73.92$ & $74.41$ & $4.62$ \\
 &  WiSE-FT  &  $85.40\pm1.61 $  & $\underline{ 71.89\pm1.79  }$ &  $78.13\pm2.92 $  & $\underline{ 76.69\pm1.76  }$ &  $74.37\pm1.79 $  &  $63.58\pm0.00 $  &  $77.98\pm0.33 $  & $\underline{ 66.48\pm0.43 }$ & $74.31$ & $74.26$ & $3.62$ \\
 &  $L^2$-SP  &  $76.83\pm8.87 $  &  $67.35\pm 0.82$  &  $78.17\pm0.02 $  &  $73.69\pm0.03 $  &  $62.35\pm0.15 $  &  $62.21\pm0.45 $  &  $76.27\pm0.32 $  &  $62.75\pm0.88 $ & $69.95$ & $69.87$ & $6.62$ \\
 &  Feature-map  & $\mathbf{ 90.13 \pm2.12  }$ &  $70.99 \pm 0.27$  & $\underline{ 83.17 \pm0.49  }$ &  $73.61 \pm0.03 $  & $\underline{ 78.74 \pm0.76  }$ &  $62.12 \pm0.02 $  & $\mathbf{ 79.99 \pm 0.12 }$ &  $65.03 \pm 0.08$ & $75.47$ & $75.25$ & $3.50$ \\
 &  BSS  &  $79.99 \pm5.89 $  &  $67.10 \pm0.93 $  &  $78.12 \pm2.32 $  &  $72.50 \pm0.51 $  &  $61.20 \pm 0.08$  &  $61.13 \pm0.95 $  &  $76.69 \pm 0.64$  &  $65.45 \pm0.89 $ & $70.27$ & $70.18$ & $7.38$ \\

\midrule
\multirow{8}{*}{Size}
% & Full-FT  & $85.96\pm 4.28$ & $87.62\pm 0.90$ & $67.41\pm 2.44$ & $81.47\pm 1.94$ & $72.03\pm 2.55$ & $54.72\pm 0.01$ & $69.71\pm 0.37$ & $61.31\pm 0.37$\\
% & LP    & $81.84\pm 0.02$ & $78.09\pm 0.00$ & $58.08\pm 0.01$ & $77.48\pm 0.00$ & $69.46\pm 0.00$ & $53.59\pm 0.00$ & $73.65\pm 0.00$ & $61.25\pm 0.00$\\
% & Surgical-FT  & $86.59\pm $ & $89.07\pm $ & $70.94\pm $ & $82.50\pm $ & $74.47\pm $ & $56.24\pm $ & $72.30\pm $ & $62.74\pm $\\
% & LP-FT  & $86.78\pm 2.69$ & $88.02\pm 1.50$ & $63.72\pm 1.85$ & $82.57\pm 0.46$ & $73.51\pm 1.77$ & $52.40\pm 0.00$  & $68.23\pm 0.87$ & $60.85\pm 0.00$\\
% & WiSE-FT & $82.44\pm $ & $87.76\pm $ & $72.89\pm $ & $81.37\pm $ & $73.67\pm $ & $55.87\pm $ & $68.85\pm $ & $60.61\pm $\\
% & $L^2$-SP & $71.03\pm $ & $81.32\pm $ & $68.82\pm $ & $70.66\pm $ & $64.69\pm $ & $52.08\pm $ & $70.91\pm $ & $56.50\pm $\\
% & Feature-map & $82.48\pm $ & $87.70\pm $ & $69.56\pm $ & $67.23\pm $ & $71.49\pm $ & $54.43\pm $ & $74.12\pm $ & $58.73\pm $\\
% & BSS & $72.42\pm $ & $82.92\pm $ & $62.76\pm $ & $72.81\pm $ & $65.79\pm $ & $52.89\pm $ & $71.91\pm $ & $57.79\pm $\\

&  Full-FT   &  $85.96\pm 4.28$  &  $87.62\pm 0.90$  &  $67.41\pm 2.44$  &  $81.47\pm 1.94$  &  $72.03\pm 2.55$  &  $54.72\pm 0.01$  &  $69.71\pm 0.37$  & $\underline{ 61.31\pm 0.37}$ & $72.53$ & $72.98$ & $3.88$ \\
 &  LP     &  $81.84\pm 0.02$  &  $78.09\pm 0.00$  &  $58.08\pm 0.01$  &  $77.48\pm 0.00$  &  $69.46\pm 0.00$  &  $53.59\pm 0.00$  & $\underline{ 73.65\pm 0.00 }$ &  $61.25\pm 0.00$ & $69.18$ & $69.67$ & $5.38$ \\
 &  Surgical-FT   & $\underline{ 86.59\pm 0.01 }$ & $\mathbf{ 89.07\pm 0.00 }$ & $\underline{ 70.94\pm0.01  }$ & $\underline{ 82.50\pm 0.00 }$ & $\mathbf{ 74.47\pm 0.00 }$ & $\mathbf{ 56.24\pm0.00  }$ &  $72.30\pm 0.00$  & $\mathbf{ 62.74\pm0.00 }$ & $74.36$ & $74.92$ & $1.62$ \\
 &  LP-FT   & $\mathbf{ 86.78\pm 2.69 }$ & $\underline{ 88.02\pm 1.50 }$ &  $63.72\pm 1.85$  & $\mathbf{ 82.57\pm 0.46 }$ &  $73.51\pm 1.77$  &  $52.40\pm 0.00$   &  $68.23\pm 0.87$  &  $60.85\pm 0.00$ & $72.01$ & $72.61$ & $4.00$ \\
 &  WiSE-FT  &  $82.44\pm3.02 $  &  $87.76\pm0.5 $  & $\mathbf{ 72.89\pm0.66  }$ &  $81.37\pm1.07 $  & $\underline{ 73.67\pm3.44  }$ & $\underline{ 55.87\pm0.01  }$ &  $68.85\pm0.84 $  &  $60.61\pm0.53 $ & $72.93$ & $73.31$ & $3.62$ \\
 &  $L^2$-SP  &  $71.03\pm3.67 $  &  $81.32\pm1.51 $  &  $68.82\pm0.06 $  &  $70.66\pm0.00 $  &  $64.69\pm0.32 $  &  $52.08\pm0.84 $  &  $70.91\pm0.34 $  &  $56.50\pm0.01 $ & $67.00$ & $67.10$ & $6.88$ \\
 &  Feature-map  &  $82.48\pm3.25 $  &  $87.70\pm0.64 $  &  $69.56\pm0.20 $  &  $67.23\pm1.93 $  &  $71.49\pm0.13 $  &  $54.43\pm0.03 $  & $\mathbf{ 74.12\pm0.09  }$ &  $58.73\pm0.04 $ & $70.72$ & $70.60$ & $4.38$ \\
 &  BSS  &  $72.42\pm0.03 $  &  $82.92\pm1.60 $  &  $62.76\pm4.23 $  &  $72.81\pm0.66 $  &  $65.79\pm 5.31$  &  $52.89\pm1.12 $  &  $71.91\pm0.44 $  &  $57.79\pm1.80 $ & $67.41$ & $67.25$ & $6.25$ \\
 \bottomrule[2pt]
\label{table:nonfew_cls_molebert}
\end{tabular}
\end{sc}
\end{small}
\end{adjustbox}
\end{center}
\end{table*}

\begin{table*}[t]
\caption{Robust fine-tuning performance on 4 \hlb{\textbf{Regression}}{30}
datasets (RMSE metrics) in the \hlr{\textbf{Non-Fewshot}}{30} setting, evaluated across 3 dataset splits (\textsc{Random, Scaffold, Size}) over \hlg{\textsc{\textbf{MoleculeSTM}}}{30} and \hlg{\textsc{\textbf{Graphium-Large}}}{30} models.  \textsc{Avg-R,Avg-R}$^*$ denote the average rank and the rank based on the average normalized performance over all the datasets for each method, respectively. Standard deviations across five replicates are shown in parentheses. We \textbf{bold} and \underline{underline} the best and second-best performances in each scenario.}
\label{tab-stm-large-rgs-non}
\begin{center}
\begin{adjustbox}{width = 1\textwidth}
\begin{small}
\begin{sc}
\begin{tabular}{cccccccc|cccccc}
\toprule[2pt]
\multirow{2}{*}{Split} &
  \multirow{2}{*}{Methods} &
  \multicolumn{6}{c}{Self-supervised Pre-training (MoleculeSTM)} &
  \multicolumn{5}{c}{Supervised Pre-training (Graphium-Large)} \\ \cmidrule(lr){3-14}
 &
   &
  Esol &
  Lipo &
  Malaria &
  Cep &
  Avg-R &
  \multicolumn{1}{c|}{Avg-R$^*$} &
  Esol &
  Lipo &
  Malaria &
  Cep &
  Avg-R &
  \multicolumn{1}{c}{Avg-R$^*$} \\ \midrule
\multirow{8}{*}{Random} 
  &  Full-FT       &  $0.901 \pm 0.063$  &  $0.660 \pm 0.013$  & $\underline{ 1.067 \pm 0.009 }$ &  $1.401 \pm 0.035$  & $3.00$ & 2 & $0.643 \pm 0.011$ & $0.605 \pm 0.011$ & $1.085 \pm 0.007$ & $1.399 \pm 0.015$ & $4.00$ & 4 \\
   &  LP            &  $1.374 \pm 0.011$  &  $1.067 \pm 0.015$  &  $1.207 \pm 0.004$  &  $1.999 \pm 0.003$  & $8.00$ & 8 & $0.699 \pm 0.000$ & $0.672 \pm 0.000$ & $1.105 \pm 0.002$ & $1.658 \pm 0.011$ & $7.75$ & 8 \\
   &  Surgical-FT   &  $1.056 \pm 0.028$  &  $0.724 \pm 0.011$  &  $1.074 \pm 0.010$  &  $1.547 \pm 0.011$  & $6.00$ & 6 & $\mathbf{0.617 \pm 0.000}$ & $\mathbf{0.582 \pm 0.000}$ & $\mathbf{1.047 \pm 0.000}$ & $\underline{1.392 \pm 0.000}$ & $1.25$ & 1 \\
   &  LP-FT         &  $0.922 \pm 0.023$  & $\underline{ 0.654 \pm 0.023 }$ &  $1.076 \pm 0.014$  & $\underline{ 1.365 \pm 0.029 }$ & $3.25$ & 3 & $\underline{0.618 \pm 0.023}$ & $\underline{0.591 \pm 0.008}$ & $\underline{1.059 \pm 0.000}$ & $\mathbf{1.355 \pm 0.008}$ & $2.00$ & 2 \\
   &  WiSE-FT       &  $0.934 \pm 0.061$  &  $0.662 \pm 0.016$  & $\mathbf{ 1.064 \pm 0.007 }$ &  $1.460 \pm 0.042$  & $3.75$ & 5 & $0.630 \pm 0.006$ & $0.606 \pm 0.008$ & $1.086 \pm 0.007$ & $1.430 \pm 0.019$ & $5.00$ & 3 \\
   &  $L^2$-SP      & $\mathbf{ 0.884 \pm 0.025 }$ &  $0.666 \pm 0.014$  &  $1.087 \pm 0.011$  &  $1.385 \pm 0.031$  & $3.75$ & 4 & $0.647 \pm 0.028$ & $0.662 \pm 0.014$ & $\underline{1.059 \pm 0.001}$ & $1.466 \pm 0.050$ & $5.75$ & 7 \\
   &  Feature-map   &  $1.018 \pm 0.024$  &  $0.789 \pm 0.018$  &  $1.106 \pm 0.005$  &  $1.536 \pm 0.008$  & $6.50$ & 7 & $0.660 \pm 0.240$ & $0.642 \pm 0.009$ & $\underline{1.059 \pm 0.001}$ & $1.419 \pm 0.037$ & $5.25$ & 5 \\
   &  BSS           & $\underline{ 0.887 \pm 0.030 }$ & $\mathbf{ 0.641 \pm 0.014 }$ &  $1.070 \pm 0.016$  & $\mathbf{ 1.351 \pm 0.016 }$ & $1.75$ & 1 & $0.619 \pm 0.030$ & $0.611 \pm 0.017$ & $1.158 \pm 0.041$ & $1.404 \pm 0.029$ & $5.00$ & 6 \\

\midrule
\multirow{8}{*}{Scaffold} &  Full-FT       &  $1.360 \pm 0.049$  &  $0.752 \pm 0.018$  &  $1.105 \pm 0.018$  &  $1.395 \pm 0.041$  & $4.50$ & 5 & $0.878 \pm 0.010$ & $0.731 \pm 0.003$ & $1.107 \pm 0.008$ & $\underline{1.409 \pm 0.037}$ & 4.50 & 5 \\
   &  LP            &  $1.608 \pm 0.030$  &  $0.983 \pm 0.006$  &  $1.133 \pm 0.002$  &  $2.009 \pm 0.004$  & $8.00$ & 8 & $0.886 \pm 0.005$ & $0.772 \pm 0.000$ & $1.103 \pm 0.000$ & $1.635 \pm 0.017$ & 6.25 & 7 \\
   &  Surgical-FT   & $\mathbf{ 1.297 \pm 0.044 }$ &  $0.765 \pm 0.013$  &  $1.105 \pm 0.013$  &  $1.518 \pm 0.010$  & $4.50$ & 6 & $\underline{0.863 \pm 0.000}$ & $\mathbf{0.675 \pm 0.000}$ & $\underline{1.090 \pm 0.000}$ & $1.480 \pm 0.000$ & 2.75 & 3 \\
   &  LP-FT         &  $1.331 \pm 0.033$  & $\underline{ 0.743 \pm 0.017 }$ &  $1.107 \pm 0.011$  &  $1.356 \pm 0.030$  & $4.00$ & 4 & $0.887 \pm 0.002$ & $0.709 \pm 0.016$ & $1.091 \pm 0.007$ & $\mathbf{1.380 \pm 0.005}$ & 3.75 & 4 \\
   &  WiSE-FT       &  $1.347 \pm 0.036$  & $\mathbf{ 0.740 \pm 0.018 }$ & $\mathbf{ 1.090 \pm 0.015 }$ &  $1.505 \pm 0.045$  & $3.00$ & 2 & $0.876 \pm 0.011$ & $0.727 \pm 0.004$ & $1.120 \pm 0.008$ & $1.430 \pm 0.041$ & 4.75 & 6 \\
   &  $L^2$-SP      & $\underline{ 1.300 \pm 0.017 }$ &  $0.756 \pm 0.017$  &  $1.106 \pm 0.005$  & $\underline{ 1.347 \pm 0.020 }$ & $3.75$ & 3 & $0.905 \pm 0.022$ & $0.778 \pm 0.009$ & $1.147 \pm 0.003$ & $1.518 \pm 0.011$ & 7.50 & 8 \\
   &  Feature-map   &  $1.383 \pm 0.008$  &  $0.824 \pm 0.009$  &  $1.098 \pm 0.004$  &  $1.518 \pm 0.003$  & $6.00$ & 7 & $\mathbf{0.853 \pm 0.005}$ & $\underline{0.692 \pm 0.002}$ & $1.149 \pm 0.002$ & $1.427 \pm 0.052$ & 3.50 & 2 \\
   &  BSS           & $\underline{ 1.300 \pm 0.024 }$ &  $0.746 \pm 0.010$  & $\underline{ 1.097 \pm 0.013 }$ & $\mathbf{ 1.319 \pm 0.023 }$ & $2.25$ & 1 & $0.873 \pm 0.024$ & $0.707 \pm 0.015$ & $\mathbf{0.166 \pm 0.000}$ & $1.431 \pm 0.016$ & 3.00 & 1 \\

   \midrule
\multirow{8}{*}{Size} &  Full-FT       &  $1.490 \pm 0.153$  &  $0.711 \pm 0.017$  & $\mathbf{ 0.883 \pm 0.008 }$ &  $1.834 \pm 0.038$  & $3.25$ & 2 & $1.020 \pm 0.009$ & $0.727 \pm 0.006$ & $0.890 \pm 0.013$ & $\mathbf{1.847 \pm 0.043}$ & 2.75 & 2 \\
   &  LP            &  $2.172 \pm 0.065$  &  $0.935 \pm 0.004$  &  $0.912 \pm 0.004$  &  $2.402 \pm 0.018$  & $8.00$ & 8 & $1.190 \pm 0.000$ & $0.852 \pm 0.000$ & $0.912 \pm 0.000$ & $2.101 \pm 0.026$ & 7.75 & 8 \\
   &  Surgical-FT   &  $1.499 \pm 0.093$  &  $0.769 \pm 0.013$  &  $0.889 \pm 0.014$  &  $1.998 \pm 0.020$  & $5.25$ & 6 & $1.105 \pm 0.000$ & $0.745 \pm 0.000$ & $\mathbf{0.871 \pm 0.000}$ & $1.902 \pm 0.000$ & 4.50 & 5 \\
   &  LP-FT         & $\underline{ 1.401 \pm 0.053 }$ & $\underline{ 0.703 \pm 0.012 }$ &  $0.897 \pm 0.009$  & $\underline{ 1.763 \pm 0.037 }$ & $3.25$ & 3 & $1.067 \pm 0.034$ & $\mathbf{0.703 \pm 0.016}$ & $0.892 \pm 0.014$ & $1.884 \pm 0.017$ & 3.75 & 4 \\
   &  WiSE-FT       &  $1.583 \pm 0.118$  &  $0.727 \pm 0.018$  &  $0.889 \pm 0.008$  &  $1.902 \pm 0.053$  & $5.25$ & 5 & $1.026 \pm 0.011$ & $0.721 \pm 0.009$ & $\underline{0.888 \pm 0.011}$ & $\underline{1.848 \pm 0.035}$ & 2.75 & 1 \\
   &  $L^2$-SP      & $\mathbf{ 1.390 \pm 0.115 }$ &  $0.725 \pm 0.019$  &  $0.896 \pm 0.007$  &  $1.786 \pm 0.022$  & $3.25$ & 4 & $\underline{1.001 \pm 0.023}$ & $0.805 \pm 0.010$ & $0.903 \pm 0.004$ & $2.008 \pm 0.130$ & 5.25 & 6 \\
   &  Feature-map   &  $1.458 \pm 0.045$  &  $0.849 \pm 0.012$  &  $0.896 \pm 0.011$  &  $2.007 \pm 0.018$  & $6.00$ & 7 & $1.028 \pm 0.032$ & $0.767 \pm 0.002$ & $0.925 \pm 0.002$ & $2.079 \pm 0.021$ & 6.50 & 7 \\
   &  BSS           &  $1.408 \pm 0.100$  & $\mathbf{ 0.700 \pm 0.020 }$ & $\underline{ 0.887 \pm 0.011 }$ & $\mathbf{ 1.725 \pm 0.026 }$ & $1.75$ & 1 & $\mathbf{0.985 \pm 0.040}$ & $\underline{0.720 \pm 0.013}$ & $0.901 \pm 0.006$ & $1.883 \pm 0.023$ & 2.75 & 3 \\

\midrule[2pt]
\end{tabular}%
\end{sc}
\end{small}
\end{adjustbox}
\end{center}
\end{table*}

\begin{table*}[t]
\caption{Robust fine-tuning performance on 5 \hlb{\textbf{Classification}}{30}
datasets (AUC metrics) in the \hlr{\textbf{Fewshot}}{30} setting (covering \textsc{Fewshot-50, Fewshot-100, Fewshot-500}), evaluated across 3 dataset splits (\textsc{Random, Scaffold, Size}) over \hlg{\textsc{\textbf{MoleculeSTM}}}{30} and \hlg{\textsc{\textbf{Graphium-Large}}}{30} models.  We \textbf{bold} and \underline{underline} the best and second-best performances in each scenario.}
\label{tab-stm-large-clf-few}
\begin{center}
\begin{adjustbox}{width = 1\textwidth}
\begin{small}
\begin{sc}
\begin{tabular}{cccccccccc|cccccccc}
\toprule[2pt]
 \multirow{2}{*}{Split}&\multirow{2}{*}{Methods} & \multicolumn{8}{c}{Self-supervised Pre-training (MoleculeSTM)} & \multicolumn{8}{c}{Supervised Pre-training (Graphium-Large)} \\
\cmidrule(lr){3-10}\cmidrule(lr){10-18}        
  &   &   ClinTox &  BBBP  &     Bace    & HIV & Sider & avg & avg-F & avg-R &   ClinTox &  BBBP  &     Bace    & HIV & Sider & avg & avg-F & avg-R  \\
\midrule
\multicolumn{18}{c}{{\textbf{Fewshot-50}}}\\
\midrule
\multirow{8}{*}{Random}

&  Full-FT       &  $49.60 \pm 2.85$  &  $84.86 \pm 1.30$  &  $74.74 \pm 1.44$  & $\underline{ 60.58 \pm 1.47 }$ &  $49.47 \pm 0.90$  & $63.85$ & $61.64$ & $4.80$ &$74.25 \pm 0.00$&$82.09 \pm 0.77$&$81.04 \pm 0.00$&$62.83 \pm 0.00$&$52.55 \pm 0.00$&$70.55$&$72.71$&$6.00$\\
   &  LP            &  $52.66 \pm 3.14$  &  $78.85 \pm 1.75$  &  $58.02 \pm 3.19$  &  $52.39 \pm 0.52$  &  $50.23 \pm 0.47$  & $58.43$ & $54.36$ & $6.40$ &$64.37 \pm 0.00$&$86.23 \pm 0.00$&$81.47 \pm 0.00$&$60.00 \pm 0.00$&$54.28 \pm 0.00$&$69.27$&$68.61$&$5.00$\\
   &  Surgical-FT   & $\underline{ 54.43 \pm 4.39 }$ & $\mathbf{ 86.64 \pm 0.96 }$ & $\underline{ 74.92 \pm 0.95 }$ & $\mathbf{ 61.71 \pm 0.64 }$ &  $51.10 \pm 0.82$  & $65.76$ & $63.69$ & $2.00$ &$\underline{76.03 \pm 0.00 }$&$\underline{ 87.04 \pm 0.00 }$&$\mathbf{ 82.44 \pm 0.00 }$&$62.09 \pm 0.00$&$53.09 \pm 0.00$&$72.14$&$73.52$&$3.60$\\
   &  LP-FT         &  $47.71 \pm 2.16$  &  $84.36 \pm 2.65$  & $\underline{ 74.92 \pm 0.95 }$ &  $55.82 \pm 1.53$  & $\mathbf{ 51.62 \pm 0.37 }$ & $62.89$ & $60.79$ & $4.60$ &$\mathbf{76.40 \pm 0.00 }$&$82.10 \pm 0.00$&$73.86 \pm 0.00$&$\underline{ 64.86 \pm 0.00 }$&$54.11 \pm 0.00$&$70.27$&$71.71$&$4.20$\\
   &  WiSE-FT       & $\mathbf{ 55.69 \pm 5.37 }$ &  $84.62 \pm 1.45$  &  $74.02 \pm 1.36$  &  $60.05 \pm 1.26$  &  $49.41 \pm 0.89$  & $64.76$ & $63.25$ & $4.60$ &$75.77 \pm 0.00$&$84.05 \pm 0.85$&$81.30 \pm 0.00$&$62.46 \pm 0.00$&$\mathbf{ 55.49 \pm 0.00  }$&$71.81$&$73.18$&$3.60$\\
\end{tabular}
\end{sc}
\end{small}
\end{adjustbox}
\end{center}
\end{table*}

\begin{table*}[t]
\caption{Robust fine-tuning performance on 4 \hlb{\textbf{Regression}}{30}
datasets (RMSE metrics) in the \hlr{\textbf{Fewshot}}{30} setting (covering \textsc{Fewshot-50, Fewshot-100}, and \textsc{Fewshot-500}), evaluated across 3 dataset splits (\textsc{Random, Scaffold, Size}) over \hlg{\textsc{\textbf{MoleculeSTM}}}{30} and \hlg{\textsc{\textbf{Graphium-Large}}}{30} models.  \textsc{Avg-R, Avg-R}$^*$ denote the average rank and the rank based on the average normalized performance over all the datasets for each evavluated method, respectively. Standard deviations across five replicates are shown in parentheses. We \textbf{bold} and \underline{underline} the best and second-best performances in each scenario. }
\label{tab-stm-large-rgs-few}
\begin{center}
\begin{adjustbox}{width = 1\textwidth}
\begin{small}
\begin{sc}
\begin{tabular}{cccccccc|cccccc}
\midrule[2pt]
\multirow{2}{*}{Split} &
  \multirow{2}{*}{Methods} &
  \multicolumn{6}{c}{Self-supervised Pre-training (MoleculeSTM)} &
  \multicolumn{5}{c}{Supervised Pre-training (Graphium-Large)} \\ \cmidrule(lr){3-14} 
 &
   &
  Esol &
  Lipo &
  Malaria &
  CEP &
  Avg-R &
  Avg-R$^*$ &
  Esol &
  Lipo &
  Malaria &
  CEP & Avg-R &
  Avg-R$^*$ \\ \midrule
  \multicolumn{13}{c}{{\textbf{Fewshot-50}}}\\
  \midrule
\midrule \multirow{8}{*}{Random}   &  Full-FT       &  $2.128 \pm 0.072$  &  $1.247 \pm 0.031$  &  $1.310 \pm 0.025$  &  $3.433 \pm 0.226$  & $5.00$ & 6&$1.125 \pm 0.000$ & $1.156 \pm 0.019$ & $\underline{ 1.277 \pm 0.000 }$ & $2.198 \pm 0.001$ & $5.75$ &7 \\  
    &  LP            &  $2.971 \pm 0.017$  &  $1.638 \pm 0.014$  &  $1.309 \pm 0.012$  &  $3.519 \pm 0.052$  & $6.75$ & 8&$1.176 \pm 0.000$ & $1.131 \pm 0.000$ & $1.294 \pm 0.000$ & $2.113 \pm 0.000$ & $6.50$ &8 \\  
    &  Surgical-FT   &  $2.315 \pm 0.081$  &  $1.327 \pm 0.017$  &  $1.317 \pm 0.024$  &  $3.272 \pm 0.199$  & $6.50$ & 7&$\mathbf{ 1.055 \pm 0.000 }$ & $\underline{ 1.076 \pm 0.000 }$ & $1.283 \pm 0.000$ & $2.192 \pm 0.000$ & $4.00$ &4 \\  
    &  LP-FT         &  $1.600 \pm 0.129$  &  $1.181 \pm 0.030$  &  $1.356 \pm 0.011$  &  $2.358 \pm 0.037$  & $4.25$ & 4&$\underline{ 1.096 \pm 0.000 }$ & $\mathbf{ 1.032 \pm 0.002 }$ & $1.293 \pm 0.000$ & $\underline{ 2.092 \pm 0.002 }$ & $3.00$ &1 \\  
    &  WiSE-FT       &  $2.135 \pm 0.072$  &  $1.261 \pm 0.035$  & $\underline{ 1.298 \pm 0.023 }$ &  $3.576 \pm 0.235$  & $5.50$ & 5&$1.116 \pm 0.000$ & $1.151 \pm 0.024$ & $1.278 \pm 0.000$ & $\mathbf{ 2.075 \pm 0.004 }$ & $4.00$ &3 \\  
    &  $L^2$-SP      & $\underline{ 1.472 \pm 0.036 }$ & $\mathbf{ 1.165 \pm 0.037 }$ & $\mathbf{ 1.297 \pm 0.006 }$ & $\underline{ 2.304 \pm 0.055 }$ & $1.50$ & 1&$1.161 \pm 0.000$ & $1.077 \pm 0.019$ & $\mathbf{ 1.276 \pm 0.000 }$ & $2.127 \pm 0.015$ & $4.00$ &5 \\  
    &  Feature-map   &  $1.632 \pm 0.028$  &  $1.257 \pm 0.025$  &  $1.301 \pm 0.009$  &  $2.398 \pm 0.037$  & $4.00$ & 3&$1.133 \pm 0.002$ & $1.106 \pm 0.003$ & $\underline{ 1.277 \pm 0.001 }$ & $2.108 \pm 0.002$ & $3.75$ &2 \\  
    &  BSS           & $\mathbf{ 1.450 \pm 0.057 }$ & $\underline{ 1.171 \pm 0.021 }$ &  $1.314 \pm 0.018$  & $\mathbf{ 2.244 \pm 0.036 }$ & $2.50$ & 2&$1.188 \pm 0.004$ & $1.109 \pm 0.021$ & $\mathbf{ 1.276 \pm 0.000 }$ & $2.108 \pm 0.029$ & $4.25$ &6 \\  
 \midrule
\multirow{8}{*}{Scaffold} 
&  Full-FT       &  $2.790 \pm 0.116$  &  $1.434 \pm 0.072$  &  $1.195 \pm 0.025$  &  $3.395 \pm 0.191$  & $5.75$ & 6&
$1.237 \pm 0.000$ & $1.079 \pm 0.000$ & $1.175 \pm 0.000$ & $2.051 \pm 0.000$ & 4.00 & 7 \\
   &  LP            &  $3.538 \pm 0.075$  &  $1.755 \pm 0.021$  &  $1.206 \pm 0.012$  &  $3.870 \pm 0.038$  & $7.75$ & 8&
$\mathbf{ 0.929 \pm 0.000 }$ & $1.096 \pm 0.000$ & $1.170 \pm 0.000$ & $2.053 \pm 0.000$ & 3.75 & 1 \\
   &  Surgical-FT   &  $3.018 \pm 0.118$  &  $1.491 \pm 0.085$  &  $1.191 \pm 0.004$  &  $3.304 \pm 0.347$  & $5.75$ & 7&
$1.240 \pm 0.000$ & $\underline{ 1.044 \pm 0.000 }$ & $1.180 \pm 0.000$ & $\underline{ 2.009 \pm 0.000 }$ & 4.00 & 2 \\
   &  LP-FT         & $\underline{ 1.636 \pm 0.021 }$ &  $1.181 \pm 0.029$  &  $1.263 \pm 0.009$  &  $2.294 \pm 0.024$  & $4.00$ & 4&
$1.241 \pm 0.000$ & $1.085 \pm 0.000$ & $1.176 \pm 0.000$ & $2.044 \pm 0.000$ & 5.00 & 8 \\
   &  WiSE-FT       &  $2.762 \pm 0.091$  &  $1.405 \pm 0.067$  & $\mathbf{ 1.181 \pm 0.008 }$ &  $3.496 \pm 0.199$  & $4.50$ & 5&
$1.247 \pm 0.000$ & $1.099 \pm 0.000$ & $\underline{ 1.166 \pm 0.000 }$ & $2.024 \pm 0.000$ & 4.25 & 4 \\
   &  $L^2$-SP      &  $1.654 \pm 0.086$  & $\underline{ 1.178 \pm 0.022 }$ &  $1.185 \pm 0.008$  & $\mathbf{ 2.255 \pm 0.026 }$ & $2.25$ & 2&
$1.280 \pm 0.003$ & $1.107 \pm 0.002$ & $1.175 \pm 0.000$ & $\mathbf{ 1.997 \pm 0.016 }$ & 5.50 & 6 \\
   &  Feature-map   &  $1.783 \pm 0.034$  &  $1.252 \pm 0.012$  &  $1.195 \pm 0.008$  &  $2.401 \pm 0.028$  & $4.50$ & 3&
$1.267 \pm 0.110$ & $\mathbf{ 1.037 \pm 0.006 }$ & $1.170 \pm 0.143$ & $2.073 \pm 0.016$ & 4.75 & 5 \\
   &  BSS           & $\mathbf{ 1.632 \pm 0.048 }$ & $\mathbf{ 1.173 \pm 0.022 }$ & $\underline{ 1.182 \pm 0.016 }$ & $\underline{ 2.287 \pm 0.028 }$ & $1.50$ & 1&
$\underline{ 1.159 \pm 0.007 }$ & $1.100 \pm 0.002$ & $\mathbf{ 1.162 \pm 0.000 }$ & $2.060 \pm 0.009$ & 4.25 & 3 \\

\midrule
\multirow{8}{*}{Size} 
&  Full-FT       &  $3.457 \pm 0.086$  &  $1.407 \pm 0.088$  &  $1.064 \pm 0.067$  &  $3.311 \pm 0.158$  & $6.25$ & 7&
$\mathbf{ 1.499 \pm 0.000 }$ & $1.108 \pm 0.000$ & $\underline{ 0.909 \pm 0.000 }$ & $2.321 \pm 0.000$ & 3.50 & 4 \\
   &  LP            &  $3.758 \pm 0.010$  &  $1.773 \pm 0.025$  &  $0.990 \pm 0.056$  &  $4.114 \pm 0.042$  & $6.75$ & 8&
$2.025 \pm 0.000$ & $1.325 \pm 0.000$ & $0.917 \pm 0.000$ & $2.358 \pm 0.000$ & 7.50 & 8 \\
   &  Surgical-FT   &  $3.429 \pm 0.139$  &  $1.543 \pm 0.083$  &  $0.990 \pm 0.054$  &  $3.195 \pm 0.306$  & $5.25$ & 6&
$1.675 \pm 0.000$ & $1.089 \pm 0.000$ & $0.916 \pm 0.000$ & $\mathbf{ 2.271 \pm 0.000 }$ & 4.50 & 1 \\
   &  LP-FT         & $\mathbf{ 2.035 \pm 0.080 }$ &  $1.208 \pm 0.078$  &  $1.102 \pm 0.018$  &  $2.500 \pm 0.045$  & $4.00$ & 4&
$1.540 \pm 0.000$ & $1.079 \pm 0.001$ & $0.994 \pm 0.000$ & $2.347 \pm 0.001$ & 5.75 & 7 \\
   &  WiSE-FT       &  $3.527 \pm 0.112$  &  $1.392 \pm 0.062$  & $\mathbf{ 0.983 \pm 0.053 }$ &  $3.386 \pm 0.142$  & $5.00$ & 5&
$1.536 \pm 0.000$ & $1.149 \pm 0.000$ & $0.911 \pm 0.000$ & $2.321 \pm 0.000$ & 4.50 & 5 \\
   &  $L^2$-SP      & $\underline{ 2.111 \pm 0.091 }$ & $\underline{ 1.159 \pm 0.037 }$ & $\underline{ 0.988 \pm 0.032 }$ & $\underline{ 2.421 \pm 0.045 }$ & $2.00$ & 1&
$1.673 \pm 0.030$ & $\underline{ 1.072 \pm 0.002 }$ & $0.948 \pm 0.007$ & $\underline{ 2.304 \pm 0.022 }$ & 4.25 & 6 \\
   &  Feature-map   &  $2.331 \pm 0.050$  &  $1.225 \pm 0.049$  &  $1.000 \pm 0.034$  &  $2.439 \pm 0.024$  & $4.00$ & 3&
$1.594 \pm 0.010$ & $\mathbf{ 1.070 \pm 0.012 }$ & $0.915 \pm 0.001$ & $2.306 \pm 0.008$ & 3.25 & 3 \\
   &  BSS           &  $2.197 \pm 0.084$  & $\mathbf{ 1.106 \pm 0.027 }$ &  $1.019 \pm 0.033$  & $\mathbf{ 2.419 \pm 0.045 }$ & $2.75$ & 2&
$\underline{ 1.516 \pm 0.008 }$ & $1.076 \pm 0.043$ & $\mathbf{ 0.907 \pm 0.000 }$ & $2.313 \pm 0.049$ & 2.50 & 2 \\

\midrule
    \multicolumn{13}{c}{{\textbf{Fewshot-100}}}\\
  \midrule
\multirow{8}{*}{Random}
&  Full-FT       &  $1.842 \pm 0.208$   &  $1.205 \pm 0.059$   &  $1.289 \pm 0.032$   &  $2.784 \pm 0.110$   & $5.75$ & 6&
$1.121 \pm 0.000$ & $1.187 \pm 0.020$ & $\underline{ 1.259 \pm 0.000 }$ & $1.902 \pm 0.011$ & 5.00 & 6 \\
   &  LP            &  $2.391 \pm 0.044$   &  $1.623 \pm 0.011$   &  $1.279 \pm 0.007$   &  $3.176 \pm 0.093$   & $7.00$ & 8&
$\underline{ 0.912 \pm 0.000 }$ & $1.068 \pm 0.000$ & $1.286 \pm 0.000$ & $1.920 \pm 0.014$ & 4.75 & 4 \\
   &  Surgical-FT   &  $1.650 \pm 0.063$   &  $1.301 \pm 0.037$   &  $1.277 \pm 0.012$   &  $2.777 \pm 0.181$   & $5.00$ & 4&
$0.952 \pm 0.000$ & $\underline{ 1.061 \pm 0.000 }$ & $1.269 \pm 0.000$ & $\mathbf{ 1.881 \pm 0.000 }$ & 2.25 & 2 \\
   &  LP-FT         & $\underline{ 1.540 \pm 0.123  }$ &  $1.234 \pm 0.030$   &  $1.350 \pm 0.016$   &  $2.203 \pm 0.030$   & $4.50$ & 7&
$1.061 \pm 0.005$ & $1.126 \pm 0.000$ & $1.290 \pm 0.011$ & $1.918 \pm 0.005$ & 6.00 & 7 \\
   &  WiSE-FT       &  $1.790 \pm 0.147$   &  $1.207 \pm 0.058$   &  $1.282 \pm 0.017$   &  $2.842 \pm 0.123$   & $5.50$ & 5&
$1.064 \pm 0.000$ & $1.121 \pm 0.050$ & $\mathbf{ 1.258 \pm 0.000 }$ & $1.905 \pm 0.015$ & 3.75 & 3 \\
   &  $L^2$-SP      & $\mathbf{ 1.486 \pm 0.105  }$ & $\mathbf{ 1.190 \pm 0.038  }$ & $\mathbf{ 1.267 \pm 0.007  }$ &  $2.207 \pm 0.046$   & $1.75$ & 1&
$1.109 \pm 0.082$ & $1.094 \pm 0.007$ & $1.276 \pm 0.000$ & $1.916 \pm 0.022$ & 5.00 & 5 \\
   &  Feature-map   &  $1.557 \pm 0.034$   &  $1.252 \pm 0.007$   & $\underline{ 1.269 \pm 0.002  }$ & $\mathbf{ 2.130 \pm 0.020  }$ & $3.25$ & 2&
$\mathbf{ 0.897 \pm 0.009 }$ & $\mathbf{ 1.053 \pm 0.007 }$ & $1.273 \pm 0.000$ & $\underline{ 1.881 \pm 0.011 }$ & 1.75 & 1 \\
   &  BSS           &  $1.543 \pm 0.044$   & $\mathbf{ 1.190 \pm 0.031  }$ &  $1.285 \pm 0.011$   & $\underline{ 2.170 \pm 0.028  }$ & $3.25$ & 3&
$1.159 \pm 0.012$ & $1.129 \pm 0.022$ & $1.276 \pm 0.004$ & $2.036 \pm 0.139$ & 7.00 & 8 \\

   \midrule
   \multirow{8}{*}{Scaffold}
   &  Full-FT       &  $2.036 \pm 0.119$   &  $1.108 \pm 0.017$   &  $1.205 \pm 0.050$   &  $2.942 \pm 0.208$   & $5.75$ & 6&
$1.238 \pm 0.000$ & $1.027 \pm 0.000$ & $1.187 \pm 0.000$ & $1.986 \pm 0.019$ & 6.75 & 7 \\
   &  LP            &  $2.906 \pm 0.093$   &  $1.389 \pm 0.008$   &  $1.180 \pm 0.017$   &  $3.635 \pm 0.051$   & $6.75$ & 8&
$1.184 \pm 0.013$ & $0.998 \pm 0.000$ & $1.163 \pm 0.000$ & $1.935 \pm 0.000$ & 3.25 & 3 \\
   &  Surgical-FT   &  $1.956 \pm 0.170$   &  $1.190 \pm 0.027$   &  $1.183 \pm 0.016$   &  $2.848 \pm 0.120$   & $5.50$ & 5&
$\underline{ 1.121 \pm 0.000 }$ & $\underline{ 0.977 \pm 0.000 }$ & $1.172 \pm 0.000$ & $\mathbf{ 1.914 \pm 0.000 }$ & 2.50 & 1 \\
   &  LP-FT         &  $1.775 \pm 0.178$   &  $1.103 \pm 0.024$   &  $1.288 \pm 0.012$   &  $2.310 \pm 0.034$   & $4.75$ & 7&
$1.210 \pm 0.001$ & $1.062 \pm 0.003$ & $1.206 \pm 0.000$ & $\underline{ 1.918 \pm 0.002 }$ & 6.00 & 8 \\
   &  WiSE-FT       &  $2.052 \pm 0.082$   &  $1.112 \pm 0.023$   &  $1.188 \pm 0.027$   &  $3.049 \pm 0.246$   & $6.25$ & 4&
$1.199 \pm 0.000$ & $1.002 \pm 0.000$ & $\underline{ 1.160 \pm 0.000 }$ & $1.988 \pm 0.028$ & 4.50 & 5 \\
   &  $L^2$-SP      & $\mathbf{ 1.559 \pm 0.047  }$ & $\mathbf{ 1.069 \pm 0.044  }$ & $\underline{ 1.166 \pm 0.004  }$ &  $2.227 \pm 0.036$   & $1.75$ & 1&
$1.210 \pm 0.030$ & $0.999 \pm 0.035$ & $1.176 \pm 0.015$ & $2.000 \pm 0.009$ & 5.75 & 6 \\
   &  Feature-map   & $\underline{ 1.576 \pm 0.028  }$ &  $1.123 \pm 0.009$   &  $1.181 \pm 0.005$   & $\underline{ 2.216 \pm 0.014  }$ & $3.50$ & 3&
$\mathbf{ 1.106 \pm 0.025 }$ & $\mathbf{ 0.957 \pm 0.008 }$ & $\mathbf{ 1.159 \pm 0.003 }$ & $2.047 \pm 0.008$ & 2.75 & 2 \\
   &  BSS           &  $1.680 \pm 0.098$   & $\underline{ 1.081 \pm 0.019  }$ & $\mathbf{ 1.163 \pm 0.004  }$ & $\mathbf{ 2.212 \pm 0.018  }$ & $1.75$ & 2&
$1.169 \pm 0.035$ & $1.025 \pm 0.000$ & $1.170 \pm 0.014$ & $1.938 \pm 0.030$ & 4.25 & 4 \\

   \midrule
\multirow{8}{*}{Size} 
&  Full-FT       &  $2.527 \pm 0.152$   &  $1.113 \pm 0.054$   &  $1.022 \pm 0.046$   &  $2.587 \pm 0.100$   & $6.25$ & 7&
$1.675 \pm 0.003$ & $1.132 \pm 0.000$ & $0.909 \pm 0.000$ & $2.317 \pm 0.000$ & 5.25 & 6 \\
   &  LP            &  $3.020 \pm 0.061$   &  $1.492 \pm 0.039$   &  $0.951 \pm 0.011$   &  $3.408 \pm 0.041$   & $6.75$ & 8&
$1.740 \pm 0.000$ & $1.245 \pm 0.000$ & $0.934 \pm 0.000$ & $2.355 \pm 0.000$ & 7.75 & 8 \\
   &  Surgical-FT   &  $2.435 \pm 0.119$   &  $1.119 \pm 0.037$   &  $0.970 \pm 0.020$   &  $2.607 \pm 0.040$   & $6.25$ & 6&
$1.501 \pm 0.000$ & $1.091 \pm 0.000$ & $\mathbf{ 0.902 \pm 0.000 }$ & $\mathbf{ 2.241 \pm 0.000 }$ & 2.50 & 1 \\
   &  LP-FT         &  $1.937 \pm 0.120$   & $\mathbf{ 1.050 \pm 0.052  }$ &  $1.045 \pm 0.012$   &  $2.506 \pm 0.042$   & $4.25$ & 5&
$1.662 \pm 0.009$ & $1.228 \pm 0.002$ & $0.939 \pm 0.003$ & $2.310 \pm 0.005$ & 6.50 & 7 \\
   &  WiSE-FT       &  $2.580 \pm 0.096$   &  $1.086 \pm 0.051$   &  $0.962 \pm 0.043$   &  $2.556 \pm 0.089$   & $5.00$ & 4&
$1.605 \pm 0.001$ & $1.159 \pm 0.000$ & $\underline{ 0.907 \pm 0.000 }$ & $2.300 \pm 0.000$ & 4.00 & 5 \\
   &  $L^2$-SP      & $\underline{ 1.860 \pm 0.183  }$ & $\underline{ 1.063 \pm 0.006  }$ & $\mathbf{ 0.931 \pm 0.007  }$ & $\underline{ 2.436 \pm 0.043  }$ & $1.75$ & 1&
$\underline{ 1.474 \pm 0.031 }$ & $\underline{ 1.047 \pm 0.097 }$ & $0.915 \pm 0.008$ & $\underline{ 2.256 \pm 0.020 }$ & 2.75 & 3 \\
   &  Feature-map   &  $1.921 \pm 0.086$   &  $1.098 \pm 0.036$   & $\underline{ 0.936 \pm 0.009  }$ & $\mathbf{ 2.374 \pm 0.011  }$ & $2.75$ & 2&
$1.494 \pm 0.038$ & $1.085 \pm 0.012$ & $0.915 \pm 0.000$ & $2.303 \pm 0.006$ & 3.75 & 4 \\
   &  BSS           & $\mathbf{ 1.854 \pm 0.109  }$ &  $1.075 \pm 0.032$   &  $0.962 \pm 0.017$   &  $2.444 \pm 0.014$   & $3.00$ & 3&
$\mathbf{ 1.325 \pm 0.017 }$ & $\mathbf{ 1.011 \pm 0.045 }$ & $0.909 \pm 0.002$ & $2.322 \pm 0.002$ & 3.00 & 2 \\

\midrule
  \multicolumn{13}{c}{{\textbf{Fewshot-500}}}\\
  \midrule
\multirow{8}{*}{Random}
&  Full-FT       &  $1.093 \pm 0.085$   &  $0.834 \pm 0.014$   &  $1.245 \pm 0.018$   &  $1.874 \pm 0.042$   & $5.00$ & 6&
$0.702 \pm 0.006$ & $0.849 \pm 0.006$ & $\underline{ 1.217 \pm 0.000 }$ & $\underline{ 1.801 \pm 0.018 }$ & 5.00 & 5 \\
   &  LP            &  $1.542 \pm 0.011$   &  $1.136 \pm 0.006$   &  $1.253 \pm 0.003$   &  $2.435 \pm 0.019$   & $8.00$ & 8&
$0.732 \pm 0.000$ & $0.829 \pm 0.000$ & $1.225 \pm 0.000$ & $1.809 \pm 0.011$ & 6.25 & 7 \\
   &  Surgical-FT   &  $1.177 \pm 0.043$   &  $0.888 \pm 0.010$   &  $1.233 \pm 0.009$   &  $1.948 \pm 0.005$   & $6.00$ & 7&
$\mathbf{ 0.643 \pm 0.000 }$ & $\underline{ 0.800 \pm 0.000 }$ & $\underline{ 1.207 \pm 0.000 }$ & $\mathbf{ 1.775 \pm 0.000 }$ & 1.50 & 1 \\
   &  LP-FT         &  $1.001 \pm 0.020$   &  $0.838 \pm 0.020$   &  $1.244 \pm 0.011$   &  $1.850 \pm 0.019$   & $4.00$ & 5&
$0.664 \pm 0.001$ & $0.837 \pm 0.019$ & $\mathbf{ 1.204 \pm 0.000 }$ & $1.809 \pm 0.019$ & 3.75 & 2 \\
   &  WiSE-FT       &  $1.076 \pm 0.074$   & $\underline{ 0.833 \pm 0.007  }$ &  $1.236 \pm 0.012$   &  $1.898 \pm 0.051$   & $4.25$ & 4&
$\underline{ 0.661 \pm 0.009 }$ & $0.848 \pm 0.005$ & $1.207 \pm 0.000$ & $1.802 \pm 0.025$ & 3.50 & 3 \\
   &  $L^2$-SP      & $\underline{ 0.992 \pm 0.034  }$ &  $0.838 \pm 0.009$   & $\underline{ 1.225 \pm 0.005  }$ & $\underline{ 1.839 \pm 0.024  }$ & $2.75$ & 1&
$0.714 \pm 0.041$ & $0.827 \pm 0.011$ & $1.223 \pm 0.006$ & $1.830 \pm 0.014$ & 5.75 & 8 \\
   &  Feature-map   &  $1.070 \pm 0.020$   &  $0.948 \pm 0.010$   & $\mathbf{ 1.216 \pm 0.002  }$ &  $1.904 \pm 0.003$   & $4.50$ & 3&
$0.671 \pm 0.014$ & $\mathbf{ 0.791 \pm 0.007 }$ & $1.210 \pm 0.002$ & $1.849 \pm 0.002$ & 4.25 & 4 \\
   &  BSS           & $\mathbf{ 0.990 \pm 0.046  }$ & $\mathbf{ 0.829 \pm 0.018  }$ &  $1.231 \pm 0.009$   & $\mathbf{ 1.835 \pm 0.023  }$ & $1.50$ & 2&
$0.715 \pm 0.035$ & $0.816 \pm 0.015$ & $1.228 \pm 0.003$ & $1.808 \pm 0.009$ & 5.50 & 6 \\

\midrule
\multirow{8}{*}{Scaffold}
&  Full-FT       &  $1.434 \pm 0.044$   &  $0.885 \pm 0.028$   &  $1.186 \pm 0.017$   &  $1.910 \pm 0.022$   & $5.00$ & 6&
$1.025 \pm 0.011$ & $0.856 \pm 0.016$ & $1.125 \pm 0.000$ & $1.808 \pm 0.023$ & 5.50 & 6 \\
   &  LP            &  $2.047 \pm 0.020$   &  $1.026 \pm 0.003$   &  $1.168 \pm 0.005$   &  $2.572 \pm 0.018$   & $7.25$ & 8&
$0.929 \pm 0.003$ & $0.841 \pm 0.000$ & $1.151 \pm 0.000$ & $1.787 \pm 0.000$ & 4.50 & 3 \\
   &  Surgical-FT   &  $\mathbf{ 1.323 \pm 0.053 }$ &  $0.940 \pm 0.016$   &  $1.159 \pm 0.014$   &  $1.920 \pm 0.010$   & $4.50$ & 5&
$\underline{ 0.943 \pm 0.000 }$ & $\underline{ 0.812 \pm 0.000 }$ & $\underline{ 1.138 \pm 0.000 }$ & $\underline{ 1.793 \pm 0.000 }$ & 2.00 & 2 \\
   &  LP-FT         &  $1.394 \pm 0.025$   &  $0.888 \pm 0.017$   &  $1.204 \pm 0.015$   &  $1.876 \pm 0.024$   & $5.00$ & 7&
$0.962 \pm 0.004$ & $0.847 \pm 0.001$ & $1.133 \pm 0.003$ & $1.809 \pm 0.019$ & 4.75 & 4 \\
   &  WiSE-FT       &  $1.423 \pm 0.032$   &  $0.885 \pm 0.023$   &  $1.170 \pm 0.014$   &  $1.926 \pm 0.035$   & $5.50$ & 4&
$0.995 \pm 0.013$ & $0.851 \pm 0.010$ & $1.123 \pm 0.000$ & $1.807 \pm 0.020$ & 5.50 & 5 \\
   &  $L^2$-SP      &  $1.375 \pm 0.030$   &  $\mathbf{ 0.879 \pm 0.008 }$ &  $\mathbf{ 1.139 \pm 0.001 }$ &  $\underline{ 1.870 \pm 0.032 }$ & $1.75$ & 1&
$0.996 \pm 0.044$ & $0.861 \pm 0.014$ & $1.122 \pm 0.005$ & $1.828 \pm 0.004$ & 5.75 & 7 \\
   &  Feature-map   &  $1.453 \pm 0.028$   &  $0.903 \pm 0.004$   &  $1.154 \pm 0.003$   &  $1.913 \pm 0.016$   & $5.25$ & 3&
$\mathbf{ 0.881 \pm 0.005 }$ & $\mathbf{ 0.808 \pm 0.003 }$ & $\mathbf{ 1.145 \pm 0.000 }$ & $\mathbf{ 1.747 \pm 0.014 }$ & 1.00 & 1 \\
   &  BSS           &  $\underline{ 1.367 \pm 0.043 }$ &  $\underline{ 0.881 \pm 0.024 }$ &  $\underline{ 1.150 \pm 0.020 }$ &  $\mathbf{ 1.866 \pm 0.018 }$ & $1.75$ & 2&
$0.976 \pm 0.029$ & $0.859 \pm 0.009$ & $1.158 \pm 0.010$ & $1.817 \pm 0.013$ & 3.00 & 8 \\

\midrule
\multirow{8}{*}{Size} 
&  Full-FT       &  $1.797 \pm 0.088$   &  $0.793 \pm 0.019$   &  $0.997 \pm 0.019$   &  $2.353 \pm 0.033$   & $5.50$ & 7&
$1.198 \pm 0.000$ & $0.863 \pm 0.001$ & $0.926 \pm 0.000$ & $2.235 \pm 0.011$ & 6.00 & 7 \\
   &  LP            &  $2.581 \pm 0.049$   &  $1.030 \pm 0.004$   &  $0.943 \pm 0.005$   &  $2.990 \pm 0.030$   & $6.75$ & 8&
$1.375 \pm 0.000$ & $0.934 \pm 0.000$ & $0.938 \pm 0.000$ & $2.300 \pm 0.000$ & 8.00 & 8 \\
   &  Surgical-FT   & $\mathbf{ 1.540 \pm 0.078  }$ &  $0.846 \pm 0.011$   &  $0.944 \pm 0.010$   &  $2.403 \pm 0.038$   & $4.50$ & 4&
$1.289 \pm 0.000$ & $\underline{ 0.820 \pm 0.000 }$ & $0.917 \pm 0.000$ & $\mathbf{ 2.198 \pm 0.000 }$ & 3.75 & 4 \\
   &  LP-FT         &  $1.717 \pm 0.077$   &  $0.809 \pm 0.004$   &  $0.956 \pm 0.014$   & $\underline{ 2.287 \pm 0.043  }$ & $4.50$ & 5&
$\underline{ 1.147 \pm 0.000 }$ & $0.855 \pm 0.015$ & $0.907 \pm 0.005$ & $2.220 \pm 0.031$ & 3.25 & 3 \\
   &  WiSE-FT       &  $1.874 \pm 0.084$   &  $0.805 \pm 0.012$   &  $0.955 \pm 0.019$   &  $2.363 \pm 0.035$   & $5.50$ & 6&
$1.189 \pm 0.000$ & $0.873 \pm 0.001$ & $0.908 \pm 0.000$ & $2.233 \pm 0.007$ & 4.75 & 5 \\
   &  $L^2$-SP      &  $1.592 \pm 0.089$   & $\underline{ 0.788 \pm 0.014  }$ & $\underline{ 0.930 \pm 0.008  }$ &  $2.297 \pm 0.014$   & $2.75$ & 1&
$\mathbf{ 1.114 \pm 0.038 }$ & $\mathbf{ 0.805 \pm 0.030 }$ & $\underline{ 0.903 \pm 0.009 }$ & $2.220 \pm 0.012$ & 1.75 & 1 \\
   &  Feature-map   & $\underline{ 1.580 \pm 0.070  }$ &  $0.873 \pm 0.016$   & $\mathbf{ 0.921 \pm 0.002  }$ & $\mathbf{ 2.286 \pm 0.036  }$ & $2.75$ & 2&
$1.241 \pm 0.116$ & $0.833 \pm 0.010$ & $0.917 \pm 0.001$ & $2.236 \pm 0.024$ & 5.50 & 6 \\
   &  BSS           &  $1.617 \pm 0.117$   & $\mathbf{ 0.783 \pm 0.018  }$ &  $0.957 \pm 0.007$   &  $2.295 \pm 0.038$   & $3.75$ & 3&
$1.189 \pm 0.029$ & $0.829 \pm 0.022$ & $\mathbf{ 0.901 \pm 0.007 }$ & $\underline{ 2.219 \pm 0.000 }$ & 2.25 & 2 \\

   \midrule[2pt]
   
\end{tabular}%
\end{sc}
\end{small}
\end{adjustbox}
\end{center}

\end{table*}

\begin{table*}[t]
\caption{Robust fine-tuning performance on 8 \hlb{\textbf{Classification}}{30}
datasets (AUC metrics) in the \hlr{\textbf{Non-Fewshot}}{30} setting, evaluated across 3 dataset splits (\textsc{Random, Scaffold, Size}), over \hlg{\textsc{\textbf{GraphMAE}}}{30} and \hlg{\textsc{\textbf{GraphGPS}}}{30} models. \textsc{Avg, Avg-F, Avg-R} denote the average AUC, average AUC without max and min values, and average rank over all the  datasets for each method, respectively. Standard deviations across five replicates are shown in parentheses. We \textbf{bold} and \underline{underline} the best and second-best performances in each scenario. 
}\label{tab-mae-gps-clf-non}
\begin{center}
\begin{adjustbox}{width = 1.0\textwidth}
\begin{small}
\begin{sc}
\begin{tabular}{ccccccccccccc}

\toprule[2pt]

 Split & Methods  &   ClinTox &  BBBP  &     Bace    & HIV     & MUV & Sider &  Tox21  &   ToxCast & Avg & Avg-F & Avg-R\\ \midrule
\multicolumn{13}{c}{{\textbf{Self-supervised Pre-training (GraphMAE)}}}  \\ \midrule
\multirow{8}{*}{Random} 

&  Full-FT   &  $83.22  \pm 2.07$  & $\underline{ 94.70  \pm 0.32 }$ &  $89.26  \pm 0.40$  &  $85.31  \pm 0.29$  &  $80.71  \pm 0.58$  &  $61.53  \pm 0.48$  &  $82.35  \pm 0.15$  &  $73.01  \pm 0.16$ & $81.26$ & $82.31$ & $4.00$ \\
 &  LP     &  $78.82  \pm 1.55$  &  $83.16  \pm 0.58$  &  $77.65  \pm 1.27$  &  $74.45  \pm 0.31$  &  $78.54  \pm 1.16$  &  $61.51  \pm 0.35$  &  $73.57  \pm 0.16$  &  $66.96  \pm 0.16$ & $74.33$ & $75.00$ & $7.50$ \\
 &  Surgical-FT   &  $83.85  \pm 1.52$  &  $92.11  \pm 0.35$  &  $86.77  \pm 0.09$  &  $84.56  \pm 0.30$  & $\mathbf{ 82.71  \pm 0.81 }$ &  $61.79  \pm 0.19$  &  $79.90  \pm 0.14$  &  $71.51  \pm 0.21$ & $80.40$ & $81.55$ & $4.50$ \\
 &  LP-FT   & $\mathbf{ 88.09  \pm 1.04 }$ &  $94.68  \pm 0.19$  & $\underline{ 89.58  \pm 0.23 }$ & $\mathbf{ 86.06  \pm 0.43 }$ &  $80.75  \pm 1.53$  &  $61.69  \pm 0.26$  & $\mathbf{ 82.50  \pm 0.21 }$ & $\mathbf{ 73.66  \pm 0.07 }$ & $82.13$ & $83.44$ & $2.25$ \\
 &  WiSE-FT  &  $80.01  \pm 4.00$  &  $93.04  \pm 0.46$  & $\mathbf{ 90.15  \pm 0.50 }$ &  $85.42  \pm 0.52$  & $\underline{ 82.07  \pm 2.10 }$ & $\underline{ 62.18  \pm 0.49 }$ &  $81.55  \pm 0.43$  &  $72.48  \pm 0.26$ & $80.86$ & $81.95$ & $3.38$ \\
 &  L2-SP  &  $83.39  \pm 1.88$  &  $93.89  \pm 0.28$  &  $88.70  \pm 0.10$  &  $80.22  \pm 0.17$  &  $73.35  \pm 1.54$  & $\mathbf{ 62.36  \pm 0.43 }$ &  $77.45  \pm 0.47$  &  $68.71  \pm 0.31$ & $78.51$ & $78.64$ & $5.00$ \\
 &  Feature-map  &  $73.08  \pm 0.89$  &  $85.36  \pm 0.46$  &  $75.88  \pm 0.75$  &  $77.04  \pm 0.26$  &  $79.53  \pm 1.25$  &  $62.06  \pm 0.32$  &  $75.36  \pm 0.13$  &  $65.69  \pm 0.24$  & $74.25$ & $74.43$ & $6.75$ \\
 &  BSS  & $\underline{ 83.98  \pm 3.00 }$ & $\mathbf{ 94.85  \pm 0.31 }$ &  $89.31  \pm 0.21$  & $\underline{ 86.05  \pm 0.40 }$ &  $80.55  \pm 0.75$  &  $61.92  \pm 0.21$  & $\underline{ 82.48  \pm 0.28 }$ & $\underline{ 73.22  \pm 0.07}$ & $81.54$ & $82.60$ & $2.62$ \\
\midrule
\multirow{8}{*}{Scaffold}

&  Full-FT   &  $74.74  \pm 0.93$  &  $66.35  \pm 0.65$  &  $80.33  \pm 0.37$  & $\underline{ 77.22  \pm 0.38 }$ &  $77.47  \pm 1.33$  &  $60.98  \pm 0.19$  & $\underline{ 76.18  \pm 0.31 }$ & $\underline{ 64.27  \pm 0.36 }$ & $72.19$ & $72.70$ & $3.88$ \\
 &  LP     & $71.34  \pm 1.48$  &  $64.36  \pm 0.24$  &  $61.70  \pm 7.34$  &  $70.62  \pm 0.64$  & $\underline{ 79.13  \pm 1.20 }$ &  $58.23  \pm 1.29$  &  $70.89  \pm 0.10$  &  $60.03  \pm 0.13$ & $67.04$ & $66.49$ & $6.75$ \\
 &  Surgical-FT   &  $71.88  \pm 1.07$  &  $66.81  \pm 0.29$  &  $80.24  \pm 0.90$  &  $76.90  \pm 0.30$  & $\mathbf{ 79.20  \pm 0.50 }$ & $\mathbf{ 64.00  \pm 0.09 }$ &  $74.18  \pm 0.40$  &  $62.60  \pm 0.27$ & $71.98$ & $72.16$ & $4.12$ \\
 &  LP-FT   &  $74.88  \pm 2.00$  &  $67.39  \pm 0.55$  &  $80.67  \pm 0.57$  & $\mathbf{ 77.97  \pm 0.38 }$ &  $75.13  \pm 1.06$  &  $60.76  \pm 0.45$  & $\underline{ 76.18  \pm 0.20 }$ & $\mathbf{ 64.29  \pm 0.23 }$ & $72.16$ & $72.64$ & $3.25$ \\
 &  WiSE-FT  & $\mathbf{ 77.30  \pm 5.30 }$ & $\mathbf{ 69.29  \pm 2.34 }$ & $\mathbf{ 82.16  \pm 1.50 }$ &  $76.75  \pm 0.69$  &  $77.76  \pm 1.55$  &  $59.76  \pm 0.86$  &  $74.99  \pm 0.44$  &  $63.61  \pm 0.34$  & $72.70$ & $73.28$ & $3.25$ \\
 &  L2-SP  &  $73.40  \pm 0.45$  &  $67.39  \pm 0.90$  &  $80.36  \pm 0.92$  &  $74.63  \pm 0.44$  &  $73.20  \pm 0.90$  & $\underline{ 63.40  \pm 0.29 }$ &  $73.16  \pm 0.14$  &  $61.29  \pm 0.38$ & $70.85$ & $70.86$ & $5.00$ \\
 &  Feature-map  &  $64.74  \pm 0.62$  &  $62.46  \pm 0.26$  &  $69.22  \pm 2.06$  &  $72.34  \pm 0.58$  &  $75.63  \pm 0.54$  &  $57.13  \pm 1.08$  &  $71.25  \pm 0.13$  &  $57.78  \pm 0.26$ & $66.32$ & $66.30$ & $7.38$ \\
 &  BSS  & $\underline{ 75.80  \pm 1.11 }$ & $\underline{ 67.46  \pm 1.35 }$ & $\underline{ 80.82  \pm 0.62 }$ &  $77.10  \pm 0.77$  &  $78.53  \pm 1.03$  &  $62.29  \pm 0.51$  & $\mathbf{ 76.45  \pm 0.24 }$ &  $64.03  \pm 0.09$ & $72.81$ & $73.23$ & $2.38$ \\
\midrule
\multirow{8}{*}{Size}

&  Full-FT   &  $56.52  \pm 0.81$  &  $80.05  \pm 2.01$  &  $59.94  \pm 1.83$  &  $77.21  \pm 0.94$  &  $74.64  \pm 1.72$  &  $53.04  \pm 0.74$  &  $70.87  \pm 0.24$  &  $60.80  \pm 0.50$  & $66.63$ & $66.66$ & $4.62$ \\
 &  LP     &  $57.44  \pm 0.94$  &  $73.52  \pm 1.68$  &  $51.46  \pm 0.97$  &  $73.91  \pm 0.89$  &  $65.97  \pm 3.36$  &  $51.84  \pm 0.31$  &  $67.56  \pm 0.10$  &  $57.49  \pm 0.11$  & $62.40$ & $62.30$ & $7.38$ \\
 &  Surgical-FT   &  $57.47  \pm 1.16$  & $\underline{ 81.96  \pm 0.78 }$ &  $55.85  \pm 2.81$  & $\mathbf{ 80.48  \pm 0.18 }$ & $\underline{ 75.86  \pm 2.96 }$ & $\underline{ 54.32  \pm 0.43 }$ & $\underline{ 71.19  \pm 0.30 }$ &  $59.45  \pm 0.18$ & $67.07$ & $66.72$ & $3.12$ \\
 &  LP-FT   &  $56.35  \pm 0.62$  &  $76.80  \pm 2.24$  & $\underline{ 61.61  \pm 1.01 }$ &  $77.14  \pm 0.69$  & $\mathbf{ 79.10  \pm 0.89 }$ &  $52.69  \pm 0.35$  & $\mathbf{ 71.33  \pm 0.26 }$ & $\underline{ 60.98  \pm 0.27}$ & $67.00$ & $67.37$ & $4.00$ \\
 &  WiSE-FT  & $\mathbf{ 59.25  \pm 3.49 }$ & $\mathbf{ 82.99  \pm 1.91 }$ &  $61.16  \pm 2.31$  &  $75.90  \pm 1.94$  &  $75.09  \pm 3.95$  & $\mathbf{ 55.74  \pm 1.28 }$ &  $70.94  \pm 0.42$  & $\mathbf{ 61.53  \pm 0.56}$ & $67.83$ & $67.31$ & $2.50$ \\
 &  L2-SP  & $\underline{ 59.11  \pm 0.88 }$ &  $80.40  \pm 1.50$  &  $61.10  \pm 1.54$  &  $76.67  \pm 1.61$  &  $65.11  \pm 0.75$  &  $53.81  \pm 0.72$  &  $68.96  \pm 0.47$  &  $57.85  \pm 0.36$ & $65.38$ & $64.80$ & $4.88$ \\
 &  Feature-map  &  $59.02  \pm 0.89$  &  $77.60  \pm 0.45$  &  $43.17  \pm 0.32$  & $\underline{ 79.17  \pm 0.23 }$ &  $73.54  \pm 0.29$  &  $52.23  \pm 0.32$  &  $68.74  \pm 0.09$  &  $53.39  \pm 0.51$  & $63.36$ & $64.09$ & $5.75$ \\
 &  BSS  &  $58.58  \pm 1.31$  &  $80.86  \pm 1.92$  & $\mathbf{ 61.96  \pm 2.00 }$ &  $79.14  \pm 0.79$  &  $73.35  \pm 1.27$  &  $53.14  \pm 0.63$  &  $70.76  \pm 0.37$  &  $60.62  \pm 0.35$ & $67.30$ & $67.40$ & $3.75$ \\

\midrule
\multicolumn{13}{c}{{\textbf{Supervised Pre-training (GraphGPS)}}}  \\ \midrule

\multirow{8}{*}{Random}
&  Full-FT     &  $99.77 \pm 0.01$ &  $99.99 \pm 0.01$ &  $\mathbf{ 100.00 \pm 0.00 }$ &  $84.80 \pm 0.33$ &  $57.06 \pm 0.00$ &  $87.13 \pm 0.39$ &  $87.17 \pm 0.48$ &  $86.90 \pm 0.17$ &  $87.85$ &  $90.96$ &  $4.00$\\
&  LP          &  $99.48 \pm 0.04$ &  $86.96 \pm 0.40$ &  $80.94 \pm 0.45$ &  $86.70 \pm 0.42$ &  $\underline{ 63.97 \pm 0.80 }$ &  $84.77 \pm 0.08$ &  $82.70 \pm 0.14$ &  $83.93 \pm 0.04$ &  $83.68$ &  $84.33$ &  $5.50$\\
&  Surgical-FT &  $99.65 \pm 0.05$ &  $99.16 \pm 0.00$ &  $98.14 \pm 0.04$ &  $86.58 \pm 0.03$ &  $60.52 \pm 0.64$ &  $47.74 \pm 0.95$ &  $51.53 \pm 0.00$ &  $51.71 \pm 0.00$ &  $74.38$ &  $74.61$ &  $5.88$\\
&  LP-FT       &  $99.54 \pm 0.14$ &  $89.67 \pm 5.14$ &  $84.88 \pm 7.57$ &  $85.71 \pm 1.11$ &  $63.96 \pm 0.80$ &  $85.97 \pm 2.43$ &  $83.98 \pm 2.45$ &  $84.48 \pm 1.09$ &  $84.77$ &  $85.78$ &  $5.12$\\
&  WiSE-FT     &  $97.04 \pm 1.00$ &  $58.14 \pm 3.98$ &  $68.29 \pm 2.24$ &  $67.14 \pm 4.36$ &  $49.94 \pm 0.01$ &  $80.52 \pm 0.07$ &  $67.81 \pm 0.12$ &  $77.50 \pm 0.03$ &  $70.80$ &  $69.90$ &  $7.62$\\
&  $L^2$-SP    &  $\mathbf{ 99.84 \pm 0.03 }$ &  $\mathbf{ 100.00 \pm 0.00 }$ &  $\underline{ 100.00 \pm 0.00 }$ &  $\underline{ 97.75 \pm 0.07 }$ &  $\mathbf{ 74.51 \pm 1.12 }$ &  $\mathbf{ 92.16 \pm 0.44 }$ &  $\mathbf{ 92.28 \pm 0.46 }$ &  $\mathbf{ 89.79 \pm 0.07 }$ &  $93.29$ &  $95.30$ &  $1.25$\\
&  Feature-map &  $\underline{ 99.79 \pm 0.09 }$ &  $\underline{ 100.00 \pm 0.00 }$ &  $100.00 \pm 0.00$ &  $\mathbf{ 99.42 \pm 0.01 }$ &  $53.07 \pm 0.82$ &  $\underline{ 91.64 \pm 0.06 }$ &  $\underline{ 91.61 \pm 0.16 }$ &  $\underline{ 89.39 \pm 0.06 }$ &  $90.62$ &  $95.31$ &  $2.62$\\
&  BSS         &  $99.77 \pm 0.00$ &  $100.00 \pm 0.00$ &  $100.00 \pm 0.00$ &  $84.87 \pm 0.02$ &  $58.93 \pm 3.25$ &  $87.61 \pm 0.05$ &  $87.52 \pm 0.10$ &  $86.75 \pm 0.05$ &  $88.18$ &  $91.09$ &  $4.00$\\
\midrule
\multirow{8}{*}{Scaffold}
&  Full-FT     &  $99.76 \pm 0.04$ &  $99.99 \pm 0.01$ &  $\mathbf{ 100.00 \pm 0.00 }$ &  $83.67 \pm 1.61$ &  $57.08 \pm 1.77$ &  $87.26 \pm 0.15$ &  $87.16 \pm 0.21$ &  $86.71 \pm 0.12$ &  $87.70$ &  $90.76$ &  $4.12$\\
&  LP          &  $99.47 \pm 0.04$ &  $86.84 \pm 0.49$ &  $81.04 \pm 0.53$ &  $86.66 \pm 0.44$ &  $\underline{ 63.98 \pm 0.82 }$ &  $84.74 \pm 0.08$ &  $82.70 \pm 0.14$ &  $83.93 \pm 0.04$ &  $83.67$ &  $84.32$ &  $5.75$\\
&  Surgical-FT &  $99.64 \pm 0.08$ &  $99.33 \pm 0.14$ &  $98.14 \pm 0.06$ &  $87.61 \pm 0.63$ &  $61.75 \pm 0.39$ &  $76.46 \pm 1.75$ &  $72.53 \pm 1.99$ &  $55.58 \pm 0.35$ &  $81.38$ &  $82.64$ &  $5.75$\\
&  LP-FT       &  $99.54 \pm 0.15$ &  $89.53 \pm 5.24$ &  $84.35 \pm 6.50$ &  $84.81 \pm 2.36$ &  $62.46 \pm 1.48$ &  $85.96 \pm 2.47$ &  $83.96 \pm 2.42$ &  $84.52 \pm 1.17$ &  $84.39$ &  $85.52$ &  $5.38$\\
&  WiSE-FT     &  $97.32 \pm 0.16$ &  $64.59 \pm 3.69$ &  $\underline{ 100.00 \pm 0.00 }$ &  $67.98 \pm 4.58$ &  $49.84 \pm 0.72$ &  $80.53 \pm 0.07$ &  $68.04 \pm 0.17$ &  $77.53 \pm 0.02$ &  $75.73$ &  $76.00$ &  $7.00$\\
&  $L^2$-SP    &  $\underline{ 99.83 \pm 0.03 }$ &  $\mathbf{ 100.00 \pm 0.00 }$ &  $100.00 \pm 0.00$ &  $\underline{ 98.35 \pm 0.43 }$ &  $\mathbf{ 74.63 \pm 0.95 }$ &  $\mathbf{ 92.33 \pm 0.21 }$ &  $\mathbf{ 92.43 \pm 0.34 }$ &  $\mathbf{ 89.85 \pm 0.17 }$ &  $93.43$ &  $95.46$ &  $1.50$\\
&  Feature-map &  $\mathbf{ 99.85 \pm 0.01 }$ &  $\underline{ 100.00 \pm 0.00 }$ &  $100.00 \pm 0.00$ &  $\mathbf{ 99.26 \pm 0.13 }$ &  $55.32 \pm 0.31$ &  $\underline{ 91.63 \pm 0.04 }$ &  $\underline{ 91.61 \pm 0.11 }$ &  $\underline{ 89.30 \pm 0.06 }$ &  $90.87$ &  $95.27$ &  $2.62$\\
&  BSS         &  $99.81 \pm 0.04$ &  $99.99 \pm 0.01$ &  $100.00 \pm 0.00$ &  $85.03 \pm 0.57$ &  $60.82 \pm 4.94$ &  $89.80 \pm 3.20$ &  $87.36 \pm 0.09$ &  $86.85 \pm 0.12$ &  $88.71$ &  $91.47$ &  $3.88$\\
\midrule
\multirow{8}{*}{Size}
&  Full-FT     &  $99.76 \pm 0.03$ &  $99.99 \pm 0.01$ &  $\mathbf{ 100.00 \pm 0.00 }$ &  $83.42 \pm 1.75$ &  $56.61 \pm 1.51$ &  $87.41 \pm 0.51$ &  $87.06 \pm 0.10$ &  $86.90 \pm 0.13$ &  $87.64$ &  $90.76$ &  $4.12$\\
&  LP          &  $99.47 \pm 0.05$ &  $86.56 \pm 0.34$ &  $80.81 \pm 0.52$ &  $86.66 \pm 0.44$ &  $64.02 \pm 0.78$ &  $84.74 \pm 0.08$ &  $82.38 \pm 0.15$ &  $83.95 \pm 0.04$ &  $83.57$ &  $84.18$ &  $5.75$\\
&  Surgical-FT &  $99.21 \pm 0.00$ &  $99.30 \pm 0.15$ &  $98.09 \pm 0.07$ &  $86.08 \pm 0.07$ &  $60.69 \pm 0.81$ &  $76.45 \pm 1.71$ &  $82.17 \pm 1.95$ &  $85.13 \pm 0.03$ &  $85.89$ &  $87.86$ &  $5.88$\\
&  LP-FT       &  $99.52 \pm 0.14$ &  $89.35 \pm 5.33$ &  $84.80 \pm 7.61$ &  $84.41 \pm 2.92$ &  $63.79 \pm 0.60$ &  $85.99 \pm 2.52$ &  $83.71 \pm 2.54$ &  $84.49 \pm 1.07$ &  $84.51$ &  $85.46$ &  $5.38$\\
&  WiSE-FT     &  $96.03 \pm 1.22$ &  $57.52 \pm 3.31$ &  $70.92 \pm 2.97$ &  $66.52 \pm 4.13$ &  $49.80 \pm 0.26$ &  $80.55 \pm 0.06$ &  $67.69 \pm 0.21$ &  $77.52 \pm 0.02$ &  $70.82$ &  $70.12$ &  $7.88$\\
&  $L^2$-SP    &  $\underline{ 99.84 \pm 0.03 }$ &  $99.99 \pm 0.01$ &  $\underline{ 100.00 \pm 0.00 }$ &  $97.87 \pm 0.09$ &  $\mathbf{ 75.36 \pm 0.79 }$ &  $\mathbf{ 92.22 \pm 0.19 }$ &  $\mathbf{ 92.55 \pm 0.60 }$ &  $\mathbf{ 90.00 \pm 0.07 }$ &  $93.48$ &  $95.41$ &  $1.88$\\
&  Feature-map &  $\mathbf{ 99.85 \pm 0.02 }$ &  $\mathbf{ 100.00 \pm 0.00 }$ &  $100.00 \pm 0.00$ &  $\mathbf{ 99.36 \pm 0.08 }$ &  $\underline{ 65.70 \pm 0.23 }$ &  $\underline{ 91.61 \pm 0.06 }$ &  $\underline{ 91.43 \pm 0.15 }$ &  $\underline{ 89.49 \pm 0.03 }$ &  $92.18$ &  $95.29$ &  $1.75$\\
&  BSS         &  $99.79 \pm 0.05$ &  $\underline{ 100.00 \pm 0.00 }$ &  $100.00 \pm 0.00$ &  $\underline{ 98.71 \pm 0.03 }$ &  $59.16 \pm 2.37$ &  $87.40 \pm 0.33$ &  $88.34 \pm 0.15$ &  $86.95 \pm 0.14$ &  $90.04$ &  $93.53$ &  $3.38$\\
 \bottomrule[2pt]
\label{table:nonfew_cls_molebert}
\end{tabular}
\end{sc}
\end{small}
\end{adjustbox}
\end{center}
\end{table*}

\begin{table*}[t]
\caption{Robust fine-tuning performance on 4 \hlb{\textbf{Regression}}{30}
datasets (RMSE metrics) in the \hlr{\textbf{Non-Fewshot}}{30} setting, evaluated across 3 dataset splits (\textsc{Random, Scaffold, Size}) over \hlg{\textsc{\textbf{GraphMAE}}}{30} and \hlg{\textsc{\textbf{GraphGPS}}}{30} models.  \textsc{Avg-R,Avg-R}$^*$ denote the average rank and the rank based on the average normalized performance over all the datasets for each method, respectively. Standard deviations across five replicates are shown in parentheses. We \textbf{bold} and \underline{underline} the best and second-best performances in each scenario.}
\label{tab-mae-gps-rgs-non}
\begin{center}
\begin{adjustbox}{width = 1\textwidth}
\begin{small}
\begin{sc}
\begin{tabular}{cccccccc|ccccccc}
\toprule[2pt]
\multirow{2}{*}{Split} &
  \multirow{2}{*}{Methods} &
  \multicolumn{6}{c}{Self-supervised Pre-training (GraphMAE)} &
  \multicolumn{5}{c}{Supervised Pre-training (GraphGPS)} \\ \cmidrule(lr){3-14}
 &
   &
  Esol &
  Lipo &
  Malaria &
  Cep &
  Avg-R &
  \multicolumn{1}{c|}{Avg-R$^*$} &
  Esol &
  Lipo &
  Malaria &
  Cep &
  Avg-R &
  \multicolumn{1}{c}{Avg-R$^*$} \\ \midrule
\multirow{8}{*}{Random} &
  Full-FT &
  $0.987 \pm 0.013$ &
  $\underline{ 0.734 \pm 0.007 }$ &
  $1.109 \pm 0.015$ &
  $1.342 \pm 0.015$ &
  $3.00$ &
  3 &
  $0.191 \pm 0.019$ &
  $0.211 \pm 0.012$ &
  $0.955 \pm 0.008$ &
  $0.587 \pm 0.000$ &
  $4.50$ &
  4 \\
 &
  LP &
  $1.394 \pm 0.012$ &
  $1.156 \pm 0.001$ &
  $1.263 \pm 0.002$ &
  $3.079 \pm 0.105$ &
  $8.00$ &
  8 &
  $0.737 \pm 0.005$ &
  $0.877 \pm 0.004$ &
  $1.031 \pm 0.003$ &
  $1.602 \pm 0.006$ &
  $6.50$ &
  6 \\
 &
  Surgical-FT &
  $1.088 \pm 0.011$ &
  $0.883 \pm 0.007$ &
  $1.120 \pm 0.012$ &
  $1.697 \pm 0.012$ &
  $6.25$ &
  6 &
  $1.565 \pm 0.313$ &
  $2.284 \pm 0.179$ &
  $0.800 \pm 0.022$ &
  $0.881 \pm 0.000$ &
  $6.00$ &
  7 \\
 &
  LP-FT &
  $\mathbf{ 0.953 \pm 0.009 }$ &
  $0.743 \pm 0.006$ &
  $\underline{ 1.096 \pm 0.009 }$ &
  $\mathbf{ 1.322 \pm 0.025 }$ &
  $1.75$ &
  1 &
  $\mathbf{ 0.139 \pm 0.016 }$ &
  $0.197 \pm 0.003$ &
  $0.925 \pm 0.007$ &
  $0.646 \pm 0.087$ &
  $3.25$ &
  3 \\
 &
  WiSE-FT &
  $1.210 \pm 0.032$ &
  $0.846 \pm 0.023$ &
  $\mathbf{ 1.060 \pm 0.008 }$ &
  $1.531 \pm 0.030$ &
  $4.50$ &
  5 &
  $2.488 \pm 0.137$ &
  $1.224 \pm 0.007$ &
  $1.187 \pm 0.001$ &
  $2.574 \pm 0.015$ &
  $7.75$ &
  8 \\
 &
  L2-SP &
  $0.995 \pm 0.024$ &
  $0.787 \pm 0.008$ &
  $1.115 \pm 0.006$ &
  $1.363 \pm 0.040$ &
  $4.25$ &
  4 &
  $\underline{ 0.169 \pm 0.009 }$ &
  $\underline{ 0.194 \pm 0.010 }$ &
  $\underline{ 0.559 \pm 0.022 }$ &
  $\underline{ 0.451 \pm 0.036 }$ &
  $2.00$ &
  2 \\
 &
  Feature-map &
  $1.297 \pm 0.007$ &
  $1.080 \pm 0.002$ &
  $1.115 \pm 0.016$ &
  $1.473 \pm 0.018$ &
  $6.25$ &
  7 &
  $\mathbf{ 0.187 \pm 0.026 }$ &
  $\mathbf{ 0.134 \pm 0.008 }$ &
  $\mathbf{ 0.243 \pm 0.009 }$ &
  $\mathbf{ 0.215 \pm 0.026 }$ &
  $1.75$ &
  1 \\
 &
  BSS &
  $\underline{ 0.975 \pm 0.019 }$ &
  $\mathbf{ 0.725 \pm 0.011 }$ &
  $1.100 \pm 0.004$ &
  $\underline{ 1.334 \pm 0.004 }$ &
  $2.00$ &
  2 &
  $\underline{ 0.177 \pm 0.013 }$ &
  $0.213 \pm 0.005$ &
  $0.921 \pm 0.013$ &
  $0.651 \pm 0.079$ &
  $4.25$ &
  5 \\ \midrule

\multirow{8}{*}{Scaffold} &
  Full-FT      &  $1.332 \pm 0.015$  &  $0.808 \pm 0.008$  &  $1.104 \pm 0.007$  &  $1.327 \pm 0.017$  &  $3.50$  &  3  &  $0.218 \pm 0.054$  &  $0.202 \pm 0.022$  &  $0.929 \pm 0.011$  &  $0.528 \pm 0.123$  &  $4.25$  &  4 \\
&  LP           &  $1.703 \pm 0.016$  &  $1.043 \pm 0.006$  &  $1.150 \pm 0.003$  &  $3.102 \pm 0.136$  &  $7.50$  &  8  &  $0.752 \pm 0.006$  &  $0.849 \pm 0.005$  &  $1.008 \pm 0.000$  &  $1.539 \pm 0.009$  &  $6.75$  &  7 \\
&  Surgical-FT  &  $1.335 \pm 0.005$  &  $0.884 \pm 0.007$  &  $1.111 \pm 0.013$  &  $1.669 \pm 0.022$  &  $5.50$  &  5  &  $1.574 \pm 0.314$  &  $0.362 \pm 0.013$  &  $0.818 \pm 0.007$  &  $0.917 \pm 0.000$  &  $5.50$  &  6 \\
&  LP-FT        &  $\mathbf{ 1.312 \pm 0.024 }$  &  $\mathbf{ 0.788 \pm 0.005 }$  &  $1.104 \pm 0.006$  &  $\underline{ 1.318 \pm 0.017 }$  &  $1.75$  &  1  &  $\mathbf{ 0.145 \pm 0.020 }$  &  $\underline{ 0.181 \pm 0.012 }$  &  $0.944 \pm 0.015$  &  $0.585 \pm 0.036$  &  $3.25$  &  3 \\
&  WiSE-FT      &  $1.617 \pm 0.031$  &  $0.891 \pm 0.009$  &  $\mathbf{ 1.077 \pm 0.004 }$  &  $1.498 \pm 0.034$  &  $5.00$  &  7  &  $2.338 \pm 0.519$  &  $1.262 \pm 0.015$  &  $1.220 \pm 0.017$  &  $2.610 \pm 0.082$  &  $8.00$  &  8 \\
&  L2-SP        &  $1.329 \pm 0.030$  &  $0.835 \pm 0.011$  &  $1.108 \pm 0.011$  &  $1.325 \pm 0.021$  &  $3.50$  &  4  &  $0.208 \pm 0.037$  &  $0.183 \pm 0.004$  &  $\underline{ 0.733 \pm 0.151 }$  &  $\underline{ 0.462 \pm 0.050 }$  &  $2.75$  &  2 \\
&  Feature-map  &  $1.551 \pm 0.013$  &  $0.994 \pm 0.004$  &  $\underline{ 1.097 \pm 0.008 }$  &  $1.415 \pm 0.030$  &  $5.00$  &  6  &  $0.194 \pm 0.009$  &  $\mathbf{ 0.142 \pm 0.004 }$  &  $\mathbf{ 0.327 \pm 0.034 }$  &  $\mathbf{ 0.232 \pm 0.026 }$  &  $1.50$  &  1 \\
&  BSS          &  $\underline{ 1.326 \pm 0.029 }$  &  $\underline{ 0.803 \pm 0.013 }$  &  $1.104 \pm 0.009$  &  $\mathbf{ 1.302 \pm 0.012 }$  &  $2.00$  &  2  &  $\underline{ 0.181 \pm 0.008 }$  &  $0.206 \pm 0.016$  &  $0.899 \pm 0.024$  &  $0.622 \pm 0.021$  &  $4.00$  &  5 \\ \midrule

\multirow{8}{*}{Size} &
  Full-FT      &  $1.822 \pm 0.099$  &  $\underline{0.814 \pm 0.013}$  &  $0.908 \pm 0.005$  &  $1.722 \pm 0.016$  &  $3.25$  &  3  &  $0.192 \pm 0.022$  &  $0.221 \pm 0.013$  &  $0.836 \pm 0.044$  &  $0.474 \pm 0.042$  &  $3.75$  &  3 \\
 &
  LP           &  $2.309 \pm 0.030$  &  $1.024 \pm 0.014$  &  $0.927 \pm 0.010$  &  $3.814 \pm 0.175$  &  $7.75$  &  8  &  $0.752 \pm 0.006$  &  $0.881 \pm 0.004$  &  $0.996 \pm 0.005$  &  $1.540 \pm 0.015$  &  $6.75$  &  7 \\
 &
  Surgical-FT  &  $1.915 \pm 0.036$  &  $0.886 \pm 0.013$  &  $0.925 \pm 0.003$  &  $2.135 \pm 0.038$  &  $6.00$  &  5  &  $1.589 \pm 0.314$  &  $0.353 \pm 0.005$  &  $0.787 \pm 0.018$  &  $0.943 \pm 0.000$  &  $5.25$  &  6 \\
 &
  LP-FT        &  $\mathbf{1.754 \pm 0.075}$  &  $\mathbf{0.795 \pm 0.005}$  &  $0.907 \pm 0.020$  &  $\mathbf{1.710 \pm 0.010}$  &  $1.75$  &  1  &  $\mathbf{0.145 \pm 0.007}$  &  $\underline{0.195 \pm 0.007}$  &  $0.902 \pm 0.067$  &  $0.575 \pm 0.058$  &  $3.25$  &  4 \\
 &
  WiSE-FT      &  $2.323 \pm 0.041$  &  $0.974 \pm 0.016$  &  $\underline{0.895 \pm 0.011}$  &  $1.982 \pm 0.039$  &  $5.50$  &  7  &  $2.264 \pm 0.336$  &  $1.226 \pm 0.006$  &  $1.189 \pm 0.002$  &  $2.683 \pm 0.151$  &  $8.00$  &  8 \\
 &
  L2-SP        &  $1.849 \pm 0.041$  &  $0.849 \pm 0.025$  &  $0.911 \pm 0.006$  &  $1.748 \pm 0.041$  &  $4.50$  &  4  &  $0.192 \pm 0.014$  &  $0.196 \pm 0.009$  &  $\underline{0.787 \pm 0.029}$  &  $\underline{0.456 \pm 0.109}$  &  $3.00$  &  2 \\
 &
  Feature-map  &  $2.136 \pm 0.030$  &  $1.007 \pm 0.015$  &  $\mathbf{0.891 \pm 0.012}$  &  $1.947 \pm 0.013$  &  $4.75$  &  6  &  $0.209 \pm 0.014$  &  $\mathbf{0.153 \pm 0.009}$  &  $\mathbf{0.354 \pm 0.007}$  &  $\mathbf{0.227 \pm 0.048}$  &  $2.00$  &  1 \\
 &
  BSS          &  $\underline{1.808 \pm 0.039}$  &  $0.818 \pm 0.020$  &  $0.899 \pm 0.006$  &  $\underline{1.712 \pm 0.021}$  &  $2.50$  &  2  &  $\underline{0.188 \pm 0.019}$  &  $0.211 \pm 0.006$  &  $0.946 \pm 0.006$  &  $0.550 \pm 0.000$  &  $4.00$  &  5 \\ 

\midrule[2pt]

\end{tabular}%
\end{sc}
\end{small}
\end{adjustbox}
\end{center}
\end{table*}

\begin{table*}[t]
\caption{Robust fine-tuning performance on 5 \hlb{\textbf{Classification}}{30}
datasets (AUC metrics) in the \hlr{\textbf{Fewshot}}{30} setting (covering \textsc{Fewshot-50, Fewshot-100, Fewshot-500}), evaluated across 3 dataset splits (\textsc{Random, Scaffold, Size}) over \hlg{\textsc{\textbf{GraphMAE}}}{30} and \hlg{\textsc{\textbf{GraphGPS}}}{30} models.  We \textbf{bold} and \underline{underline} the best and second-best performances in each scenario.}
\label{tab-mae-gps-clf-few}
\begin{center}
\begin{adjustbox}{width = 1\textwidth}
\begin{small}
\begin{sc}
\begin{tabular}{cccccccccc|cccccccc}
\toprule[2pt]
 \multirow{2}{*}{Split}&\multirow{2}{*}{Methods} & \multicolumn{8}{c}{Self-supervised Pre-training (GraphMAE)} & \multicolumn{8}{c}{Supervised Pre-training (GraphGPS)} \\
\cmidrule(lr){3-10}\cmidrule(lr){10-18}        
  &   &   ClinTox &  BBBP  &     Bace    & HIV & Sider & avg & avg-F & avg-R &   ClinTox &  BBBP  &     Bace    & HIV & Sider & avg & avg-F & avg-R  \\
\midrule
\multicolumn{18}{c}{{\textbf{Fewshot-50}}}\\
\midrule
\multirow{8}{*}{Random}
 & Full-FT       & $59.67 \pm 3.35$ & $83.04 \pm 0.39$ & $74.97 \pm 1.30$ & $\mathbf{62.63 \pm 0.92}$ & $52.52 \pm 0.19$ & $66.57$ & $65.76$ & $4.20$ 
                 & $98.37 \pm 0.28$ & $56.80 \pm 1.97$ & $59.17 \pm 2.03$ & $72.16 \pm 3.04$ & $85.04 \pm 0.42$ & $74.31$ & $72.12$ & $3.80$ \\
 & LP            & $57.56 \pm 4.09$ & $71.69 \pm 0.89$ & $72.96 \pm 0.91$ & $48.27 \pm 4.06$           & $55.09 \pm 0.22$ & $61.11$ & $61.45$ & $6.20$ 
                 & $97.63 \pm 0.56$ & $56.07 \pm 1.71$ & $56.41 \pm 0.20$ & $63.26 \pm 0.78$ & $82.77 \pm 2.22$ & $71.23$ & $67.48$ & $6.40$ \\
 & Surgical-FT   & $59.83 \pm 2.64$ & $78.37 \pm 1.06$ & $\underline{75.25 \pm 0.92}$ & $53.35 \pm 0.81$       & $54.97 \pm 0.43$ & $64.35$ & $63.35$ & $4.40$ 
                 & $97.73 \pm 0.49$ & $\underline{60.73 \pm 3.76}$ & $58.30 \pm 0.86$ & $72.63 \pm 0.13$ & $\mathbf{86.59 \pm 0.75}$ & $75.20$ & $73.32$ & $3.20$ \\
 & LP-FT         & $60.20 \pm 2.14$ & $\mathbf{84.54 \pm 0.41}$ & $\mathbf{76.82 \pm 0.34}$ & $\underline{62.24 \pm 0.58}$ & $54.41 \pm 0.32$ & $67.64$ & $66.42$ & $2.60$ 
                 & $97.37 \pm 0.89$ & $55.13 \pm 0.23$ & $56.28 \pm 0.08$ & $61.68 \pm 3.35$ & $83.52 \pm 1.15$ & $70.80$ & $67.16$ & $7.20$ \\
\end{tabular}
\end{sc}
\end{small}
\end{adjustbox}
\end{center}
\end{table*}

\begin{table*}[t]
\caption{Robust fine-tuning performance on 4 \hlb{\textbf{Regression}}{30}
datasets (RMSE metrics) in the \hlr{\textbf{Fewshot}}{30} setting (covering \textsc{Fewshot-50, Fewshot-100}, and \textsc{Fewshot-500}), evaluated across 3 dataset splits (\textsc{Random, Scaffold, Size}) over \hlg{\textsc{\textbf{GraphMAE}}}{30} and \hlg{\textsc{\textbf{GraphGPS}}}{30} models.  \textsc{Avg-R, Avg-R}$^*$ denote the average rank and the rank based on the average normalized performance over all the datasets for each evavluated method, respectively. Standard deviations across five replicates are shown in parentheses. We \textbf{bold} and \underline{underline} the best and second-best performances in each scenario. }
\label{tab-mae-gps-rgs-few}
\begin{center}
\begin{adjustbox}{width = 1\textwidth}
\begin{small}
\begin{sc}
\begin{tabular}{cccccccc|cccccc}
\midrule[2pt]
\multirow{2}{*}{Split} &
  \multirow{2}{*}{Methods} &
  \multicolumn{6}{c}{Self-supervised Pre-training (GraphMAE)} &
  \multicolumn{5}{c}{Supervised Pre-training (GraphGPS)} \\ \cmidrule(lr){3-14} 
 &
   &
  Esol &
  Lipo &
  Malaria &
  CEP &
  Avg-R &
  Avg-R$^*$ &
  Esol &
  Lipo &
  Malaria &
  CEP &
  Avg-R &
  Avg-R$^*$ \\ \midrule
  \multicolumn{14}{c}{{\textbf{Fewshot-50}}}\\
  \midrule
\multirow{8}{*}{Random} &
  FULL-FT &
  $\underline{ 1.432 \pm 0.019 }$ &
  $1.328 \pm 0.051$ &
  $1.297 \pm 0.015$ &
  $2.927 \pm 0.226$ &
  $4.25$ & 6 &
  $0.896 \pm 0.015$ &
  $1.221 \pm 0.016$ &
  $1.192 \pm 0.017$ &
  $\underline{ 2.072 \pm 0.050 }$ &
  $4.00$ & 4 \\
& LP &
  $1.646 \pm 0.027$ &
  $1.395 \pm 0.076$ &
  $1.334 \pm 0.009$ &
  $4.133 \pm 0.372$ &
  $7.50$ & 8 &
  $1.183 \pm 0.012$ &
  $1.223 \pm 0.007$ &
  $1.193 \pm 0.009$ &
  $2.249 \pm 0.016$ &
  $6.00$ & 6 \\
& Surgical-FT &
  $1.497 \pm 0.017$ &
  $\underline{ 1.303 \pm 0.051 }$ &
  $1.309 \pm 0.017$ &
  $3.300 \pm 0.406$ &
  $5.00$ & 7 &
  $3.573 \pm 0.101$ &
  $2.168 \pm 0.089$ &
  $1.203 \pm 0.010$ &
  $4.263 \pm 0.096$ &
  $7.75$ & 8 \\
& LP-FT &
  $\mathbf{ 1.386 \pm 0.022 }$ &
  $\mathbf{ 1.217 \pm 0.021 }$ &
  $1.399 \pm 0.033$ &
  $2.840 \pm 0.226$ &
  $3.75$ & 5 &
  $1.037 \pm 0.209$ &
  $\underline{ 1.199 \pm 0.041 }$ &
  $\mathbf{ 1.178 \pm 0.014 }$ &
  $2.156 \pm 0.145$ &
  $3.50$ & 5 \\
& WiSE-FT &
  $1.622 \pm 0.053$ &
  $1.343 \pm 0.010$ &
  $\mathbf{ 1.248 \pm 0.008 }$ &
  $2.385 \pm 0.026$ &
  $3.75$ & 2 &
  $2.488 \pm 0.137$ &
  $1.224 \pm 0.007$ &
  $1.180 \pm 0.019$ &
  $2.574 \pm 0.053$ &
  $6.00$ & 7 \\
& L2-SP &
  $1.444 \pm 0.027$ &
  $1.354 \pm 0.052$ &
  $1.294 \pm 0.005$ &
  $\mathbf{ 2.315 \pm 0.106 }$ &
  $3.75$ & 1 &
  $\underline{ 0.881 \pm 0.037 }$ &
  $1.203 \pm 0.022$ &
  $\underline{ 1.184 \pm 0.013 }$ &
  $2.091 \pm 0.049$ &
  $2.50$ & 3 \\
& Feature-map &
  $1.655 \pm 0.027$ &
  $1.312 \pm 0.020$ &
  $\underline{ 1.278 \pm 0.003 }$ &
  $\underline{ 2.363 \pm 0.127 }$ &
  $3.75$ & 3 &
  $0.882 \pm 0.059$ &
  $\mathbf{ 1.173 \pm 0.013 }$ &
  $1.193 \pm 0.006$ &
  $\mathbf{ 2.050 \pm 0.029 }$ &
  $2.75$ & 2 \\
& BSS &
  $1.439 \pm 0.029$ &
  $1.351 \pm 0.051$ &
  $1.294 \pm 0.005$ &
  $2.682 \pm 0.115$ &
  $4.00$ & 4 &
  $\mathbf{ 0.822 \pm 0.024 }$ &
  $1.204 \pm 0.021$ &
  $1.189 \pm 0.011$ &
  $2.109 \pm 0.036$ &
  $3.25$ & 1 \\
\midrule

\multirow{8}{*}{Scaffold} &
  FULL-FT &
  $1.717 \pm 0.028$ &
  $1.214 \pm 0.051$ &
  $1.169 \pm 0.005$ &
  $2.612 \pm 0.178$ &
  $5.25$ & 6 &
  $\mathbf{ 0.859 \pm 0.065 }$ &
  $1.219 \pm 0.025$ &
  $1.426 \pm 0.243$ &
  $\underline{ 2.100 \pm 0.031 }$ &
  $4.00$ & 3 \\
& LP &
  $2.209 \pm 0.039$ &
  $1.183 \pm 0.045$ &
  $1.170 \pm 0.004$ &
  $2.656 \pm 0.048$ &
  $6.00$ & 8 &
  $1.213 \pm 0.015$ &
  $1.223 \pm 0.006$ &
  $1.194 \pm 0.012$ &
  $2.261 \pm 0.019$ &
  $5.50$ & 6 \\
& SURGICAL-FT &
  $1.834 \pm 0.031$ &
  $1.198 \pm 0.049$ &
  $\underline{ 1.166 \pm 0.001 }$ &
  $3.142 \pm 0.589$ &
  $5.25$ & 7 &
  $3.589 \pm 0.101$ &
  $2.168 \pm 0.089$ &
  $1.204 \pm 0.010$ &
  $4.261 \pm 0.096$ &
  $7.00$ & 8 \\
& LP-FT &
  $\mathbf{ 1.642 \pm 0.026 }$ &
  $\mathbf{ 1.147 \pm 0.038 }$ &
  $1.300 \pm 0.061$ &
  $2.879 \pm 0.264$ &
  $4.25$ & 4 &
  $1.053 \pm 0.180$ &
  $\underline{ 1.198 \pm 0.043 }$ &
  $\mathbf{ 1.174 \pm 0.019 }$ &
  $2.633 \pm 0.625$ &
  $4.00$ & 4 \\
& WISE-FT &
  $2.221 \pm 0.047$ &
  $1.175 \pm 0.016$ &
  $1.166 \pm 0.002$ &
  $\mathbf{ 2.326 \pm 0.031 }$ &
  $4.00$ & 3 &
  $1.020 \pm 0.045$ &
  $1.259 \pm 0.027$ &
  $1.238 \pm 0.012$ &
  $2.123 \pm 0.035$ &
  $6.00$ & 7 \\
& L2-SP &
  $1.718 \pm 0.053$ &
  $1.200 \pm 0.053$ &
  $1.202 \pm 0.062$ &
  $\underline{ 2.366 \pm 0.059 }$ &
  $5.00$ & 5 &
  $0.897 \pm 0.058$ &
  $\mathbf{ 1.196 \pm 0.030 }$ &
  $1.205 \pm 0.032$ &
  $2.105 \pm 0.032$ &
  $3.00$ & 2 \\
& FEATURE-MAP &
  $2.197 \pm 0.075$ &
  $\underline{ 1.148 \pm 0.023 }$ &
  $\mathbf{ 1.163 \pm 0.003 }$ &
  $2.400 \pm 0.175$ &
  $3.00$ & 1 &
  $0.898 \pm 0.040$ &
  $1.200 \pm 0.013$ &
  $1.229 \pm 0.014$ &
  $2.115 \pm 0.031$ &
  $4.25$ & 5 \\
& BSS &
  $\underline{ 1.712 \pm 0.056 }$ &
  $1.168 \pm 0.050$ &
  $1.168 \pm 0.002$ &
  $2.551 \pm 0.121$ &
  $3.25$ & 2 &
  $\underline{ 0.861 \pm 0.024 }$ &
  $1.208 \pm 0.016$ &
  $\underline{ 1.186 \pm 0.019 }$ &
  $\mathbf{ 2.081 \pm 0.037 }$ &
  $2.25$ & 1 \\
\midrule

\multirow{8}{*}{SIZE} &
  FULL-FT &
  $2.654 \pm 0.075$ &
  $1.557 \pm 0.093$ &
  $0.943 \pm 0.026$ &
  $2.550 \pm 0.053$ &
  $4.25$ & 5 &
  $0.886 \pm 0.054$ &
  $1.209 \pm 0.011$ &
  $1.173 \pm 0.017$ &
  $\underline{ 2.098 \pm 0.048 }$ &
  $3.75$ & 4 \\
& LP &
  $2.818 \pm 0.087$ &
  $1.676 \pm 0.115$ &
  $0.963 \pm 0.030$ &
  $5.414 \pm 0.036$ &
  $7.00$ & 8 &
  $1.176 \pm 0.011$ &
  $1.232 \pm 0.007$ &
  $1.181 \pm 0.009$ &
  $2.248 \pm 0.016$ &
  $6.75$ & 7 \\
& SURGICAL-FT &
  $2.658 \pm 0.088$ &
  $1.641 \pm 0.114$ &
  $0.929 \pm 0.027$ &
  $3.423 \pm 0.550$ &
  $5.75$ & 6 &
  $3.589 \pm 0.101$ &
  $2.168 \pm 0.089$ &
  $1.192 \pm 0.010$ &
  $4.258 \pm 0.095$ &
  $8.00$ & 8 \\
& LP-FT &
  $\mathbf{ 2.440 \pm 0.056 }$ &
  $\mathbf{ 1.422 \pm 0.111 }$ &
  $1.166 \pm 0.053$ &
  $\mathbf{ 2.339 \pm 0.049 }$ &
  $2.75$ & 1 &
  $1.049 \pm 0.186$ &
  $1.204 \pm 0.047$ &
  $1.174 \pm 0.016$ &
  $2.167 \pm 0.129$ &
  $5.25$ & 6 \\
& WISE-FT &
  $3.050 \pm 0.087$ &
  $\underline{ 1.513 \pm 0.049 }$ &
  $0.969 \pm 0.001$ &
  $3.223 \pm 0.224$ &
  $5.75$ & 7 &
  $1.045 \pm 0.054$ &
  $1.230 \pm 0.015$ &
  $\underline{ 1.171 \pm 0.025 }$ &
  $2.126 \pm 0.039$ &
  $4.50$ & 5 \\
& L2-SP &
  $ 2.606 \pm 0.085 $ &
  $1.614 \pm 0.112$ &
  $\mathbf{ 0.914 \pm 0.016 }$ &
  $2.466 \pm 0.079$ &
  $3.00$ & 2 &
  $\underline{ 0.851 \pm 0.036 }$ &
  $\underline{ 1.194 \pm 0.015 }$ &
  $\mathbf{ 1.169 \pm 0.005 }$ &
  $2.101 \pm 0.022$ &
  $2.00$ & 1 \\
& Feature-map &
  $2.630 \pm 0.036$ &
  $1.697 \pm 0.080$ &
  $\underline{ 0.920 \pm 0.007 }$ &
  $\underline{ 2.408 \pm 0.057 }$ &
  $4.00$ & 4 &
  $0.867 \pm 0.035$ &
  $\mathbf{ 1.180 \pm 0.014 }$ &
  $1.183 \pm 0.007$ &
  $\mathbf{ 2.039 \pm 0.022 }$ &
  $3.00$ & 3 \\
& BSS &
  $\underline{ 2.579 \pm 0.066 }$ &
  $1.613 \pm 0.110$ &
  $0.926 \pm 0.018$ &
  $2.580 \pm 0.157$ &
  $3.50$ & 3 &
  $\mathbf{ 0.844 \pm 0.007 }$ &
  $1.200 \pm 0.028$ &
  $1.171 \pm 0.033$ &
  $2.104 \pm 0.032$ &
  $2.75$ & 2 \\
\midrule

    \multicolumn{14}{c}{{\textbf{Fewshot-100}}}\\
  \midrule
\multirow{8}{*}{Random} &
  FULL-FT &
  $\mathbf{1.304 \pm 0.041}$ &
  $1.239 \pm 0.032$ &
  $1.289 \pm 0.003$ &
  $3.028 \pm 0.310$ &
  $3.25$ &
  1 &
  $0.412 \pm 0.033$ &
  $1.061 \pm 0.017$ &
  $\underline{1.140 \pm 0.016}$ &
  $1.976 \pm 0.031$ &
  $3.00$ &
  3 \\
& LP &
  $1.609 \pm 0.032$ &
  $1.285 \pm 0.043$ &
  $1.334 \pm 0.009$ &
  $4.562 \pm 0.047$ &
  $7.50$ &
  8 &
  $0.902 \pm 0.037$ &
  $1.185 \pm 0.007$ &
  $1.174 \pm 0.004$ &
  $2.239 \pm 0.010$ &
  $7.25$ &
  7 \\
& Surgical-FT &
  $1.356 \pm 0.022$ &
  $\mathbf{1.219 \pm 0.016}$ &
  $1.298 \pm 0.008$ &
  $3.100 \pm 0.805$ &
  $4.50$ &
  5 &
  $3.371 \pm 0.120$ &
  $1.925 \pm 0.045$ &
  $1.162 \pm 0.013$ &
  $4.076 \pm 0.046$ &
  $7.50$ &
  8 \\
& LP-FT &
  $\underline{1.310 \pm 0.021}$ &
  $\underline{1.226 \pm 0.021}$ &
  $1.374 \pm 0.045$ &
  $3.241 \pm 0.438$ &
  $4.75$ &
  6 &
  $0.735 \pm 0.230$ &
  $1.144 \pm 0.049$ &
  $1.153 \pm 0.030$ &
  $2.158 \pm 0.100$ &
  $5.50$ &
  6 \\
& WiSE-FT &
  $1.600 \pm 0.051$ &
  $1.324 \pm 0.013$ &
  $\mathbf{1.245 \pm 0.017}$ &
  $2.294 \pm 0.024$ &
  $4.75$ &
  7 &
  $0.671 \pm 0.104$ &
  $1.068 \pm 0.049$ &
  $1.159 \pm 0.036$ &
  $2.017 \pm 0.095$ &
  $5.00$ &
  5 \\
& L2-SP &
  $1.323 \pm 0.034$ &
  $1.253 \pm 0.029$ &
  $\underline{1.276 \pm 0.014}$ &
  $\mathbf{2.271 \pm 0.065}$ &
  $3.25$ &
  2 &
  $\mathbf{0.405 \pm 0.034}$ &
  $\underline{1.055 \pm 0.022}$ &
  $\mathbf{1.129 \pm 0.016}$ &
  $\underline{1.951 \pm 0.045}$ &
  $1.75$ &
  1 \\
& Feature-map &
  $1.526 \pm 0.030$ &
  $1.243 \pm 0.027$ &
  $1.276 \pm 0.004$ &
  $\underline{2.271 \pm 0.116}$ &
  $3.75$ &
  3 &
  $0.422 \pm 0.021$ &
  $\mathbf{1.014 \pm 0.006}$ &
  $1.170 \pm 0.013$ &
  $\mathbf{1.883 \pm 0.012}$ &
  $3.25$ &
  4 \\
& BSS &
  $1.322 \pm 0.033$ &
  $1.251 \pm 0.028$ &
  $1.293 \pm 0.006$ &
  $2.541 \pm 0.128$ &
  $4.25$ &
  4 &
  $\underline{0.405 \pm 0.060}$ &
  $\underline{1.050 \pm 0.003}$ &
  $1.147 \pm 0.014$ &
  $1.980 \pm 0.018$ &
  $2.75$ &
  2 \\
\midrule

\multirow{8}{*}{SCAFFOLD} &
  FULL-FT &
  $1.695 \pm 0.045$ &
  $1.168 \pm 0.030$ &
  $1.167 \pm 0.003$ &
  $3.087 \pm 0.765$ &
  $4.50$ & 2 &
  $0.497 \pm 0.045$ &
  $1.125 \pm 0.034$ &
  $1.215 \pm 0.015$ &
  $2.036 \pm 0.073$ &
  $4.75$ & 6 \\
& LP &
  $2.045 \pm 0.044$ &
  $1.211 \pm 0.064$ &
  $1.173 \pm 0.004$ &
  $4.579 \pm 0.037$ &
  $7.50$ & 8 &
  $0.971 \pm 0.036$ &
  $1.185 \pm 0.008$ &
  $1.174 \pm 0.004$ &
  $2.247 \pm 0.005$ &
  $6.25$ & 5 \\
& SURGICAL-FT &
  $1.693 \pm 0.019$ &
  $\underline{1.146 \pm 0.017}$ &
  $1.169 \pm 0.003$ &
  $3.226 \pm 0.563$ &
  $4.50$ & 1 &
  $3.386 \pm 0.120$ &
  $1.927 \pm 0.041$ &
  $1.162 \pm 0.013$ &
  $4.073 \pm 0.048$ &
  $7.00$ & 8 \\
& LP-FT &
  $\mathbf{1.626 \pm 0.016}$ &
  $\mathbf{1.123 \pm 0.011}$ &
  $1.312 \pm 0.023$ &
  $2.782 \pm 0.364$ &
  $3.75$ & 5 &
  $0.730 \pm 0.236$ &
  $1.136 \pm 0.029$ &
  $\underline{1.154 \pm 0.029}$ &
  $2.167 \pm 0.117$ &
  $4.75$ & 3 \\
& WiSE-FT &
  $2.069 \pm 0.066$ &
  $1.205 \pm 0.014$ &
  $\mathbf{1.158 \pm 0.008}$ &
  $\mathbf{2.244 \pm 0.068}$ &
  $4.25$ & 7 &
  $1.069 \pm 0.332$ &
  $1.124 \pm 0.023$ &
  $1.228 \pm 0.016$ &
  $2.143 \pm 0.115$ &
  $6.00$ & 7 \\
& L2-SP &
  $\underline{1.679 \pm 0.045}$ &
  $1.201 \pm 0.048$ &
  $1.168 \pm 0.003$ &
  $\underline{2.327 \pm 0.030}$ &
  $3.50$ & 4 &
  $0.497 \pm 0.060$ &
  $1.098 \pm 0.015$ &
  $1.155 \pm 0.022$ &
  $2.031 \pm 0.061$ &
  $3.25$ & 2 \\
& FEATURE-MAP &
  $1.964 \pm 0.034$ &
  $1.164 \pm 0.029$ &
  $\underline{1.164 \pm 0.001}$ &
  $2.341 \pm 0.095$ &
  $3.50$ & 6 &
  $\underline{0.489 \pm 0.040}$ &
  $\mathbf{1.039 \pm 0.014}$ &
  $1.185 \pm 0.010$ &
  $\underline{2.008 \pm 0.022}$ &
  $2.75$ & 4 \\
& BSS &
  $1.681 \pm 0.043$ &
  $1.191 \pm 0.046$ &
  $1.169 \pm 0.004$ &
  $2.566 \pm 0.149$ &
  $4.50$ & 3 &
  $\mathbf{0.396 \pm 0.010}$ &
  $\underline{1.054 \pm 0.033}$ &
  $\mathbf{1.139 \pm 0.005}$ &
  $\mathbf{1.972 \pm 0.010}$ &
  $1.25$ & 1 \\
\midrule

\multirow{8}{*}{SIZE} &
  FULL-FT &
  $\underline{2.414 \pm 0.081}$ &
  $\underline{1.283 \pm 0.070}$ &
  $0.911 \pm 0.008$ &
  $2.677 \pm 0.139$ &
  $3.00$ & 1 &
  $0.431 \pm 0.059$ &
  $1.039 \pm 0.026$ &
  $\mathbf{1.118 \pm 0.014}$ &
  $1.968 \pm 0.056$ &
  $3.25$ & 4 \\
& LP &
  $2.859 \pm 0.078$ &
  $1.493 \pm 0.115$ &
  $0.951 \pm 0.030$ &
  $5.420 \pm 0.033$ &
  $7.50$ & 8 &
  $0.901 \pm 0.037$ &
  $1.192 \pm 0.007$ &
  $1.163 \pm 0.004$ &
  $2.236 \pm 0.011$ &
  $7.25$ & 7 \\
& SURGICAL-FT &
  $2.537 \pm 0.059$ &
  $1.301 \pm 0.074$ &
  $\underline{0.909 \pm 0.003}$ &
  $3.707 \pm 0.589$ &
  $4.75$ & 6 &
  $3.386 \pm 0.120$ &
  $1.933 \pm 0.038$ &
  $1.151 \pm 0.012$ &
  $4.077 \pm 0.040$ &
  $7.50$ & 8 \\
& LP-FT &
  $\mathbf{2.217 \pm 0.047}$ &
  $\mathbf{1.146 \pm 0.022}$ &
  $1.065 \pm 0.020$ &
  $2.562 \pm 0.076$ &
  $3.50$ & 2 &
  $0.733 \pm 0.232$ &
  $1.166 \pm 0.029$ &
  $1.147 \pm 0.022$ &
  $2.138 \pm 0.123$ &
  $5.50$ & 6 \\
& WISE-FT &
  $2.507 \pm 0.098$ &
  $1.297 \pm 0.038$ &
  $\mathbf{0.904 \pm 0.002}$ &
  $2.823 \pm 0.031$ &
  $3.75$ & 3 &
  $0.708 \pm 0.099$ &
  $1.079 \pm 0.040$ &
  $1.147 \pm 0.040$ &
  $1.987 \pm 0.050$ &
  $4.75$ & 5 \\
& L2-SP &
  $2.442 \pm 0.047$ &
  $1.362 \pm 0.082$ &
  $0.916 \pm 0.009$ &
  $\underline{2.451 \pm 0.093}$ &
  $4.50$ & 5 &
  $\underline{0.409 \pm 0.024}$ &
  $\underline{1.037 \pm 0.030}$ &
  $\underline{1.125 \pm 0.016}$ &
  $\underline{1.942 \pm 0.032}$ &
  $2.00$ & 1 \\
& FEATURE-MAP &
  $2.716 \pm 0.026$ &
  $1.551 \pm 0.085$ &
  $0.912 \pm 0.003$ &
  $\mathbf{2.424 \pm 0.039}$ &
  $5.00$ & 7 &
  $0.419 \pm 0.016$ &
  $\mathbf{1.009 \pm 0.013}$ &
  $1.160 \pm 0.010$ &
  $\mathbf{1.886 \pm 0.031}$ &
  $3.00$ & 3 \\
& BSS &
  $2.434 \pm 0.046$ &
  $1.358 \pm 0.084$ &
  $0.912 \pm 0.005$ &
  $2.533 \pm 0.103$ &
  $3.75$ & 3 &
  $\mathbf{0.387 \pm 0.020}$ &
  $1.038 \pm 0.021$ &
  $1.136 \pm 0.013$ &
  $1.967 \pm 0.023$ &
  $2.50$ & 2 \\
\midrule

  \multicolumn{14}{c}{{\textbf{Fewshot-500}}}\\
  \midrule
\multirow{8}{*}{RANDOM} &
  FULL-FT &
  $1.042 \pm 0.017$ &
  $1.023 \pm 0.022$ &
  $1.290 \pm 0.004$ &
  $1.958 \pm 0.038$ &
  $4.00$ & 5 &
  $\underline{0.135 \pm 0.019}$ &
  $0.070 \pm 0.005$ &
  $0.787 \pm 0.009$ &
  $1.554 \pm 0.044$ &
  $3.25$ & 3 \\
& LP &
  $1.487 \pm 0.011$ &
  $1.233 \pm 0.019$ &
  $1.331 \pm 0.012$ &
  $4.602 \pm 0.019$ &
  $8.00$ & 8 &
  $0.769 \pm 0.108$ &
  $0.854 \pm 0.008$ &
  $1.035 \pm 0.001$ &
  $1.941 \pm 0.004$ &
  $6.00$ & 6 \\
& SURGICAL-FT &
  $1.164 \pm 0.010$ &
  $1.127 \pm 0.007$ &
  $\underline{1.240 \pm 0.011}$ &
  $3.577 \pm 0.498$ &
  $5.00$ & 7 &
  $2.376 \pm 0.207$ &
  $0.806 \pm 0.037$ &
  $0.803 \pm 0.010$ &
  $3.058 \pm 0.054$ &
  $6.75$ & 7 \\
& LP-FT &
  $\mathbf{0.995 \pm 0.010}$ &
  $\mathbf{0.975 \pm 0.007}$ &
  $1.310 \pm 0.019$ &
  $2.004 \pm 0.056$ &
  $3.75$ & 4 &
  $0.545 \pm 0.293$ &
  $0.605 \pm 0.352$ &
  $0.793 \pm 0.018$ &
  $1.566 \pm 0.027$ &
  $5.50$ & 5 \\
& WiSE-FT &
  $1.251 \pm 0.029$ &
  $\underline{0.976 \pm 0.010}$ &
  $\mathbf{1.231 \pm 0.016}$ &
  $1.975 \pm 0.017$ &
  $3.25$ & 2 &
  $2.512 \pm 0.245$ &
  $1.563 \pm 0.200$ &
  $1.197 \pm 0.017$ &
  $2.177 \pm 0.063$ &
  $7.75$ & 8 \\
& L2-SP &
  $1.048 \pm 0.014$ &
  $1.036 \pm 0.009$ &
  $1.241 \pm 0.007$ &
  $\mathbf{1.886 \pm 0.032}$ &
  $3.25$ & 1 &
  $0.141 \pm 0.043$ &
  $0.080 \pm 0.026$ &
  $\underline{0.781 \pm 0.010}$ &
  $1.549 \pm 0.022$ &
  $2.75$ & 2 \\
& FEATURE-MAP &
  $1.340 \pm 0.007$ &
  $1.202 \pm 0.014$ &
  $1.241 \pm 0.007$ &
  $1.992 \pm 0.013$ &
  $5.75$ & 6 &
  $0.155 \pm 0.021$ &
  $\underline{0.104 \pm 0.005}$ &
  $\mathbf{0.778 \pm 0.004}$ &
  $1.565 \pm 0.028$ &
  $3.25$ & 4 \\
& BSS &
  $\underline{1.031 \pm 0.013}$ &
  $1.020 \pm 0.006$ &
  $1.272 \pm 0.007$ &
  $\underline{1.896 \pm 0.034}$ &
  $3.00$ & 3 &
  $\mathbf{0.129 \pm 0.018}$ &
  $\mathbf{0.018 \pm 0.004}$ &
  $0.779 \pm 0.007$ &
  $\mathbf{1.543 \pm 0.028}$ &
  $1.25$ & 1 \\
\midrule

\multirow{8}{*}{SCAFFOLD} &
  FULL-FT &
  $1.406 \pm 0.016$ &
  $0.945 \pm 0.021$ &
  $1.199 \pm 0.025$ &
  $2.057 \pm 0.072$ &
  $4.75$ & 5 &
  $0.145 \pm 0.023$ &
  $0.072 \pm 0.005$ &
  $\mathbf{ 0.776 \pm 0.006 }$ &
  $1.564 \pm 0.033$ &
  $2.50$ & 3 \\
& LP &
  $1.849 \pm 0.028$ &
  $1.102 \pm 0.019$ &
  $1.182 \pm 0.007$ &
  $4.607 \pm 0.020$ &
  $7.00$ & 8 &
  $0.771 \pm 0.018$ &
  $0.854 \pm 0.008$ &
  $1.035 \pm 0.001$ &
  $1.941 \pm 0.004$ &
  $6.50$ & 6 \\
& SURGICAL-FT &
  $1.436 \pm 0.010$ &
  $1.020 \pm 0.006$ &
  $1.156 \pm 0.010$ &
  $2.874 \pm 0.652$ &
  $5.00$ & 6 &
  $2.377 \pm 0.207$ &
  $0.805 \pm 0.041$ &
  $0.802 \pm 0.011$ &
  $3.053 \pm 0.051$ &
  $6.50$ & 6 \\
& LP-FT &
  $\mathbf{ 1.354 \pm 0.011 }$ &
  $\mathbf{ 0.940 \pm 0.012 }$ &
  $1.278 \pm 0.044$ &
  $2.052 \pm 0.053$ &
  $3.75$ & 4 &
  $0.546 \pm 0.293$ &
  $0.605 \pm 0.352$ &
  $0.949 \pm 0.123$ &
  $1.803 \pm 0.196$ &
  $5.25$ & 5 \\
& WiSE-FT &
  $1.707 \pm 0.029$ &
  $1.028 \pm 0.025$ &
  $\mathbf{ 1.125 \pm 0.008 }$ &
  $\mathbf{ 1.906 \pm 0.020 }$ &
  $3.50$ & 3 &
  $2.476 \pm 0.626$ &
  $1.459 \pm 0.258$ &
  $1.207 \pm 0.030$ &
  $2.173 \pm 0.061$ &
  $7.75$ & 8 \\
& L2-SP &
  $1.413 \pm 0.045$ &
  $0.943 \pm 0.022$ &
  $1.156 \pm 0.012$ &
  $1.931 \pm 0.054$ &
  $3.25$ & 2 &
  $\underline{ 0.137 \pm 0.017 }$ &
  $\underline{ 0.070 \pm 0.009 }$ &
  $0.782 \pm 0.005$ &
  $\underline{ 1.524 \pm 0.014 }$ &
  $2.25$ & 2 \\
& FEATURE-MAP &
  $1.880 \pm 0.021$ &
  $1.081 \pm 0.006$ &
  $\underline{ 1.129 \pm 0.006 }$ &
  $1.992 \pm 0.008$ &
  $5.25$ & 7 &
  $0.163 \pm 0.010$ &
  $0.111 \pm 0.002$ &
  $0.786 \pm 0.005$ &
  $1.592 \pm 0.013$ &
  $4.00$ & 4 \\
& BSS &
  $\underline{ 1.404 \pm 0.042 }$ &
  $\underline{ 0.941 \pm 0.019 }$ &
  $1.199 \pm 0.029$ &
  $\underline{ 1.926 \pm 0.041 }$ &
  $3.00$ & 1 &
  $\mathbf{ 0.127 \pm 0.015 }$ &
  $\mathbf{ 0.068 \pm 0.004 }$ &
  $\underline{ 0.777 \pm 0.008 }$ &
  $\mathbf{ 1.513 \pm 0.007 }$ &
  $1.25$ & 1 \\
\midrule

\multirow{8}{*}{Size} &
  Full-FT      &  $2.102 \pm 0.080$  &  $\underline{0.968 \pm 0.032}$  &  $0.955 \pm 0.031$  &  $2.283 \pm 0.060$  &  $3.50$  &  4  &  $0.142 \pm 0.049$  &  $0.070 \pm 0.003$  &  $0.723 \pm 0.008$  &  $1.548 \pm 0.011$  &  $3.00$  &  3 \\
 &
  LP           &  $2.486 \pm 0.040$  &  $1.140 \pm 0.046$  &  $0.968 \pm 0.027$  &  $5.452 \pm 0.018$  &  $7.50$  &  8  &  $0.771 \pm 0.018$  &  $0.855 \pm 0.009$  &  $1.008 \pm 0.004$  &  $1.938 \pm 0.004$  &  $6.50$  &  6 \\
 &
  Surgical-FT  &  $2.142 \pm 0.062$  &  $0.982 \pm 0.014$  &  $0.949 \pm 0.032$  &  $3.765 \pm 0.499$  &  $4.50$  &  7  &  $2.384 \pm 0.212$  &  $0.812 \pm 0.042$  &  $0.745 \pm 0.011$  &  $3.070 \pm 0.035$  &  $6.50$  &  7 \\
 &
  LP-FT        &  $\underline{2.003 \pm 0.037}$  &  $\mathbf{0.889 \pm 0.017}$  &  $0.985 \pm 0.033$  &  $2.339 \pm 0.049$  &  $3.75$  &  3  &  $0.550 \pm 0.287$  &  $0.812 \pm 0.042$  &  $0.740 \pm 0.027$  &  $\mathbf{1.533 \pm 0.003}$  &  $4.25$  &  5 \\
 &
  WiSE-FT      &  $2.302 \pm 0.057$  &  $1.040 \pm 0.015$  &  $\mathbf{0.906 \pm 0.003}$  &  $2.437 \pm 0.032$  &  $5.00$  &  6  &  $2.559 \pm 0.295$  &  $1.599 \pm 0.242$  &  $1.196 \pm 0.027$  &  $2.189 \pm 0.053$  &  $7.75$  &  8 \\
 &
  L2-SP        &  $2.030 \pm 0.059$  &  $1.012 \pm 0.030$  &  $0.951 \pm 0.030$  &  $\mathbf{2.208 \pm 0.030}$  &  $3.25$  &  2  &  $\mathbf{0.124 \pm 0.013}$  &  $0.073 \pm 0.008$  &  $0.721 \pm 0.008$  &  $\underline{1.533 \pm 0.014}$  &  $2.25$  &  1 \\
 &
  Feature-map  &  $2.253 \pm 0.017$  &  $1.174 \pm 0.023$  &  $\underline{0.908 \pm 0.001}$  &  $2.341 \pm 0.027$  &  $5.25$  &  5  &  $0.157 \pm 0.020$  &  $0.103 \pm 0.005$  &  $\mathbf{0.705 \pm 0.008}$  &  $1.553 \pm 0.014$  &  $3.50$  &  4 \\
 &
  BSS          &  $\mathbf{1.980 \pm 0.051}$  &  $0.989 \pm 0.025$  &  $0.956 \pm 0.041$  &  $\underline{2.237 \pm 0.058}$  &  $3.25$  &  1  &  $\underline{0.126 \pm 0.013}$  &  $\mathbf{0.064 \pm 0.004}$  &  $\underline{0.710 \pm 0.014}$  &  $1.551 \pm 0.028$  &  $2.25$  &  2 \\ 

\midrule[2pt]

\end{tabular}%
\end{sc}
\end{small}
\end{adjustbox}
\end{center}

\end{table*}

\begin{table*}[t]
\vspace{-2mm}
\caption{\proj performance on 4 \hlb{\textbf{Regression}}{30}
datasets (RMSE metrics) in the \hlr{\textbf{Fewshot}}{30} setting with 50,100, 500 samples, evaluated across 3 dataset splits (\textsc{Random, Scaffold, Size}) given \hlg{\textbf{\textsc{Mole-BERT}}}{30} model. \textsc{Avg-R} denote the average rank. Standard deviations across five replicates are shown in parentheses. We \textbf{bold} and \underline{underline} the best and second-best performances in each scenario. 
% \pan{what is TOP?; Why do we need to compare with top? If we cannot get consistently better than top, what is the sense to put it here? We can just leave the statement that our methods achieve the best average rank, and even achieve xxx top performance in xxx (how many over how many) datasets. } \Shikun{Since there is a reviewer questioning if we are comparing DWiSE-FT to all baselines, I'm wondering if we gonna keep the "best" row}
}
\vspace{-2mm}
\begin{center}
\begin{adjustbox}{width = \textwidth}
\begin{small}
\begin{sc}
\begin{tabular}{lcccccc|ccccc|ccccc}
\toprule
& & \multicolumn{5}{c}{Fewshot 50} & \multicolumn{5}{c}{Fewshot 100} & \multicolumn{5}{c}{Fewshot 500}\\
\cmidrule(lr){3-7}\cmidrule(lr){8-12}\cmidrule(lr){13-17}
 Split & Methods  &   esol &  lipo  &     malaria    & cep  & AVG &   esol &  lipo  &     malaria    & cep  & AVG &   esol &  lipo  &     malaria    & cep  & AVG  \\
\midrule
\multirow{4}{*}{Random} 

% & WiSE-FT & $1.384 \pm 0.047 $ & $1.212 \pm 0.020 $ & $1.276 \pm 0.007 $ & $2.410 \pm 0.051 $ & & $1.189 \pm 0.030 $ & $1.142 \pm 0.025 $ & $1.256 \pm 0.006 $ & $2.211 \pm 0.028 $ & & $0.995 \pm 0.010 $ & $0.855 \pm 0.011 $ & $1.193 \pm 0.003 $ & $1.893 \pm 0.021 $\\
% & $L^2$-SP & $1.372  \pm 0.029$ & $1.196  \pm 0.019$ & $1.277  \pm 0.006$ & $2.280  \pm 0.031$ & & $1.161  \pm 0.016$ & $1.149  \pm 0.007$ & $1.260  \pm 0.004$ & $2.131  \pm 0.014$ & & $0.878  \pm 0.026$ & $0.806  \pm 0.007$ & $1.192  \pm 0.004$ & $1.893  \pm 0.018$\\
% & Top & $1.329 \pm 0.021$ & $1.164  \pm 0.010$ & $1.271  \pm 0.007$ & $2.275  \pm 0.022$ & & $1.120 \pm 0.038$ & $1.139 \pm 0.017$ & $1.256 \pm 0.006$ & $2.131 \pm 0.014$ & & $ 0.878\pm 0.026$ & $0.806 \pm 0.007$ & $1.192 \pm 0.004$ & $1.862 \pm 0.010$\\
% & DWiSE-FT & $ 1.378 \pm 0.055$ & $ 1.189 \pm 0.020$ & $ 1.273 \pm 0.009$ & $ 2.222 \pm 0.059$ & & $ 1.132\pm 0.025$ & $ 1.138\pm 0.028$ & $ 1.256\pm 0.004$ & $ 2.129\pm 0.020$ & & $ 0.918\pm 0.012$ & $ 0.818\pm 0.013$ & $ 1.192\pm 0.004$ & $ 1.865\pm 0.030$\\

&  WiSE-FT  &  $1.384 \pm 0.047 $  &  $1.212 \pm 0.020 $  &  $1.276 \pm 0.007 $  &  $2.410 \pm 0.051 $  & $3.75$ &  $1.189 \pm 0.030 $  &  $1.142 \pm 0.025 $  & $\mathbf{ 1.256 \pm 0.006  }$ &  $2.211 \pm 0.028 $  & $3.00$ &  $0.995 \pm 0.010 $  &  $0.855 \pm 0.011 $  &  $1.193 \pm 0.003 $  &  $1.893 \pm 0.021 $ & $3.75$ \\
 &  $L^2$-SP  & $\underline{ 1.372  \pm 0.029 }$ &  $1.196  \pm 0.019$  &  $1.277  \pm 0.006$  &  $2.280  \pm 0.031$  & $3.00$ &  $1.161  \pm 0.016$  &  $1.149  \pm 0.007$  &  $1.260  \pm 0.004$  & $\underline{ 2.131  \pm 0.014 }$ & $3.25$ & $\mathbf{ 0.878  \pm 0.026 }$ & $\mathbf{ 0.806  \pm 0.007 }$ & $\mathbf{ 1.192  \pm 0.004 }$ &  $1.893  \pm 0.018$ & $1.50$ \\
 &  Top  & $\mathbf{ 1.329 \pm 0.021 }$ & $\mathbf{ 1.164  \pm 0.010 }$ & $\mathbf{ 1.271  \pm 0.007 }$ & $\underline{ 2.275  \pm 0.022 }$ & $1.25$ & $\mathbf{ 1.120 \pm 0.038 }$ & $\underline{ 1.139 \pm 0.017 }$ & $\mathbf{ 1.256 \pm 0.006 }$ & $\underline{ 2.131 \pm 0.014 }$ & $1.50$ & $\mathbf{  0.878\pm 0.026 }$ & $\mathbf{ 0.806 \pm 0.007 }$ & $\mathbf{ 1.192 \pm 0.004 }$ & $\mathbf{ 1.862 \pm 0.010}$ & $1.00$ \\
 &  DWiSE-FT  &  $ 1.378 \pm 0.055$  & $\underline{  1.189 \pm 0.020 }$ & $\underline{  1.273 \pm 0.009 }$ & $\mathbf{  2.222 \pm 0.059 }$ & $2.00$ & $\underline{  1.132\pm 0.025 }$ & $\mathbf{  1.138\pm 0.028 }$ & $\mathbf{  1.256\pm 0.004 }$ & $\mathbf{  2.129\pm 0.020 }$ & $1.25$ &  $ 0.918\pm 0.012$  &  $ 0.818\pm 0.013$  & $\mathbf{  1.192\pm 0.004 }$ & $\underline{  1.865\pm 0.030}$ & $2.25$ \\

\midrule
\multirow{4}{*}{Scaffold} 

% & WiSE-FT & $1.842 \pm 0.056 $ & $1.177 \pm 0.009 $ & $1.162 \pm 0.004 $ & $2.454 \pm 0.043 $ & & $1.544 \pm 0.063 $ & $1.041 \pm 0.017 $ & $1.151 \pm 0.007 $ & $2.301 \pm 0.042 $ & & $1.388 \pm 0.023 $ & $0.834 \pm 0.012 $ & $1.114 \pm 0.002 $ & $1.936 \pm 0.037 $\\
% & $L^2$-SP & $1.699  \pm 0.049$ & $1.086  \pm 0.009$ & $1.162  \pm 0.002$ & $2.331  \pm 0.024$ & & $1.473  \pm 0.009$ & $0.961  \pm 0.003$ & $1.153  \pm 0.002$ & $2.201  \pm 0.038$ & & $1.163  \pm 0.026$ & $0.813  \pm 0.010$ & $1.126  \pm 0.011$ & $1.885  \pm 0.011$\\
% & Top & $1.680  \pm 0.042$ & $1.036  \pm 0.007$ & $1.159  \pm 0.000$ & $2.292  \pm 0.026$ & & $1.436 \pm 0.054$ & $0.937 \pm 0.008$ & $1.149 \pm 0.003$ & $2.187 \pm 0.034$ & & $ 1.112\pm 0.015$ & $0.802 \pm 0.003$ & $1.114 \pm 0.002$ & $1.881 \pm 0.010$\\
% & DWiSE-FT & $ 1.616 \pm 0.047$ & $ 1.110 \pm 0.013$ & $ 1.173 \pm 0.005$ & $ 2.306 \pm 0.030$ & & $ 1.485\pm 0.041$ & $ 0.979\pm 0.014$ & $ 1.158\pm 0.009$ & $2.149 \pm 0.040$ & & $ 1.266\pm 0.021$ & $ 0.823\pm 0.010$ & $ 1.121\pm 0.004$ & $ 1.900\pm 0.019$\\

&  WiSE-FT  &  $1.842 \pm 0.056 $  &  $1.177 \pm 0.009 $  & $\underline{ 1.162 \pm 0.004  }$ &  $2.454 \pm 0.043 $  & $3.50$ &  $1.544 \pm 0.063 $  &  $1.041 \pm 0.017 $  & $\underline{ 1.151 \pm 0.007  }$ &  $2.301 \pm 0.042 $  & $3.50$ &  $1.388 \pm 0.023 $  &  $0.834 \pm 0.012 $  & $\mathbf{ 1.114 \pm 0.002  }$ &  $1.936 \pm 0.037 $ & $3.25$ \\
 &  $L^2$-SP  &  $1.699  \pm 0.049$  & $\underline{ 1.086  \pm 0.009 }$ & $\underline{ 1.162  \pm 0.002 }$ &  $2.331  \pm 0.024$  & $2.50$ & $\underline{ 1.473  \pm 0.009 }$ & $\underline{ 0.961  \pm 0.003 }$ &  $1.153  \pm 0.002$  &  $2.201  \pm 0.038$  & $2.50$ & $\underline{ 1.163  \pm 0.026 }$ & $\underline{ 0.813  \pm 0.010 }$ &  $1.126  \pm 0.011$  & $\underline{ 1.885  \pm 0.011}$ & $2.50$ \\
 &  Top  & $\underline{ 1.680  \pm 0.042 }$ & $\mathbf{ 1.036  \pm 0.007 }$ & $\mathbf{ 1.159  \pm 0.000 }$ & $\mathbf{ 2.292  \pm 0.026 }$ & $1.25$ & $\mathbf{ 1.436 \pm 0.054 }$ & $\mathbf{ 0.937 \pm 0.008 }$ & $\mathbf{ 1.149 \pm 0.003 }$ & $\underline{ 2.187 \pm 0.034 }$ & $1.25$ & $\mathbf{  1.112\pm 0.015 }$ & $\mathbf{ 0.802 \pm 0.003 }$ & $\mathbf{ 1.114 \pm 0.002 }$ & $\mathbf{ 1.881 \pm 0.010}$ & $1.00$ \\
 &  DWiSE-FT  & $\mathbf{  1.616 \pm 0.047 }$ &  $ 1.110 \pm 0.013$  &  $ 1.173 \pm 0.005$  & $\underline{  2.306 \pm 0.030 }$ & $2.50$ &  $ 1.485\pm 0.041$  &  $ 0.979\pm 0.014$  &  $ 1.158\pm 0.009$  & $\mathbf{ 2.149 \pm 0.040 }$ & $2.75$ &  $ 1.266\pm 0.021$  &  $ 0.823\pm 0.010$  &  $ 1.121\pm 0.004$  &  $ 1.900\pm 0.019$ & $3.00$ \\

\midrule
\multirow{4}{*}{Size} 

% & WiSE-FT & $2.615 \pm 0.072 $ & $1.391 \pm 0.042 $ & $0.929 \pm 0.004 $ & $2.762 \pm 0.053 $ & & $2.216 \pm 0.056 $ & $1.124 \pm 0.031 $ & $0.917 \pm 0.004 $ & $2.543 \pm 0.027 $ & & $2.071 \pm 0.078 $ & $0.902 \pm 0.016 $ & $0.912 \pm 0.003 $ & $2.379 \pm 0.086 $\\
% & $L^2$-SP & $2.393  \pm 0.068$ & $1.306  \pm 0.037$ & $0.915  \pm 0.002$ & $2.497  \pm 0.019$ & & $1.731  \pm 0.071$ & $1.025  \pm 0.028$ & $0.905  \pm 0.002$ & $2.424  \pm 0.024$ & & $1.629  \pm 0.084$ & $0.821  \pm 0.011$ & $0.904  \pm 0.003$ & $2.368  \pm 0.013$\\
% & Top & $2.369  \pm 0.075$ & $1.297  \pm 0.040$ & $0.911  \pm 0.002$ & $2.497  \pm 0.019$ & & $1.731 \pm 0.071$ & $1.025 \pm 0.028$ & $0.898 \pm 0.003$ & $2.424 \pm 0.024$ & & $1.629 \pm 0.084$ & $0.803 \pm 0.006$ & $0.895 \pm 0.002$ & $2.328 \pm 0.017$\\
% & DWiSE-FT & $ 1.488 \pm 0.101$ & $ 1.113 \pm 0.021$ & $ 0.913 \pm 0.007$ & $ 2.539 \pm 0.023$ & & $ 1.469\pm 0.052$ & $ 1.031\pm 0.022$ & $ 0.920\pm 0.006$ & $2.390 \pm 0.025$ & & $ 1.466\pm 0.040$ & $ 0.816\pm 0.022$ & $0.915 \pm 0.003 $ & $ 2.322\pm 0.031$\\

&  WiSE-FT  &  $2.615 \pm 0.072 $  &  $1.391 \pm 0.042 $  &  $0.929 \pm 0.004 $  &  $2.762 \pm 0.053 $  & $4.00$ &  $2.216 \pm 0.056 $  &  $1.124 \pm 0.031 $  &  $0.917 \pm 0.004 $  &  $2.543 \pm 0.027 $  & $3.75$ &  $2.071 \pm 0.078 $  &  $0.902 \pm 0.016 $  &  $0.912 \pm 0.003 $  &  $2.379 \pm 0.086 $ & $3.75$ \\
 &  $L^2$-SP  &  $2.393  \pm 0.068$  &  $1.306  \pm 0.037$  &  $0.915  \pm 0.002$  & $\mathbf{ 2.497  \pm 0.019 }$ & $2.50$ & $\underline{ 1.731  \pm 0.071 }$ & $\mathbf{ 1.025  \pm 0.028 }$ & $\underline{ 0.905  \pm 0.002 }$ & $\underline{ 2.424  \pm 0.024 }$ & $1.75$ & $\underline{ 1.629  \pm 0.084 }$ &  $0.821  \pm 0.011$  & $\underline{ 0.904  \pm 0.003 }$ &  $2.368  \pm 0.013$ & $2.50$ \\
 &  Top  & $\underline{ 2.369  \pm 0.075 }$ & $\underline{ 1.297  \pm 0.040 }$ & $\mathbf{ 0.911  \pm 0.002 }$ & $\mathbf{ 2.497  \pm 0.019 }$ & $1.50$ & $\underline{ 1.731 \pm 0.071 }$ & $\mathbf{ 1.025 \pm 0.028 }$ & $\mathbf{ 0.898 \pm 0.003 }$ & $\underline{ 2.424 \pm 0.024 }$ & $1.50$ & $\underline{ 1.629 \pm 0.084 }$ & $\mathbf{ 0.803 \pm 0.006 }$ & $\mathbf{ 0.895 \pm 0.002 }$ & $\underline{ 2.328 \pm 0.017}$ & $1.50$ \\
 &  DWiSE-FT  & $\mathbf{  1.488 \pm 0.101 }$ & $\mathbf{  1.113 \pm 0.021 }$ & $\underline{  0.913 \pm 0.007 }$ &  $ 2.539 \pm 0.023$  & $1.75$ & $\mathbf{  1.469\pm 0.052 }$ &  $ 1.031\pm 0.022$  &  $ 0.920\pm 0.006$  & $\mathbf{ 2.390 \pm 0.025 }$ & $2.25$ & $\mathbf{  1.466\pm 0.040 }$ & $\underline{  0.816\pm 0.022 }$ &  $0.915 \pm 0.003 $  & $\mathbf{  2.322\pm 0.031}$ & $2.00$ \\

\bottomrule
\label{table:new_method_appendix}
\end{tabular}
\end{sc}
\end{small}
\end{adjustbox}
\end{center}
\vspace{-6mm}
\end{table*}
\vspace{-1.5mm}
\begin{table*}[t]
\centering
\caption{\proj performance on 2 \hlb{\textbf{Regression}}{30}
datasets (RMSE metrics) and 2 \hlb{\textbf{Classification}}{30} datasets (AUC) in the \hlr{\textbf{Fewshot}}{30} setting with 50 samples, evaluated across dataset splits (\textsc{Scaffold, Size}) given \hlg{\textbf{\textsc{GraphGPS}}}{30} model. \textsc{Avg-R} denote the average rank. Standard deviations across five replicates are shown in parentheses. We \textbf{bold} and \underline{underline} the best and second-best performances in each scenario. }
\resizebox{\textwidth}{!}{
\begin{tabular}{lccc|ccc|ccc|ccc}
\toprule
\multirow{2}{*}{\textbf{Method}} 
& \multicolumn{6}{c}{\textbf{Few-Shot 50 (Scaffold Split)}}  
& \multicolumn{6}{c}{\textbf{Few-Shot 50 (Size Split)}} \\
\cmidrule(lr){2-7} \cmidrule(lr){8-13}
& \textbf{BACE (AUC)} & \textbf{SIDER (AUC)} & \textbf{Avg AUC} & \textbf{ESOL (RMSE)} & \textbf{LIPO (RMSE)} & \textbf{Avg R}
& \textbf{BACE (AUC)} & \textbf{SIDER (AUC)} & \textbf{Avg AUC} & \textbf{ESOL (RMSE)} & \textbf{LIPO (RMSE)} &  \textbf{Avg R} \\
\midrule
% WiSE-FT & $54.67 \pm 0.12$ & $83.73 \pm 0.00$ & $69.23$ & $1.020 \pm 0.045$ & $1.259 \pm 0.027$ & $3$ 
% & $53.03 \pm 2.64$ & $83.77 \pm 0.34$ & $68.40$ & $1.045 \pm 0.054$ & $1.230 \pm 0.015$ & $4$ \\
% L2-SP   & $59.52 \pm 3.80$ & $84.94 \pm 0.16$ & $72.23$ & $0.897 \pm 0.058$ & $1.196 \pm 0.030$ & $2$ 
% & $61.93 \pm 3.45$ & $85.12 \pm 0.28$ & $73.53$ & $0.851 \pm 0.036$ & $1.194 \pm 0.015$ & $2.5$ \\
% TOP     & $64.90 \pm 1.55$ & $85.12 \pm 0.23$ & $75.01$ & $0.859 \pm 0.065$ & $1.196 \pm 0.030$ & $1.5$ 
% & $61.93 \pm 3.45$ & $86.48 \pm 0.70$ & $74.21$ & $0.844 \pm 0.007$ & $1.180 \pm 0.014$ & $1.5$ \\
% DWiSE-FT& $64.82 \pm 1.53$ & $85.23 \pm 0.02$ & $75.03$ & $0.859 \pm 0.071$ & $1.190 \pm 0.016$ & $1$ 
% & $61.46 \pm 0.57$ & $85.39 \pm 0.23$ & $73.43$ & $0.868 \pm 0.041$ & $1.167 \pm 0.016$ & $2$ \\
% \bottomrule
WiSE-FT 
& $54.67 \pm 0.12$ & $83.73 \pm 0.00$ & $69.23$ & $1.020 \pm 0.045$ & $1.259 \pm 0.027$ & $3$
& $53.03 \pm 2.64$ & $83.77 \pm 0.34$ & $68.40$ & $1.045 \pm 0.054$ & $1.230 \pm 0.015$ & $4$ \\

L2-SP
& $59.52 \pm 3.80$ & $84.94 \pm 0.16$ & $72.23$ & $0.897 \pm 0.058$ & \underline{$1.196 \pm 0.030$} & $2$
& \textbf{\boldmath $61.93 \pm 3.45$} & $85.12 \pm 0.28$ & \underline{$73.53$} & \underline{$0.851 \pm 0.036$} & $1.194 \pm 0.015$ & $2.5$ \\

TOP
& \textbf{\boldmath $64.90 \pm 1.55$} & \underline{$85.12 \pm 0.23$} & \underline{$75.01$} & $\mathbf{0.859 \pm 0.065}$ & $1.196 \pm 0.030$ & \underline{$1.5$}
& \underline{$61.93 \pm 3.45$} & \textbf{\boldmath $86.48 \pm 0.70$} & \textbf{\boldmath $74.21$} & $\mathbf{0.844 \pm 0.007}$ & \underline{$1.180 \pm 0.014$} & $\mathbf{1.5}$ \\

DWiSE-FT
& \underline{$64.82 \pm 1.53$} & \textbf{\boldmath $85.23 \pm 0.02$} & \textbf{\boldmath $75.03$} & \underline{$0.859 \pm 0.071$} & $\mathbf{1.190 \pm 0.016}$ & $\mathbf{1}$
& $61.46 \pm 0.57$ & \underline{$85.39 \pm 0.23$} & $73.43$ & $0.868 \pm 0.041$ & $\mathbf{1.167 \pm 0.016}$ & \underline{$2$} \\
\bottomrule
\label{table:dwise-gps}
\end{tabular}
}
\end{table*}

\begin{table}[ht]
  \centering
  \caption{XGBoost performance on both regression and classification datasets in the Fewshot setting across 3 dataset splits}
  \label{tab:xgboost}
  \footnotesize
  % Resize the entire block to text width, centered automatically
  \resizebox{\textwidth}{!}{%
    % ─────────────── Left half: Classification ───────────────
    \begin{minipage}[t]{0.6\textwidth}
      \centering
      \textbf{Classification tasks}\\[0.5ex]
      \begin{tabular}{c l r r r r}
        \toprule
        \multirow{2}{*}{\#Shots} & \multirow{2}{*}{Split}
          & \multicolumn{4}{c}{Dataset} \\
        \cmidrule(l){3-6}
        & & Clintox & BBBP   & BACE   & HIV    \\
        \midrule
        \multirow{3}{*}{50}  & Random   & 50.00 & 75.25 & 75.13 & 47.75 \\
                             & Scaffold & 68.21 & 57.32 & 58.04 & 50.00 \\
                             & Size     & 50.00 & 62.98 & 61.68 & 52.48 \\
        \midrule
        \multirow{3}{*}{100} & Random   & 68.95 & 70.39 & 82.02 & 47.51 \\
                             & Scaffold & 82.53 & 58.59 & 65.59 & 56.51 \\
                             & Size     & 62.09 & 63.60 & 63.96 & 52.31 \\
        \midrule
        \multirow{3}{*}{500} & Random   & 87.24 & 86.14 & 83.20 & 63.54 \\
                             & Scaffold & 86.06 & 64.43 & 69.26 & 66.03 \\
                             & Size     & 71.75 & 80.51 & 53.16 & 65.41 \\
        \bottomrule
      \end{tabular}
    \end{minipage}
    \hfill
    % ─────────────── Right half: Regression ───────────────
    \begin{minipage}[t]{0.6\textwidth}
      \centering
      \textbf{Regression tasks}\\[0.5ex]
      \begin{tabular}{c l r r r r}
        \toprule
        \multirow{2}{*}{\#Shots} & \multirow{2}{*}{Split}
          & \multicolumn{4}{c}{Dataset} \\
        \cmidrule(l){3-6}
        & & ESOL   & LIPO   & Malaria & CEP     \\
        \midrule
        \multirow{3}{*}{50}  & Random   & 2.1118 & 1.3447 & 1.4396 & 2.3080 \\
                             & Scaffold & 2.3763 & 1.2556 & 1.3096 & 2.6531 \\
                             & Size     & 3.3287 & 1.5481 & 1.2063 & 2.3934 \\
        \midrule
        \multirow{3}{*}{100} & Random   & 2.0708 & 1.2751 & 1.3917 & 2.2813 \\
                             & Scaffold & 2.1859 & 1.2160 & 1.2721 & 2.2624 \\
                             & Size     & 2.8140 & 1.3235 & 1.2349 & 2.4970 \\
        \midrule
        \multirow{3}{*}{500} & Random   & 1.3626 & 1.0906 & 1.3015 & 1.8142 \\
                             & Scaffold & 1.9525 & 1.1078 & 1.2221 & 1.8396 \\
                             & Size     & 2.4934 & 1.0358 & 1.1975 & 2.1820 \\
        \bottomrule
      \end{tabular}
    \end{minipage}%
  } % end \resizebox
\end{table}
\begin{table}[t]
\centering
\caption{LoRA Performance under few-shot and non-fewshot settings across classification and regression datasets with pretrained model GraphGPS.}
\resizebox{\textwidth}{!}{
\begin{tabular}{lccccc|cccc}
\toprule
\textbf{Scaffold Split} & \textbf{clintox} & \textbf{bbbp} & \textbf{bace} & \textbf{hiv} & \textbf{sider} & \textbf{esol} & \textbf{lipo} & \textbf{malaria} & \textbf{cep} \\
\midrule
fewshot 50  & $97.77 \pm 0.21$ & $60.15 \pm 3.83$ & $57.99 \pm 3.48$ & $67.95 \pm 4.21$ & $83.02 \pm 0.24$ & $0.796 \pm 0.032$ & $1.232 \pm 0.039$ & $1.188 \pm 0.008$ & $2.081 \pm 0.082$ \\
fewshot 500 & $99.79 \pm 0.02$ & $100.00 \pm 0.00$ & $99.99 \pm 0.02$ & $80.19 \pm 1.52$ & $92.16 \pm 0.25$ & $0.354 \pm 0.009$ & $0.260 \pm 0.010$ & $0.872 \pm 0.006$ & $1.569 \pm 0.040$ \\
non-fewshot & $99.80 \pm 0.02$ & $99.84 \pm 0.06$ & $99.54 \pm 0.21$ & $94.98 \pm 0.66$ & $90.84 \pm 0.11$ & $0.375 \pm 0.033$ & $0.318 \pm 0.009$ & $0.737 \pm 0.015$ & $0.632 \pm 0.017$ \\
\bottomrule
\label{table:lora}
\end{tabular}
}
\end{table}

\end{document}